\documentclass[11pt]{article}
\usepackage[preprint]{acl}

\usepackage{times}
\usepackage{latexsym}
\usepackage[T1]{fontenc}
\usepackage[utf8]{inputenc}
\usepackage{microtype}

\usepackage{amsmath,amsfonts,amssymb,bm}
\usepackage{mathtools}
\usepackage{float}

\usepackage{xcolor,colortbl,caption}
\usepackage[most]{tcolorbox}
\usepackage{comment}
\usepackage{pifont}
\usepackage{tabularx}
\usepackage{caption, booktabs}
\usepackage{setspace}
\usepackage{arydshln}

\makeatletter
\newcommand\footnoteref[1]{\protected@xdef\@thefnmark{\ref{#1}}\@footnotemark}
\makeatother
\newcolumntype{P}[1]{>{\centering\arraybackslash}p{#1}}

\usepackage{xspace}
\newcommand{\ours}[0]{\textsc{David-GRPO}\xspace}
\usepackage{graphicx}
\usepackage{adjustbox}
\makeatletter
\newcommand{\thickhline}{
    \noalign {\ifnum 0=`}\fi \hrule height 1pt
    \futurelet \reserved@a \@xhline
}
\newcolumntype{"}{@{\hskip\tabcolsep\vrule width 1pt\hskip\tabcolsep}}
\makeatother

\makeatletter
\newcommand*{\blackleq}{
  \mathrel{
    \mathpalette\@blackleq{}
  }
}
\newcommand*{\@blackleq}[2]{
  \vcenter{
    \m@th
    \setbox0=\hbox{$#1\mkern3mu$}
    \setbox2=\hbox{$#1\vcenter{}$}
    \setbox4=\hbox{\raisebox{-\ht2}[.2pt][.2pt]{$#1-$}}
    \hbox{$#1\blacktriangleleft$}
    \nointerlineskip
    \kern\wd0 
    \copy4 
  }
}
\makeatother
\usepackage{multirow}
\newcommand{\edit}[1]{#1}

\definecolor{my_blue}{RGB}{0,112,192}

\usepackage{placeins}

\usepackage{tikz}

\usepackage{fvextra}

\DefineVerbatimEnvironment{WideMinted}{Verbatim}
{breaklines, fontsize=\small, frame=lines, linenos, breakanywhere, xleftmargin=0pt, xrightmargin=0pt, width=\linewidth}

\usepackage{enumitem}
\usepackage{subcaption}

\newsavebox{\figA}
\newsavebox{\figB}
\newsavebox{\figC}

\usetikzlibrary{calc}

\usepackage{listings} % 코드 및 태그 하이라이팅용

\definecolor{my_green}{rgb}{0.1, 0.5, 0.2}
\definecolor{tagblack}{RGB}{0, 0, 0}   % search 관련 태그
\definecolor{tagblue}{RGB}{0, 102, 204}   % search 관련 태그
\definecolor{tagblue}{RGB}{0, 102, 204}   % search 관련 태그
\definecolor{tagred}{RGB}{204, 0, 0}   % search 관련 태그
\definecolor{tagpurple}{RGB}{128, 0, 128} % think 관련 태그
\definecolor{goldbg}{RGB}{245, 240, 255}  % Gold Response 연보라 배경
\definecolor{goldframe}{RGB}{200, 180, 230} % Gold Response 테두리

\makeatletter
\newcommand{\lstlinecolor}{}
\lst@AddToHook{EveryPar}{\lstlinecolor}
\makeatother

\newtcolorbox{fewshotbox}[1]{
    colback=white, colframe=gray!60, fonttitle=\bfseries\large,
    title={#1}, breakable, enhanced, attach title to upper,
    after title={\medskip\hrule\medskip}, coltitle=black, colbacktitle=white,
    boxrule=0.8pt, arc=3pt, top=10pt
}

\newtcolorbox{goldbox}{
    colback=goldbg, colframe=goldpurple, fonttitle=\bfseries,
    title=Gold Response, 
    breakable,           % <--- 이 옵션이 끊김 현상을 해결합니다
    enhanced,
    attach boxed title to top left={xshift=2mm, yshift=-2mm},
    boxed title style={colback=goldpurple, colframe=goldpurple},
    coltitle=white, sharp corners=downhill, arc=2pt,
    left=3mm, right=3mm, top=5mm, bottom=3mm,
    before skip=10pt     % 프롬프트와의 간격 조절
}
\makeatletter
\ifacl@anonymize
\newcommand{\codeurl}{https://anonymous.4open.science/r/David-GRPO-5261/}

\else
\newcommand{\codeurl}{https://github.com/AsadalJung/David-GRPO}

\fi
\makeatother

\title{\textit{Can David Beat Goliath?} On Multi-Hop Reasoning with Resource-Constrained Agents}

\author{
Hojae Han\textsuperscript{1} \quad
Heeyun Jung\textsuperscript{2} \quad
Jongyoon Kim\textsuperscript{3} \quad
Seung-won Hwang\textsuperscript{3}\thanks{Corresponding author.} \\
\textsuperscript{1}Electronics and Telecommunications Research Institute (ETRI) \\
\textsuperscript{2}The University of Hong Kong \quad
\textsuperscript{3}Seoul National University \\
\texttt{hojae.han@etri.re.kr}, \texttt{heeyun.jung@connect.hku.hk}, \\
\texttt{\{jongyoon.kim,seungwonh\}@snu.ac.kr}
}

\begin{document}

\maketitle

\begin{abstract}
\edit{Reinforcement learning (RL) trains
small language model agents to answer multi-hop questions by retrieving evidence over multiple turns, but reported gains typically rely on thousands of on-policy rollouts per update.
We study RL for such agents on commodity GPUs, where each update samples only a few rollouts per question.
Most sampled trajectories then retrieve none of the required evidence, so the outcome reward gives the policy little to learn from and small agents settle for answering without retrieval, a failure we call \emph{retrieval collapse}.
\ours addresses this with two mechanisms: (1) \emph{Expert trajectory seeding} places a handful of off-policy expert trajectories into the GRPO groups of the early updates, and (2) \emph{evidence-guided continuation} rewards evidence coverage and resumes the most promising partial trajectory.
The evidence for each training question is constructed from the corpus link graph, without annotated evidence.
On six multi-hop QA benchmarks, \ours trained on four RTX 3090 GPUs with 144 rollouts per step brings Qwen2.5-1.5B to 22.6 average EM against 11.9 for the best baseline under the same budget, is comparable to Tree-GRPO trained with 20 times more rollouts, and, unlike the baselines that stop after at most one search, learns to retrieve across turns.}\footnote{\href{\codeurl}{\codeurl}}
\end{abstract}

% \footnote{\href{https://huggingface.co/spaces/davidgrpo/David-GRPO}{https://huggingface.co/spaces/davidgrpo/David-GRPO}}

% These results demonstrate that, with the right inductive biases, small language agents can occupy an overlooked but highly desirable regime: \textit{low training cost with high accuracy}.

\section{Introduction}
\label{sec:intro}

\begin{figure}[t]
  \centering
    \includegraphics[width=0.9\columnwidth]{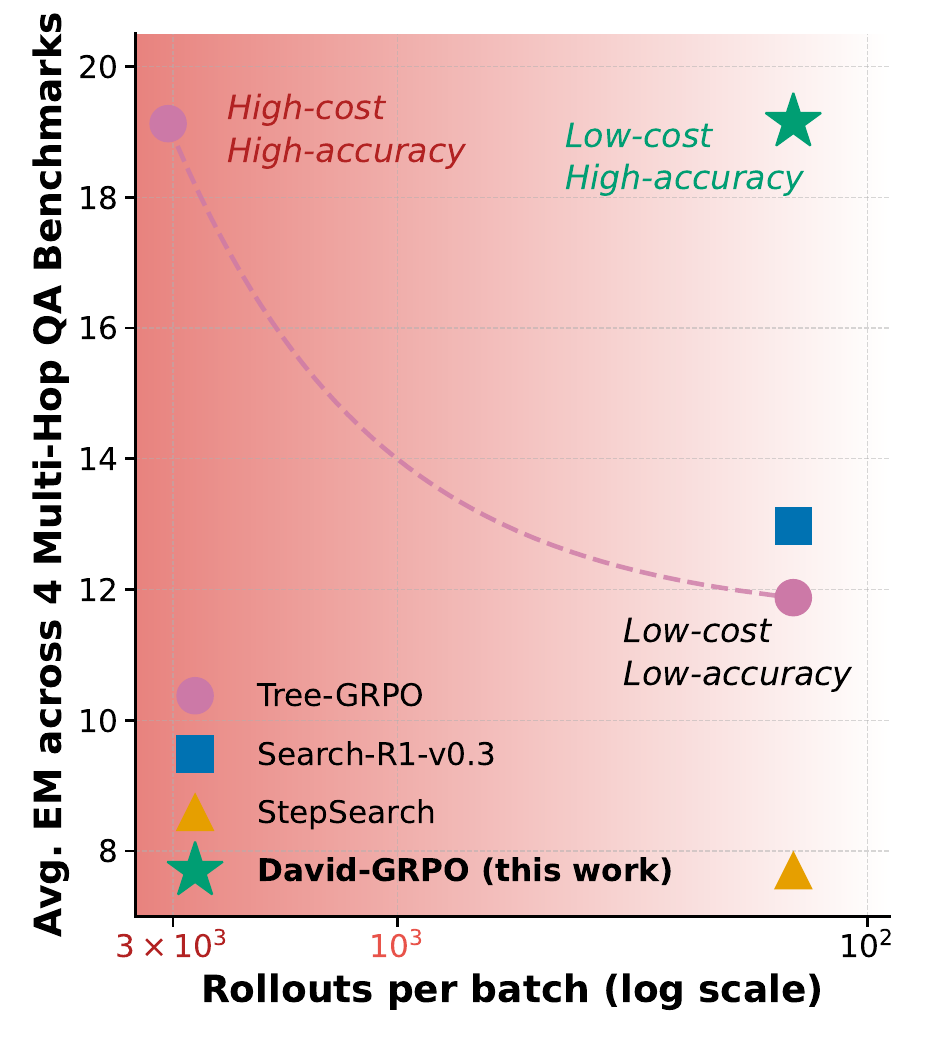}
\caption{
\edit{Average EM on four multi-hop QA benchmarks versus rollouts per batch (log scale, Qwen2.5-1.5B). Shading marks rollout intensity and the dashed line traces Tree-GRPO scaling. With 4.7\% of the high-budget rollouts, \ours reaches average EM comparable to high-budget Tree-GRPO.}
% To visualize the rollout budget, the background and x-axis tick labels are shaded from white to red, where redder indicates more rollouts per batch.
% The dashed line connects the two Tree-GRPO operating points (144 vs.\ 3{,}072 rollouts per batch) to highlight the scaling trend with increased rollout budget.
% At 144 rollouts per batch, \textbf{\ours} outperforms StepSearch, Search-R1-v0.3, and Tree-GRPO, and approaches Tree-GRPO at 3{,}072 rollouts per batch while using \textbf{only 4.7\%} of that rollout budget.
}
  \label{fig:rollout}
\end{figure}
Reinforcement learning (RL) has recently been used to train small language model agents that interleave reasoning with retrieval on long-horizon tasks such as multi-hop question answering~\citep{jin2025searchr,ji2025curriculum,zhang2025agent,ji2025treesearch,chen2025improving}.
\edit{The reported gains, however, come from high-cost training with thousands of on-policy rollouts per update and large batches (Figure~\ref{fig:rollout}), which is out of reach for most groups working on commodity hardware~\citep{kudiabor2024academics,besiroglu2024compute}.}
%We ask whether small agents can learn multi-hop reasoning when each update can afford only a few rollouts per question.}

\edit{In contrast, we study commodity GPU scenarios with a small rollout budget, where RL fails in a specific way.
GRPO-style training forms advantages from reward differences among the trajectories sampled for the same question~\citep{shao2024deepseekmath}.
In multi-hop retrieval, a trajectory is useful only if it retrieves the bridge evidence and then the answer evidence, and with few rollouts most sampled trajectories retrieve none of it.
The outcome reward then gives the policy little to learn from, and small agents settle for answering from parametric knowledge without retrieval.
We call this failure \emph{retrieval collapse}. With no rewarded retrieval in a group, GRPO gives no positive signal for multi-turn retrieval and reinforces answering without it.
Every prior low-budget agent we train undergoes retrieval collapse and converges to zero or one search per question (\S\ref{sec:analysis}).
Existing ways to add learning signal assume resources that a low rollout budget lacks. SFT initialization requires thousands of expert demonstrations~\citep{zhang2025agent}, tree expansion requires large batches to find useful prefixes~\citep{ji2025treesearch}, and retrieval-level rewards require annotated evidence for every training question~\citep{jin2025an,zheng-etal-2025-stepsearch}.}

We introduce \ours, \edit{which learns multi-turn retrieval under a low rollout budget with two mechanisms.
Expert trajectory seeding prevents retrieval collapse by placing a handful of off-policy expert trajectories into the GRPO groups of the early updates, so that each group contains rewarded multi-turn retrieval before the agent's own rollouts retrieve evidence (\S\ref{sec:warmstart}).
Evidence-guided continuation rewards the fraction of evidence a trajectory retrieves and resumes the best partial trajectory from the last successful retrieval, so that the agent extends partial retrieval chains until the evidence is complete (\S\ref{sec:evidence_guided_exploration}).
The evidence for each training question is constructed from the corpus link graph (\S\ref{sec:evidence_construction}), so no annotated evidence is required.}

\edit{On six multi-hop QA benchmarks, \ours trained on four RTX 3090 GPUs with 144 rollouts per step and seeded with four expert trajectories brings Qwen2.5-1.5B to 22.6 average EM against 11.9 for the best baseline under the same budget, and is comparable to Tree-GRPO trained with 20 times more rollouts (Table~\ref{tab:main_results}).
The improvement over the baselines holds on the Qwen2.5 and Llama backbones we test and, in a smaller experiment, on a 7B agent (\S\ref{sec:exp_results}).
Unlike the baselines, which stop after at most one search, \ours learns to retrieve across turns (\S\ref{sec:analysis}).}

\section{Related Work}
\label{sec:related_work}

% In this section, we first overview our problem space and explain three challenges while describing how existing approaches have been addressed.

% In this section, we first overview the problem space then visit each of the three key challenges in Table~\ref{tab:contrib}.

\paragraph{\edit{Search Agents Trained with RL.}}
\edit{Multi-hop QA requires an agent to retrieve several documents from a corpus and combine them~\citep{yang2018hotpotqa,ho-etal-2020-constructing,trivedi-etal-2022-musique,press2023measuring}. Retrieval serves two roles here, \textit{bridge retrieval} for intermediate entities and \textit{answer retrieval} for the final evidence~\citep{xiong-etal-2019-simple,fang-etal-2020-hierarchical,xiong2021answering}.
Recent work trains such agents with RL, mostly from outcome rewards on the final answer~\citep{jin2025searchr,zhang2025agent,chen2025improving,ji2025treesearch}, using thousands of on-policy rollouts per update.}

\paragraph{\edit{Expert Demonstrations and Rollout Expansion.}}
\edit{When an update contains few useful trajectories from the current policy, one approach distills trajectories from a stronger model through SFT before RL, as AutoCoA does with thousands of multi-turn agent trajectories~\citep{zhang2025agent}. LUFFY instead mixes an off-policy expert trace into the GRPO group of every training question~\citep{yan2025learning}.
Another is to reuse partially successful on-policy trajectories. Tree-GRPO randomly expands a reasoning tree and propagates outcome rewards to intermediate steps~\citep{ji2025treesearch}. None of these fits a low budget. SFT and LUFFY need expert trajectories for thousands of training questions, and random expansion chooses expansion points without evidence signals and so needs large batches to hit useful prefixes by chance.}

\paragraph{\edit{Retrieval Rewards and Evidence Annotation.}}
\edit{Retrieval-level rewards give denser feedback than outcome rewards. StepSearch scores each step by lexical similarity to the annotated evidence~\citep{zheng-etal-2025-stepsearch}, and Search-R1-v0.3 rewards retrieving the answer document~\citep{jin2025an}. Lexical similarity is a noisy proxy for retrieving the evidence itself, both rewards credit individual documents and can be satisfied by a single hit, and both need annotated evidence for every training question.
Existing training sets obtain this evidence from crowdworkers, who write a question over a chain of documents and mark the supporting paragraphs~\citep{yang2018hotpotqa,trivedi-etal-2022-musique}. We follow the same chain-based construction but let a language model write the questions, as in synthetic QA generation~\citep{alberti-etal-2019-synthetic}, so the chain itself serves as the evidence.}

\paragraph{Our Distinction.}
\edit{\ours needs neither large annotation sets nor large batches. It replaces the SFT stage with four expert trajectories inside early GRPO groups (\S\ref{sec:warmstart}), rewards coverage of the whole support set rather than single documents, chooses continuation points by evidence coverage rather than at random (\S\ref{sec:evidence_guided_exploration}), and obtains the evidence from the corpus graph without annotation.}

\section{Method}
\label{sec:method}
\begin{figure*}[t]
    \centering
    \captionsetup[subfloat]{margin={0pt, 20pt}}
    \def\figACaption{\makebox[0pt][l]{\hspace{-1.4cm}(a) Expert Trajectory Seeding.}}

    % subfloat들을 "한 번만" 박스로 떠둠 (카운터/label은 a->b->c 순서로 정상)
    \sbox{\figA}{%
        {\captionsetup[subfloat]{labelformat=empty}%
        \subfloat[Expert Trajectory Seeding][\figACaption\label{fig:1a}]{%
            \includegraphics[width=0.2651\linewidth]{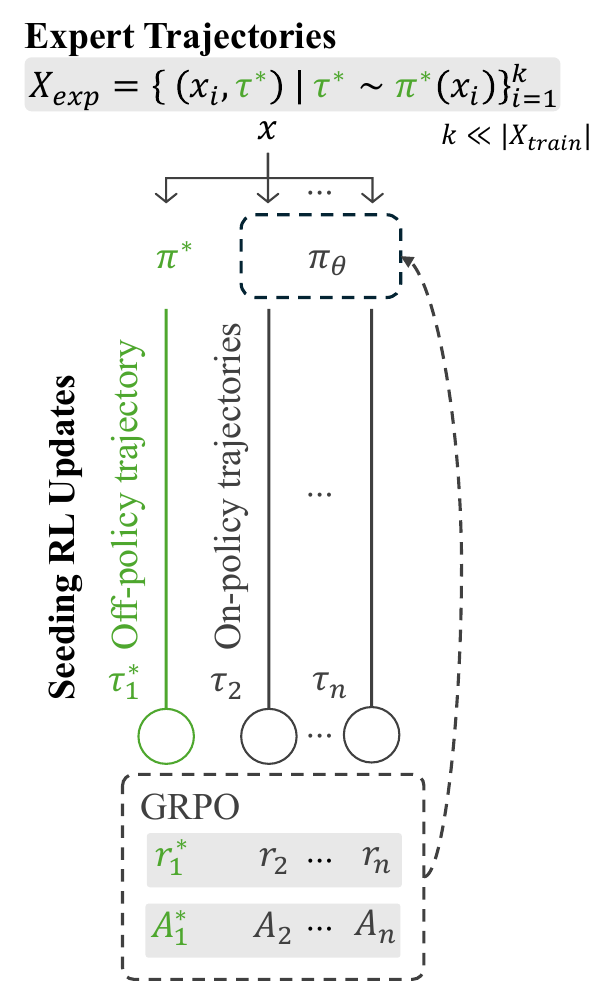}%
        }}%
    }%
    \sbox{\figB}{%
        \subfloat[Evidence-Guided Continuation.\label{fig:1b}]{%
            \includegraphics[width=0.7349\linewidth]{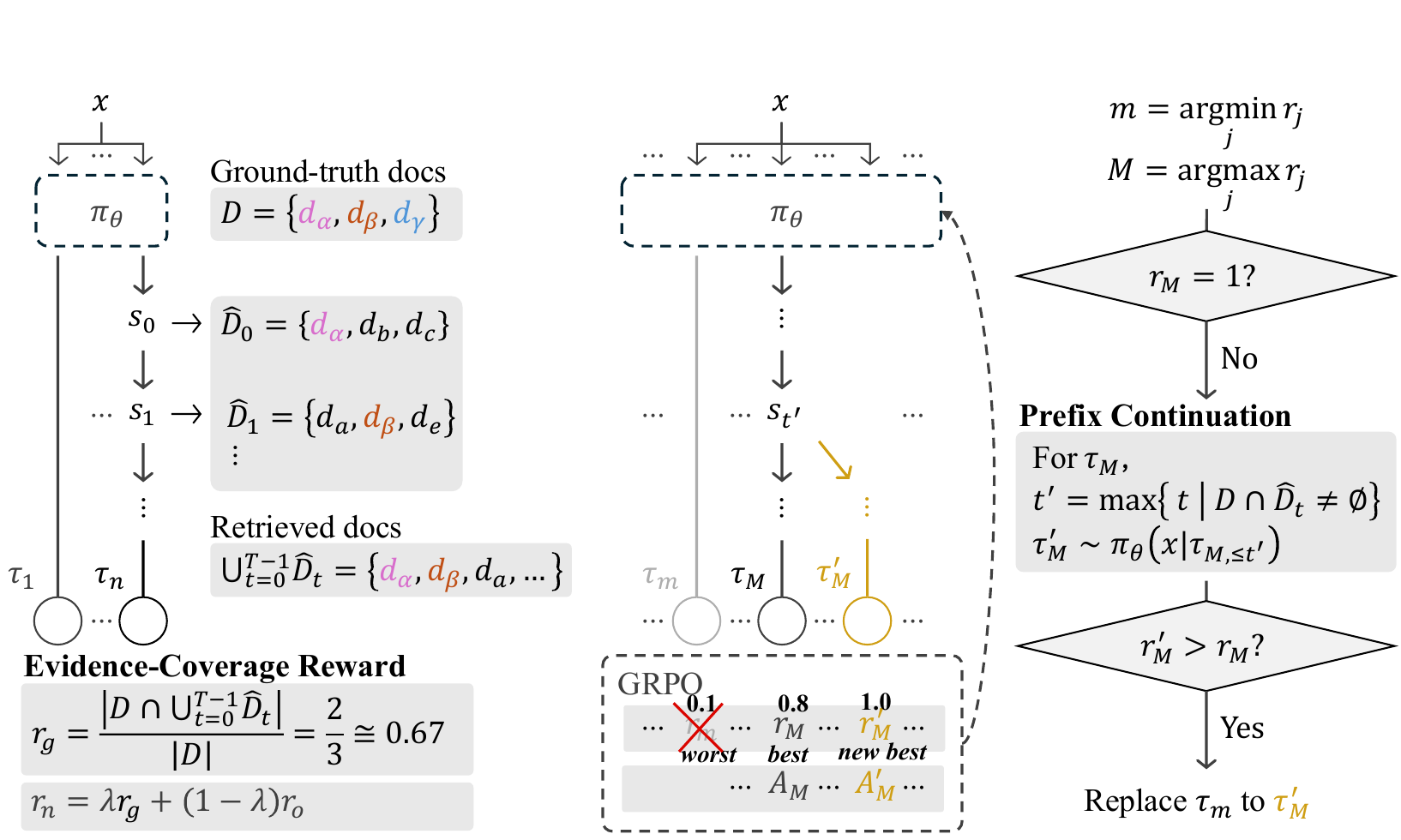}%
        }%
    }%
    % \sbox{\figC}{%
    %     \subfloat[Support-grounded continuation\label{fig:1c}]{%
    %         \raisebox{-0.8ex}{\includegraphics[width=0.4301\linewidth]{fig/fig1c.pdf}}%
    %     }%
    % }%

    \resizebox{1.0\linewidth}{!}{%
    \begin{tabular}{@{}ccc@{}}

        % (a)는 공간만 확보(안 보이게) -> 레이아웃 유지
        \phantom{\usebox{\figA}}
        &
        % (b)를 먼저 그림 (기존과 동일)
        \hspace{-0.45cm}%
        \usebox{\figB}%
        % (a)를 (b) "뒤에" 그리되, (a) 원래 자리로 되돌려서 오버레이 => (a)가 위로 올라감
        \makebox[0pt][l]{%
            \kern-\dimexpr\wd\figB+\wd\figA+2\tabcolsep-0.45cm\relax
            \usebox{\figA}%
        }%
        &
        % % (c)는 그대로
        % \hspace{-0.3cm}%
        % \usebox{\figC}
        \\

    \end{tabular}%
    }

    \caption{Overview of \ours.}
    \label{fig:overview_v2}
\end{figure*}

We first formalize the task (\S\ref{sec:formulation}), then present the two components of \ours, expert trajectory seeding (\S\ref{sec:warmstart}) and evidence-guided continuation (\S\ref{sec:evidence_guided_exploration}).
Figure~\ref{fig:overview_v2} overviews \ours with these components.
% \sw{stopped here}

\subsection{Problem Formulation}
\label{sec:formulation}
\edit{We follow the retrieval-based multi-hop QA setting, where the information needed to answer a question is available in an external corpus~\citep{yang2018hotpotqa,ho-etal-2020-constructing,trivedi-etal-2022-musique}.
A training instance $(x, y, D) \in X_{train}$ consists of a question $x$, an answer $y$, and the support set $D$, the documents that together support the answer.
\ours automatically constructs these instances from a given document corpus (\S\ref{sec:evidence_construction}).
A policy $\pi_\theta$, a language model agent with parameters $\theta$, answers $x$ over several turns. In each turn it writes reasoning text and then either issues a search query, whose retrieved documents are appended to the context, or emits the final answer.
We model this as an episodic MDP in the standard agentic RL setup (Appendix~\ref{appendix:mdp_formulation}). A step $t$ is one turn, the action $a_t$ is the query issued at step $t<T$ or the answer $\hat{y}$ at the terminal step $T$, and $\hat{D}_t$ denotes the documents retrieved at step $t$.
A rollout produces a trajectory $\tau \sim \pi_\theta(x)$ that ends when the agent answers or exhausts a budget of tokens and tool calls, and training assigns it a reward $R(\tau; y, D)$.}
\edit{The policy parameters $\theta$ are updated with GRPO~\citep{shao2024deepseekmath}. For each query $x$, GRPO samples a group of $G$ trajectories $\{\tau_1,\dots,\tau_G\}$ from the policy before the update, $\pi_{\theta_{old}}$, standardizes their rewards within the group into advantages $\hat{A}_i = (R(\tau_i) - \mathrm{mean}_j R(\tau_j)) / \mathrm{std}_j R(\tau_j)$, and updates $\pi_\theta$ by maximizing the clipped surrogate $f_{clip}(\rho_i(\theta), \hat{A}_i)$ with probability ratio $\rho_i(\theta){=} \pi_\theta(\tau_i)/\pi_{\theta_{old}}(\tau_i)$, averaged over the group and regularized by a KL term $\beta\,\mathbb{D}_{KL}$ against a reference model (Eq.~\eqref{eq:grpo} in Appendix~\ref{appendix:mdp_formulation}).}

\subsection{Expert Trajectory Seeding}
\label{sec:warmstart}
\edit{With few rollouts per question, standard GRPO rarely samples a rewarded trajectory and undergoes retrieval collapse (the no-seeding baseline in \S\ref{sec:analysis}).}
Unlike SFT-based initialization~\citep{zhang2025agent}, which uses thousands of expert trajectories, expert trajectory seeding uses a handful of expert trajectories inside RL updates.
Formally, the expert trajectory set $X_{\text{exp}}{=}\{(x_j, \tau^*_j) \mid \tau^*_j \sim \pi^*(x_j)\}_{j=1}^k$ holds $k$ trajectories generated by an expert policy $\pi^*$ (e.g., a larger model or a human).
\edit{In our experiments, $X_{\text{exp}}$ holds four HotpotQA trajectories released by AutoCoA (Appendix~\ref{appendix:fewshot_examples}), the only annotated examples that \ours uses and 99.9\% fewer (4 vs. 5,276) than the SFT stage of AutoCoA.}
Appendix~\ref{appendix:fewshot_warm_start} reports how performance changes with larger $k$.
\edit{Seeding forms the early phase of training, in which every group is built from a question in $X_{\text{exp}}$. Training then continues on $X_{train}$ with on-policy groups only (50 and 215 steps in our experiments, Appendix~\ref{appendix:implmentation}).}
As illustrated in Figure~\ref{fig:1a}, for an input $x$ with $(x, \tau^*) \in X_{\text{exp}}$ we seed a group of size $G$ with one off-policy expert trajectory $\tau^*$ and fill it with $G-1$ on-policy trajectories $\{\tau_2, \dots, \tau_G\}$ sampled from $\pi_{\theta_{old}}$.
To account for the distributional shift, we write the expert trajectory ratio as a product of the standard policy ratio and an additional expert-policy correction: $\rho_{\mathrm{mix}}^*(\theta)=\omega^*(\theta) \cdot \frac{\pi_\theta(\tau^*)}{\pi_{\theta_{old}}(\tau^*)}$
where $\omega^*(\theta)$ accounts for the gap between the expert policy and the rollout policy.
\edit{The expert trajectory receives the same reward $R(\tau^*; y, D)$ as the on-policy trajectories, and the advantage $\hat{A}(\tau^*)$ is computed by standardizing rewards within the mixed group, as in Appendix~\ref{appendix:mdp_formulation}.}
Applying the same clipped surrogate function $f_{clip}$ and KL penalty $\mathbb{D}_{KL}$ as in Eq.~\ref{eq:grpo}, the mixed objective is:
\begin{equation}
\label{eq:hybrid_rl}
\begin{split}
    \mathcal{L}_{Mixed}(\theta) = & \frac{1}{G} \Big( f_{clip}(\rho_{\mathrm{mix}}^*(\theta), \hat{A}(\tau^*)) + \\
    & \sum_{i=2}^{G} f_{clip}(\rho_i(\theta), \hat{A}(\tau_i)) \Big) - \beta \mathbb{D}_{KL}.
\end{split}
\end{equation}
\edit{The expert trajectory gives a gradient signal even when the agent's own exploration fails, and the on-policy trajectories let a high-reward trajectory of the agent itself receive a positive $\hat{A}_i$ beyond imitating $\tau^*$. Figure~\ref{fig:warm-start} shows that this outperforms SFT on $X_{\text{exp}}$ and pure off-policy RL.}
In implementation, we keep the standard policy ratio $\pi_\theta(\tau^*)/\pi_{\theta_{old}}(\tau^*)$ for $\tau^*$ but set the additional expert-policy correction $\omega^*(\theta)$ to $1$. Appendix~\ref{appendix:fewshot_warm_start} compares this choice with trajectory- and action-level corrections.

\subsection{Evidence-Guided Continuation}
\label{sec:evidence_guided_exploration}
\edit{Evidence-guided continuation scores each trajectory by evidence coverage (\S\ref{sec:grounded_reward}) and uses the score to resume the most promising partial trajectory when a group has no fully rewarded trajectory (\S\ref{sec:continuation}).}

% \subsection{Grounded Multi-Hop Retrieval Reward }
\subsubsection{Evidence-Coverage Reward}
\label{sec:grounded_reward}

\edit{Expert trajectory seeding provides trajectories with non-zero reward early in training, but with the outcome reward alone small agents still fall into retrieval collapse (see Tree-GRPO in Table~\ref{tab:retrieval_behavior}). A denser reward should score whether retrieval found the evidence. Aligning each retrieval step to a particular document would need step-level annotation, so we score the evidence that a trajectory has accumulated as a whole.}
To this end, we introduce an \textit{evidence-coverage reward} ($r_g$). As illustrated in Figure~\ref{fig:1b}, let $\hat{\mathcal{D}}_{union}$ be the set of all documents retrieved in the trajectory, the union over intermediate steps ($t < T$):
\begin{equation}
\label{eq:union}
    \hat{\mathcal{D}}_{union} = \bigcup_{t=1}^{T-1} \hat{D}_t.
\end{equation}
The evidence-coverage reward $r_g$ is the recall of the support set $D$ within this set:
\begin{equation}
    r_g(\tau) = \frac{| \hat{\mathcal{D}}_{union} \cap D |}{| D |}.
\end{equation}

\edit{Recall does not penalize an unsuccessful search while required evidence is still missing, and Appendix~\ref{appendix:grounded_retrieval_reward} compares this choice with an F1-oriented reward.}

We combine $r_g$ with the outcome reward $r_o$ (which evaluates the predicted answer $\hat{y}$ against $y$). The total reward is a weighted sum:
\begin{equation}
    R(\tau) = \lambda r_g(\tau) + (1-\lambda) r_o(\tau, y)
\end{equation}
where $\lambda$ is a balancing hyperparameter.
We set $\lambda{=}0.5$ in the main experiments and report sensitivity in Appendix~\ref{appendix:grounded_retrieval_reward}.

% sampling a trajectory which retrieves every necessary evidence and returns the correct answer, i.e., .

% Preamble에 \usepackage[table]{xcolor} 가 필요합니다.

% --- 색상 정의 ---
\definecolor{pennyColor}{HTML}{8B5A2B} 
\definecolor{abl1Color}{HTML}{B07A3C}
\definecolor{abl2Color}{HTML}{D2A679}
\definecolor{cleanColor}{HTML}{DCEFE6} % Antileak-m: post-cutoff (contamination-free) evaluation

\begin{table*}[t]
\scriptsize
\centering
\renewcommand{\arraystretch}{1.1} 
\setlength{\tabcolsep}{3pt} 

\begin{adjustbox}{max width=\textwidth}
\begin{tabular}{cl cc cc cc cc cc >{\columncolor{cleanColor}}c >{\columncolor{cleanColor}}c cc c}
\toprule
\multicolumn{2}{c}{\multirow{2}{*}{\textbf{Method}}} & \multicolumn{2}{c}{\textbf{HotpotQA}} & \multicolumn{2}{c}{\textbf{2wiki}} & \multicolumn{2}{c}{\textbf{Musique}} & \multicolumn{2}{c}{\textbf{Bamboogle}} & \multicolumn{2}{c}{\textbf{Bamtwoogle}} & \multicolumn{2}{c}{\cellcolor{cleanColor}\textbf{Antileak-m}} & \multicolumn{3}{c}{\textbf{Avg.}} \\
\cmidrule(lr){3-4} \cmidrule(lr){5-6} \cmidrule(lr){7-8} \cmidrule(lr){9-10} \cmidrule(lr){11-12} \cmidrule(lr){13-14} \cmidrule(lr){15-17}
 & & \textbf{EM} & \textbf{F1} & \textbf{EM} & \textbf{F1} & \textbf{EM} & \textbf{F1} & \textbf{EM} & \textbf{F1} & \textbf{EM} & \textbf{F1} & \textbf{EM} & \textbf{F1} & \textbf{EM} & \textbf{F1} & \textbf{Search} \\
\midrule

% --- Model 1: Qwen2.5-1.5B ---
\multirow{16}{*}{\rotatebox[origin=c]{90}{\textbf{Qwen2.5-1.5B}}} 
& \multicolumn{16}{c}{\textbf{\textsc{No Training}}} \\ \cmidrule{2-17}
& Direct Inference$^\dagger$ & \phantom{0}5.9 & - & \phantom{0}4.3 & - & \phantom{0}2.6 & -  & \phantom{0}8.0 & - & - & - & - & - & - & - & - \\ \cmidrule{2-17}

& \multicolumn{16}{c}{\textbf{\textsc{High Training Budget}}: up to batch size 512 $\times$ 6 rollout per instance $= 3,072$} \\ \cmidrule{2-17}
& Search-o1$^\dagger$ & 11.6 & - & 12.2 & - & \phantom{0}3.1 & - & 13.0 & - & - & - & - & - & - & - & - \\
& GRPO$^\dagger$ & 14.6  & - & 24.4 & - & \phantom{0}2.2 & - & \phantom{0}4.0 & - & - & - & - & - & - & - & \edit{0.00\rlap{$^\ddagger$}} \\
& Tree-GRPO$^\dagger$ & 29.5  & - & 26.8 & - & \phantom{0}6.6 & - & 13.6 & - & - & - & - & - & - & - & \edit{1.00\rlap{$^\ddagger$}} \\ \cmidrule{2-17}

& \multicolumn{16}{c}{\textbf{\textsc{Low Training Budget}}: up to batch size 24 $\times$ 6 rollout per instance $= 144$} \\ \cmidrule{2-17}
& Tree-GRPO & 12.9 & 18.6 & 20.5 & 23.3 & \phantom{0}2.1 & \phantom{0}7.2 & \underline{12.0} & \underline{15.5} & \phantom{0}\underline{5.0} & \phantom{0}\underline{8.2} & 11.7 & 15.7 & 10.7 & 14.8 & 0.00 \\
& StepSearch & 11.9 & 18.1 & 13.5 & 18.1 & \phantom{0}2.2 & \phantom{0}6.9 & \phantom{0}3.2 & \phantom{0}8.7 & \phantom{0}4.0 & \phantom{0}6.0 & 12.1 & 17.3 & \phantom{0}7.8 & 12.5 & \underline{1.11} \\
& Search-R1-v0.3 \textit{w/ retrieval reward} & \underline{19.0} & \underline{26.5} & \underline{21.3} & \underline{26.4} & \phantom{0}\underline{3.6} & \phantom{0}\underline{9.2} & \phantom{0}8.0 & 14.3 & \phantom{0}3.0 & \phantom{0}5.5 & \underline{16.7} & \underline{22.2} & \underline{11.9} & \underline{17.4} & 0.45 \\
\rowcolor{pennyColor!50} \cellcolor{white}  & \textbf{\ours} & \textbf{23.5} & \textbf{32.2} & \textbf{26.7} & \textbf{32.0} & \phantom{0}\textbf{\color{red}8.0} & \textbf{14.0} & \textbf{\color{red}18.4} & \textbf{28.1} & \textbf{22.0} & \textbf{26.2} & \textbf{36.9} & \textbf{41.4} & \textbf{22.6} & \textbf{29.0} & \textbf{3.13} \\
\midrule

% --- Model 2: Llama-3.2-1B ---
& Tree-GRPO & 12.4 & 18.4 & \textbf{20.5} & \textbf{23.4} & \phantom{0}1.6 & \phantom{0}\underline{7.1} & \phantom{0}\underline{4.0} & \phantom{0}8.9 & \phantom{0}\textbf{9.0} & \textbf{11.9} & 16.0 & 20.7 & \underline{10.6} & 15.1 & 0.00 \\
& StepSearch & \underline{16.0} & \textbf{24.5} & 11.7 & 16.2 & \phantom{0}\underline{2.7} & \phantom{0}\textbf{9.7} & \phantom{0}\underline{4.0} & \underline{11.0} & \phantom{0}\underline{3.0} & \phantom{0}5.7 & \underline{22.4} & \textbf{31.4} & 10.0 & \underline{16.4} & \underline{1.14} \\
& Search-R1-v0.3 \textit{w/ retrieval reward} & \phantom{0}8.5 & 12.4 & \underline{13.5} & 16.2 & \phantom{0}0.7 & \phantom{0}3.7 & \phantom{0}0.8 & \phantom{0}1.6 & \phantom{0}2.0 & \phantom{0}4.7 & 13.0 & 15.3 & \phantom{0}6.4 & \phantom{0}9.0 & 0.09 \\
\rowcolor{pennyColor!50} \cellcolor{white} \multirow{-4}{*}{\rotatebox[origin=c]{90}{\tiny\textbf{Llama-3.2-1B}}} & \textbf{\ours} & \textbf{16.7} & \underline{24.0} & 12.6 & \underline{17.5} & \phantom{0}\textbf{2.9} & \phantom{0}6.9 & \phantom{0}\textbf{8.8} & \textbf{16.0} & \phantom{0}\underline{3.0} & \phantom{0}\underline{6.3} & \textbf{24.4} & \underline{31.3} & \textbf{11.4} & \textbf{17.0} & \textbf{1.77} \\
\midrule

% --- Model 3: Qwen2.5-0.5B ---
& Tree-GRPO & \phantom{0}\underline{9.2} & \underline{12.8} & \textbf{20.9} & \underline{22.8} & \phantom{0}\underline{0.7} & \phantom{0}\underline{3.8} & \phantom{0}\underline{3.2} & \phantom{0}\underline{5.2} & \phantom{0}\underline{4.0} & \phantom{0}\underline{5.5} & \underline{10.3} & \underline{12.7} & \phantom{0}\underline{8.1} & \underline{10.5} & 0.00 \\
& StepSearch & \phantom{0}2.1 & \phantom{0}4.5 & \phantom{0}4.3 & \phantom{0}7.3 & \phantom{0}0.2 & \phantom{0}1.8 & \phantom{0}1.6 & \phantom{0}2.1 & \phantom{0}0.0 & \phantom{0}0.6 & \phantom{0}0.2 & \phantom{0}0.8 & \phantom{0}1.4 & \phantom{0}2.9 & \underline{0.89} \\
& Search-R1-v0.3 \textit{w/ retrieval reward} & \phantom{0}0.0 & \phantom{0}0.0 & \phantom{0}0.0 & \phantom{0}0.0 & \phantom{0}0.0 & \phantom{0}0.0 & \phantom{0}0.0 & \phantom{0}0.0 & \phantom{0}0.0 & \phantom{0}0.0 & \phantom{0}0.0 & \phantom{0}0.0 & \phantom{0}0.0 & \phantom{0}0.0 & 0.00 \\
\rowcolor{pennyColor!50} \cellcolor{white} \multirow{-4}{*}{\rotatebox[origin=c]{90}{\tiny\textbf{Qwen2.5-0.5B}}} & \textbf{\ours} & \textbf{13.1} & \textbf{19.2} & \underline{20.5} & \textbf{24.5} & \phantom{0}\textbf{2.3} & \phantom{0}\textbf{6.0} & \phantom{0}\textbf{4.0} & \phantom{0}\textbf{8.0} & \phantom{0}\textbf{5.0} & \phantom{0}\textbf{6.3} & \textbf{16.0} & \textbf{22.7} & \textbf{10.2} & \textbf{14.4} & \textbf{2.60} \\
\bottomrule
\end{tabular}
\end{adjustbox}
\caption{EM and F1 on six multi-hop QA benchmarks. Best results are \textbf{bolded} and second-best \underline{underlined}. \textcolor{red}{\textbf{Red}} marks low-budget EM above the best high-budget baseline. $^\dagger$ denotes results from \citet{ji2025treesearch}. {\ours} rows are brown. \edit{\textit{Search} is the average number of search actions per question over the six benchmarks. $^\ddagger$ is the final number of valid searches per rollout in the released training logs of \citet{ji2025treesearch} (0.998 for Tree-GRPO, 0.0 for GRPO). \colorbox{cleanColor}{Teal} Antileak-m cells are contamination-free because the questions post-date the knowledge cutoff of the backbone. Table~\ref{tab:gold_d_ablation} reports \ours trained with annotated evidence.}}
\label{tab:main_results}
\end{table*}

\subsubsection{Prefix Continuation}
\label{sec:continuation}
\edit{With few rollouts, a group often contains no trajectory that reaches the maximum reward ($R(\tau)<1$ for every $\tau$ in the group), so the advantages $\hat{A}_i$ carry little signal. Even after expert trajectory seeding, such groups still make up 20 to 33\% of a batch during early training (Appendix~\ref{appendix:grounded_expansion_dynamics}).}
% Even with the dense signal from $r_g$, simultaneously achieving perfect retrieval and correct reasoning remains a rare event under limited number of rollouts.
% Under limited number of rollouts, a sampled group may consist entirely of suboptimal trajectories ($R(\tau)<1$ for all $\tau \sim \pi_\theta$), limiting the effectiveness of the relative advantage estimation in GRPO.
% Even after our proposed warm-starts, around 20 to 30\% of batch examples fall short in this case (see Appendix~\ref{appendix:analysis}).
\edit{We introduce \textit{prefix continuation}, which keeps the prefix of the group's highest-coverage trajectory up to the last retrieval that added evidence and resamples the rest of the trajectory on policy.}

As illustrated in Figure~\ref{fig:1b}, within a sampled group of $G$ trajectories, we identify the highest-reward trajectory $\tau_M$ and the lowest-reward trajectory $\tau_m$. If even $\tau_M$ does not reach the maximum reward (i.e., $R(\tau_M) < 1$), we hypothesize that
% while the agent successfully located partial evidence, it failed to retrieve
the remaining necessary information is not recalled, known as \textit{bounded recall}~\citep{10.1145/2959100.2959190} in IR.
% to fully resolve the query.
For such \textit{near-miss} groups, we locate the last step $t'$ at which coverage increased:
\begin{equation}
    t' = \max \{ t \mid r_g(\tau_{M,\leq t}) > r_g(\tau_{M,\leq t-1}) \}
\end{equation}
where $\tau_{M,\le t}$ is a truncated trajectory of $\tau_M$ from step 1 to $t$, and $r_g$ is the evidence-coverage reward.
We then truncate $\tau_M$ at step $t'$, keeping the history up to that retrieval, and resample the rest of the trajectory to obtain a new candidate $\tau'_M \sim \pi_\theta(x | \tau_{M,\le t'})$.

\edit{If the continuation earns a higher reward than the original trajectory ($R(\tau'_M) > R(\tau_M)$), we replace $\tau_m$ with $\tau'_M$, which keeps the group size fixed.}

\begin{table*}[t]
\centering
\scriptsize
\setlength{\tabcolsep}{2.5pt}
\begin{adjustbox}{max width=\textwidth}
\begin{tabular}{l cc cc cc cc cc >{\columncolor{cleanColor}}c >{\columncolor{cleanColor}}c cc c}
\toprule
\multirow{2.5}{*}{\edit{\textbf{Method}}}
& \multicolumn{2}{c}{\edit{\textbf{HotpotQA}}}
& \multicolumn{2}{c}{\edit{\textbf{2wiki}}}
& \multicolumn{2}{c}{\edit{\textbf{Musique}}}
& \multicolumn{2}{c}{\edit{\textbf{Bamboogle}}}
& \multicolumn{2}{c}{\edit{\textbf{Bamtwoogle}}}
& \multicolumn{2}{c}{\cellcolor{cleanColor}\edit{\textbf{Antileak-m}}}
& \multicolumn{3}{c}{\edit{\textbf{Avg.}}} \\
\cmidrule(lr){2-3} \cmidrule(lr){4-5} \cmidrule(lr){6-7} \cmidrule(lr){8-9} \cmidrule(lr){10-11} \cmidrule(lr){12-13} \cmidrule(lr){14-16}
& \edit{\textbf{EM}} & \edit{\textbf{F1}}
& \edit{\textbf{EM}} & \edit{\textbf{F1}}
& \edit{\textbf{EM}} & \edit{\textbf{F1}}
& \edit{\textbf{EM}} & \edit{\textbf{F1}}
& \edit{\textbf{EM}} & \edit{\textbf{F1}}
& \edit{\textbf{EM}} & \edit{\textbf{F1}}
& \edit{\textbf{EM}} & \edit{\textbf{F1}} & \edit{\textbf{Search}} \\
\midrule
\multicolumn{16}{c}{\edit{\textbf{\textsc{No Training}}}} \\ \midrule
\edit{Direct Inference$^\dagger$} & \edit{18.3} & - & \edit{25.0} & - & \phantom{0}\edit{3.1} & - & \edit{12.0} & - & - & - & - & - & - & - & \edit{-} \\ \midrule
\multicolumn{16}{c}{\edit{\textbf{\textsc{High Training Budget}} (reported)}} \\ \midrule
\edit{Search-o1$^\dagger$} & \edit{22.1} & - & \edit{21.8} & - & \phantom{0}\edit{5.4} & - & \edit{32.0} & - & - & - & - & - & - & - & \edit{-} \\
\edit{GRPO$^\dagger$} & \edit{42.5} & - & \edit{40.7} & - & \edit{19.1} & - & \edit{43.2} & - & - & - & - & - & - & - & \edit{-} \\
\edit{Tree-GRPO$^\dagger$} & \edit{44.6} & - & \edit{42.3} & - & \edit{20.2} & - & \edit{44.0} & - & - & - & - & - & - & - & \edit{1.82\rlap{$^\P$}} \\
\edit{StepSearch$^\ddagger$} & \edit{38.0} & \edit{49.3} & \edit{38.5} & \edit{45.0} & \edit{21.6} & \edit{32.4} & \edit{46.7} & \edit{57.3} & - & - & - & - & - & - & \edit{-} \\
\edit{Search-R1-v0.3 \textit{w/ retrieval reward}$^\S$} & \edit{41.6} & - & \edit{37.5} & - & \edit{16.4} & - & \edit{38.7} & - & - & - & - & - & - & - & \edit{-} \\ \midrule
\multicolumn{16}{c}{\edit{\textbf{\textsc{Low Training Budget}}: up to batch size 24 $\times$ 6 rollout per instance $= 144$}} \\ \midrule
\edit{Tree-GRPO} & \textbf{35.5} & \textbf{46.4} & 28.8 & 35.0 & \textbf{15.3} & \textbf{23.5} & \underline{40.0} & \underline{51.1} & \underline{38.0} & \textbf{46.5} & \underline{43.5} & 52.6 & \underline{33.5} & \underline{42.5} & \edit{1.92} \\
\edit{StepSearch} & 27.5 & 37.7 & 18.2 & 25.4 & 12.9 & 21.5 & 31.2 & 44.0 & 36.0 & 40.7 & 43.3 & \underline{53.0} & 28.2 & 37.1 & \edit{\underline{2.40}} \\
\edit{Search-R1-v0.3 \textit{w/ retrieval reward}} & 34.9 & 45.5 & \underline{35.5} & \underline{41.3} & \underline{13.0} & 21.5 & 34.4 & 45.7 & 36.0 & 42.5 & 42.4 & 47.1 & 32.7 & 40.6 & \edit{1.73} \\
\rowcolor{pennyColor!50}
\edit{\textbf{\ours}} & \underline{35.2} & \underline{45.7} & \textbf{38.7} & \textbf{45.8} & \textbf{15.3} & \underline{23.2} & \textbf{43.2} & \textbf{52.6} & \textbf{42.0} & \underline{45.6} & \textbf{49.2} & \textbf{54.7} & \textbf{37.3} & \textbf{44.6} & \edit{\textbf{3.15}} \\
\bottomrule
\end{tabular}
\end{adjustbox}
\caption{\edit{{Scaling to Qwen2.5-7B.} High-budget rows are scores reported by $^\dagger$\citet{ji2025treesearch} (batch size 512), $^\ddagger$\citet{zheng-etal-2025-stepsearch} (7B-Base, batch size 256, 16$\times$H800), and $^\S$\citet{jin2025an} (7B-Base GRPO with retrieval reward, batch size 512, 8$\times$H100). Best and second-best low-budget results are \textbf{bolded} and \underline{underlined}. \textit{Search}: searches per question over the six benchmarks ($^\P$on HotpotQA with the released best checkpoint).}}
\label{tab:qwen7b_results}
\end{table*}
\section{Experiments}
\label{sec:experiment}

\subsection{Experimental Setup}
\label{sec:exp_setup}
\paragraph{\edit{Backbones.}}
\edit{We train agents with at most 1.5B parameters, \textbf{Qwen2.5} (0.5B, 1.5B) \citep{qwen2025qwen25technicalreport} and \textbf{Llama-3.2-1B} \citep{grattafiori2024llama}, and additionally \textbf{Qwen2.5-7B} under the same budget to check whether the gains hold at a larger scale.
We use these backbones deliberately: their knowledge cutoff precedes Antileak-m, so Antileak-m measures retrieval rather than memorization (Appendix~\ref{appendix:cutoff}). Newer ones such as \textbf{Qwen3-0.6B}~\citep{yang2025qwen3} do not, so we report it separately (Appendix~\ref{appendix:qwen3}).
We train the Qwen models from their base checkpoints, except for Tree-GRPO, which we train from the instruct versions because it degenerates from the base checkpoints (Appendix~\ref{appendix:tree_grpo_controlled}). For Llama-3.2-1B we use the instruct version, because the base checkpoint ships without a chat template and cannot be trained with the agent prompt.}

\paragraph{\edit{Training Data.}}
\label{sec:evidence_construction}
\edit{\ours trains on 4{,}000 instances constructed automatically from the Wikipedia corpus released with HotpotQA~\citep{yang2018hotpotqa}. For the analysis in \S\ref{sec:analysis}, we also train \ours on 5{,}276 HotpotQA training examples, with the support set $D$ taken from the supporting documents annotated in HotpotQA.
\ours needs the support set $D$ of each training question, so we construct it from the corpus. We sample a document chain from the hyperlink graph, let Qwen2.5-3B write a multi-hop question backward along the chain, and keep the instance only if the chain is necessary and sufficient to answer the question (Appendix~\ref{appendix:evidence_annotation_requirement}). The sampled chain is the support set $D$, and 45\% of the instances require three or more hops. Four expert trajectories from AutoCoA form the expert trajectory set (\S\ref{sec:warmstart}). Baselines use their released training configurations (Appendix~\ref{appendix:implmentation}).}

\paragraph{Training Budget.}
\edit{We train on 4$\times$NVIDIA RTX 3090 GPUs (24GB VRAM) with a global batch size of 24 and 6 rollouts per example on average,\footnote{\ours samples five full rollouts per example, and prefix continuation adds 1.1 rollouts per example on average that resume from a prefix of an existing trajectory (Appendix~\ref{appendix:implmentation}).} 144 rollouts per optimization step, and call this setting the \textit{Low Training Budget}.
The \textit{High Training Budget} of Tree-GRPO uses 8$\times$H20 GPUs (96GB VRAM) and 3{,}072 rollouts per step ($512 \times 6$), so the low budget is 4.7\% of the high budget. The reported 7B scores of StepSearch and Search-R1-v0.3 in Table~\ref{tab:qwen7b_results} come from batch sizes of 256 and 512.
Qwen2.5-7B does not fit on the RTX 3090s, so we train it on 8$\times$NVIDIA RTX A6000 GPUs (48GB VRAM) with the same batch size and rollout budget as the smaller models.
Appendix~\ref{appendix:computation_analysis} reports training cost and inference-time token counts, and Appendix~\ref{appendix:implmentation} the remaining implementation details.}

\paragraph{Baselines.}
\edit{We compare with three RL methods for search agents under the same low budget. \textbf{Tree-GRPO} \citep{ji2025treesearch} is the only method that reports 1.5B results under a high budget, so we also compare with the reported scores. \textbf{Search-R1-v0.3} \citep{jin2025an} trains with a \textit{retrieval reward}, and \textbf{StepSearch} \citep{zheng-etal-2025-stepsearch} with step-wise lexical similarity rewards (\S\ref{sec:related_work}). Appendix~\ref{analysis:full_sft} compares with \textbf{AutoCoA} \citep{zhang2025agent}, which initializes from full SFT on a large trajectory set, and Appendix~\ref{appendix:no_training} reports no-training controls of the base model.}

\paragraph{Benchmarks and Metrics.}
\edit{We evaluate on six multi-hop QA benchmarks. \textbf{HotpotQA} \citep{yang2018hotpotqa} covers \textit{bridge} and \textit{comparison} questions with two hops. \textbf{2WikiMultiHopQA} \citep{ho-etal-2020-constructing} and \textbf{MuSiQue} \citep{trivedi-etal-2022-musique} add compositional questions, up to four hops in MuSiQue. \textbf{Bamboogle} \citep{press2023measuring} and \textbf{BamTwoogle} \citep{aksitov2024rest} are handcrafted benchmarks with 2-hop and 2- to 4-hop questions. \textbf{AntiLeakBench-Multihop} \citep{wu-etal-2025-antileakbench} (2024 subset, 2 to 3 hops), which we call Antileak-m, contains questions that post-date the knowledge cutoff of the backbones (Appendix~\ref{appendix:benchmark_statistics}).
We report \textbf{Exact Match (EM)} and \textbf{F1}, and Appendix~\ref{appendix:llm_judge} reports LLM-as-judge evaluation. Appendix~\ref{appendix:bfcl} additionally evaluates \ours on the multi-turn long-context subset of BFCL, a tool-use benchmark outside Wikipedia QA.}

\begin{table*}[t]
\centering
\setlength{\tabcolsep}{3pt}
\scriptsize
\begin{adjustbox}{max width=\textwidth}
\begin{tabular}{l cccccc cccccc}
\toprule
\multirow{2.5}{*}{\textbf{Method}} & \multicolumn{6}{c}{\textbf{HotpotQA}} & \multicolumn{6}{c}{\textbf{MuSiQue}} \\
\cmidrule(lr){2-7} \cmidrule(lr){8-13}
 & \textbf{Search} & \textbf{Prec.} & \textbf{Any} & \textbf{All} & \textbf{Unnec.} & \textbf{EM} & \textbf{Search} & \textbf{Prec.} & \textbf{Any} & \textbf{All} & \textbf{Unnec.} & \textbf{EM} \\
\midrule
Tree-GRPO & 0.00 & - & \phantom{0}0.0 & \phantom{0}0.0 & - & 12.9 & 0.00 & - & \phantom{0}0.0 & 0.0 & - & 2.1 \\
StepSearch & 1.12 & 39.6 & 40.3 & 13.8 & 1.6 & 11.9 & 1.11 & 17.2 & 14.7 & 1.8 & 0.2 & 2.2 \\
Search-R1-v0.3 & 0.53 & \textbf{40.1} & 20.9 & \phantom{0}6.7 & 0.0 & 19.0 & 0.59 & \textbf{18.3} & \phantom{0}8.7 & 1.3 & 0.0 & 3.6 \\
\rowcolor{pennyColor!50}
\textbf{\ours} & 3.05 & 28.2 & \textbf{52.3} & \textbf{31.6} & 9.8 & \textbf{24.8} & 3.48 & 14.3 & \textbf{25.9} & \textbf{8.2} & 2.6 & \textbf{7.1} \\
\bottomrule
\end{tabular}
\end{adjustbox}
\caption{\edit{{Retrieval behavior.} Retrieval statistics and EM (\%) on HotpotQA and MuSiQue with Qwen2.5-1.5B (Qwen2.5-1.5B-Inst for Tree-GRPO). \textit{Search}: searches per question. \textit{Prec.}: \% of retrieved documents in the annotated support set $D$. \textit{Any}: \% of documents of $D$ retrieved, pooled over questions. \textit{All}: \% of questions with every document of $D$ retrieved. \textit{Unnec.}: \% of searches issued after all of $D$ is retrieved.}}
\label{tab:retrieval_behavior}
\end{table*}
\begin{table*}[t]
\scriptsize
\centering
% \renewcommand{\arraystretch}{1.2}
% 컬럼이 많아져서 tabcolsep을 조금 더 줄이거나, 필요시 resizebox 고려
\setlength{\tabcolsep}{3.0pt} 

% 마지막 두 열(Antileak-m)에 회색 배경 적용
\begin{adjustbox}{max width=\textwidth}
\begin{tabular}{l cc cc cc cc cc >{\columncolor{cleanColor}}c >{\columncolor{cleanColor}}c cc}
\toprule
\multirow{2.5}{*}{\textbf{Method}} 
& \multicolumn{2}{c}{\scriptsize\textbf{HotpotQA}} 
& \multicolumn{2}{c}{\scriptsize\textbf{2wiki}} 
& \multicolumn{2}{c}{\scriptsize\textbf{Musique}} 
& \multicolumn{2}{c}{\scriptsize\textbf{Bamboogle}} 
& \multicolumn{2}{c}{\scriptsize\textbf{Bamtwoogle}} 
& \multicolumn{2}{c}{\cellcolor{cleanColor}\scriptsize\textbf{Antileak-m}} 
& \multicolumn{2}{c}{\scriptsize\textbf{Avg.}} \\ % 헤더 배경색 명시
\cmidrule(lr){2-3} \cmidrule(lr){4-5} \cmidrule(lr){6-7} \cmidrule(lr){8-9} \cmidrule(lr){10-11} \cmidrule(lr){12-13} \cmidrule(lr){14-15}
 & \scriptsize \textbf{EM} & \scriptsize \textbf{F1} 
 & \scriptsize\textbf{EM} & \scriptsize\textbf{F1} 
 & \scriptsize\textbf{EM} & \scriptsize\textbf{F1} 
 & \scriptsize\textbf{EM} & \scriptsize\textbf{F1} 
 & \scriptsize\textbf{EM} & \scriptsize\textbf{F1} 
 & \scriptsize\textbf{EM} & \scriptsize\textbf{F1}
 & \scriptsize\textbf{EM} & \scriptsize\textbf{F1} \\
\midrule

% --- Ours ---
\rowcolor{pennyColor!50}
\textbf{\ours}
& \textbf{24.8} & \textbf{33.8}
& \textbf{27.2} & \textbf{32.3}
& \phantom{0}\textbf{7.1} & \textbf{12.6}
& \textbf{14.4} & \textbf{24.2} 
& \textbf{22.0} & \textbf{25.4} 
& \textbf{36.3} & \textbf{41.1}
& \textbf{22.0} & \textbf{28.2} \\

% --- Ablation 1 ---
\rowcolor{abl1Color!45}
\textit{\phantom{0}w/o continuation}
& \underline{23.3} & \underline{31.9}
& \underline{24.4} & \underline{30.0}
& \phantom{0}\underline{5.8} & \underline{11.3}
& \textbf{14.4} & \underline{20.9}
& \underline{19.0} & \underline{23.8}
& \underline{33.2} & \underline{38.2}
& \underline{20.0} & \underline{26.0} \\
% & \underline{23.8} & \underline{32.6}
% & \underline{24.6} & \underline{30.3}
% & \phantom{0}\underline{6.0} & \underline{11.6}
% & \textbf{14.4} & \textbf{25.8} 
% & \underline{14.0} & \underline{18.5} 
% & \underline{33.2} & \underline{38.2} \\

% --- Ablation 2 ---
\rowcolor{abl2Color!30}
\textit{\phantom{00}w/o coverage reward}
& 20.8 & 28.9
& 22.0 & 27.0
& \phantom{0}4.5 & 10.1
& 10.4 & 19.9 
& 10.0 & 12.7 
& 31.4 & 36.0
& 16.5 & 22.4 \\

\bottomrule
\end{tabular}
\end{adjustbox}
\caption{{Ablation on evidence-guided continuation.} Performance of \ours and its ablations on Qwen2.5-1.5B across multi-hop QA benchmarks. \edit{The \textit{w/o coverage reward} variant removes both components and is trained with only the outcome reward.}}
\label{tab:ablation_grounding_components}
\end{table*}

\subsection{Experimental Results}
\label{sec:exp_results}
\paragraph{\edit{Comparison with High-Budget Training.}}
\edit{Table~\ref{tab:main_results} first compares \ours with the results reported for Qwen2.5-1.5B under the high training budget of Tree-GRPO, 3{,}072 rollouts per step on 8$\times$H20 GPUs. With \textbf{4.7\%} of those rollouts on four RTX 3090s, \ours averages \textbf{19.2 EM against 19.1} for high-budget Tree-GRPO on the four benchmarks it reports, exceeding it on MuSiQue and Bamboogle (\textcolor{red}{red}), trailing on HotpotQA (23.5 vs. 29.5), and tying on 2WikiMultiHopQA (26.7 vs. 26.8). High-budget vanilla GRPO and Search-o1 average 11.3 and 10.0 EM on the same benchmarks, so the low-budget \ours also exceeds them by 8 to 9 points (Figure~\ref{fig:rollout}).}

\paragraph{\edit{Low-Budget Comparison.}}
\edit{Under the low budget, \ours reaches \textbf{22.6 average EM and 29.0 F1} with Qwen2.5-1.5B against 11.9 and 17.4 for the best baseline, Search-R1-v0.3, and leads on every benchmark. The margin is largest on the contamination-free Antileak-m, 20.2 EM points. Moving from 3{,}072 to 144 rollouts costs Tree-GRPO EM on all four benchmarks it reports, and GRPO without seeding earns no reward at all (\S\ref{sec:ablation}). \ours also has the highest average EM on Qwen2.5-0.5B and Llama-3.2-1B, where Search-R1-v0.3 collapses to 0 EM at 0.5B. On the more recent Qwen3-0.6B, StepSearch ties \ours at 19 average EM while Tree-GRPO and Search-R1-v0.3 stay below 10 (Appendix~\ref{appendix:qwen3}).}

\begin{figure}[t]
  \centering
\includegraphics[width=\columnwidth]{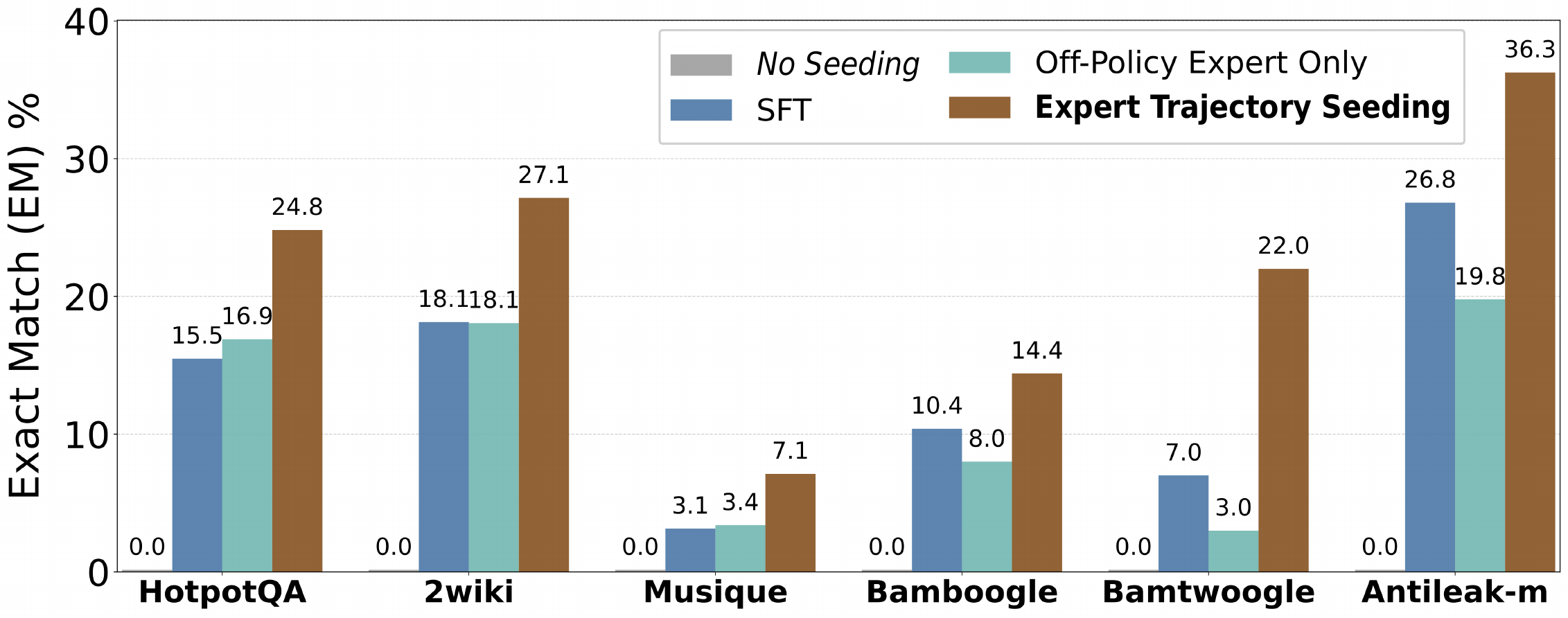}
  \caption{
    {Analysis on expert trajectory seeding.} Performance comparison of different ways to use the four expert trajectories of $X_{\text{exp}}$ followed by GRPO with the \textit{evidence-coverage reward} on Qwen2.5-1.5B.
  }
  \label{fig:warm-start}
\end{figure}
\paragraph{\edit{Multi-Turn Retrieval Under the Low Budget.}}
\edit{The Search column of Table~\ref{tab:main_results} shows that \ours issues the most searches per question on every backbone. With Qwen2.5-1.5B, \ours averages 3.1 searches over the six benchmarks, whereas Tree-GRPO, Search-R1-v0.3, and StepSearch average 0.0, 0.5, and 1.1. On Llama-3.2-1B and Qwen2.5-0.5B, \ours averages 1.8 and 2.6 searches, and no baseline exceeds 1.1. Even 3{,}072 rollouts per step do not keep an outcome-reward agent retrieving. The released training log of high-budget GRPO ends with 0.0 valid searches per rollout, and that of high-budget Tree-GRPO with 1.0.}

\paragraph{\edit{Scaling to a 7B Agent.}}
\edit{With Qwen2.5-7B under the same budget (Table~\ref{tab:qwen7b_results}), \ours has the highest average EM and F1 (37.3 and 44.6 vs. 33.5 and 42.5 for Tree-GRPO), the largest margin on Antileak-m (49.2 vs. 43.5), and the most searches per question (3.2 vs. 1.7 to 2.4 over the six benchmarks). Even the high-budget Tree-GRPO checkpoint searches less. On HotpotQA, the released 7B checkpoint averages 1.8 searches per question, and \ours averages 3.0 right after expert trajectory seeding, before the main GRPO stage (Appendix~\ref{appendix:tree_grpo_controlled}). Against the reported high-budget scores, \ours reaches 88\% of Tree-GRPO and 91\% of StepSearch on the four shared benchmarks with 4.7\% of the rollouts.}

\begin{table*}[t]
\centering
\scriptsize
\setlength{\tabcolsep}{3pt}
\begin{adjustbox}{max width=\textwidth}
\begin{tabular}{l l cc cc cc cc cc >{\columncolor{cleanColor}}c >{\columncolor{cleanColor}}c cc c}
\toprule
\multirow{2.5}{*}{\edit{\textbf{Backbone}}} & \multirow{2.5}{*}{\edit{\textbf{Evidence}}}
& \multicolumn{2}{c}{\edit{\textbf{HotpotQA}}}
& \multicolumn{2}{c}{\edit{\textbf{2wiki}}}
& \multicolumn{2}{c}{\edit{\textbf{Musique}}}
& \multicolumn{2}{c}{\edit{\textbf{Bamboogle}}}
& \multicolumn{2}{c}{\edit{\textbf{Bamtwoogle}}}
& \multicolumn{2}{c}{\cellcolor{cleanColor}\edit{\textbf{Antileak-m}}}
& \multicolumn{3}{c}{\edit{\textbf{Avg.}}} \\
\cmidrule(lr){3-4} \cmidrule(lr){5-6} \cmidrule(lr){7-8} \cmidrule(lr){9-10} \cmidrule(lr){11-12} \cmidrule(lr){13-14} \cmidrule(lr){15-17}
& & \edit{\textbf{EM}} & \edit{\textbf{F1}}
& \edit{\textbf{EM}} & \edit{\textbf{F1}}
& \edit{\textbf{EM}} & \edit{\textbf{F1}}
& \edit{\textbf{EM}} & \edit{\textbf{F1}}
& \edit{\textbf{EM}} & \edit{\textbf{F1}}
& \edit{\textbf{EM}} & \edit{\textbf{F1}}
& \edit{\textbf{EM}} & \edit{\textbf{F1}} & \edit{\textbf{Search}} \\
\midrule
 & \edit{Gold} & \textbf{35.2} & 45.6 & 37.7 & 44.5 & 14.2 & 21.6 & 40.0 & 52.3 & \textbf{44.0} & \textbf{47.1} & \textbf{55.2} & \textbf{59.8} & \textbf{37.7} & \textbf{45.1} & \edit{\textbf{3.95}} \\
\rowcolor{pennyColor!50} \cellcolor{white} \multirow{-2}{*}{\edit{Qwen2.5-7B}} & \edit{Synth} & \textbf{35.2} & \textbf{45.7} & \textbf{38.7} & \textbf{45.8} & \textbf{15.3} & \textbf{23.2} & \textbf{43.2} & \textbf{52.6} & 42.0 & 45.6 & 49.2 & 54.7 & 37.3 & 44.6 & \edit{3.15} \\
\midrule
 & \edit{Gold} & \textbf{24.8} & \textbf{33.8} & \textbf{27.2} & \textbf{32.3} & \phantom{0}7.1 & 12.6 & 14.4 & 24.2 & \textbf{22.0} & 25.4 & 36.3 & 41.1 & 22.0 & 28.2 & \edit{3.07} \\
\rowcolor{pennyColor!50} \cellcolor{white} \multirow{-2}{*}{\edit{Qwen2.5-1.5B}} & \edit{Synth} & 23.5 & 32.2 & 26.7 & 32.0 & \textbf{\phantom{0}8.0} & \textbf{14.0} & \textbf{18.4} & \textbf{28.1} & \textbf{22.0} & \textbf{26.2} & \textbf{36.9} & \textbf{41.4} & \textbf{22.6} & \textbf{29.0} & \edit{\textbf{3.13}} \\
\midrule
 & \edit{Gold} & \textbf{17.7} & \textbf{25.2} & \textbf{16.1} & \textbf{21.4} & \textbf{\phantom{0}3.2} & \textbf{\phantom{0}8.5} & \phantom{0}8.0 & 14.5 & \textbf{\phantom{0}3.0} & \textbf{\phantom{0}6.6} & 23.3 & \textbf{31.7} & \textbf{11.9} & \textbf{18.0} & \edit{1.30} \\
\rowcolor{pennyColor!50} \cellcolor{white} \multirow{-2}{*}{\edit{Llama-3.2-1B}} & \edit{Synth} & 16.7 & 24.0 & 12.6 & 17.5 & \phantom{0}2.9 & \phantom{0}6.9 & \textbf{\phantom{0}8.8} & \textbf{16.0} & \textbf{\phantom{0}3.0} & \phantom{0}6.3 & \textbf{24.4} & 31.3 & 11.4 & 17.0 & \edit{\textbf{1.77}} \\
\midrule
 & \edit{Gold} & 10.8 & 16.0 & 17.4 & 20.8 & \phantom{0}2.0 & \phantom{0}5.3 & \textbf{\phantom{0}4.8} & \textbf{\phantom{0}8.1} & \textbf{\phantom{0}6.0} & \textbf{\phantom{0}7.0} & 10.6 & 14.4 & \phantom{0}8.6 & 11.9 & \edit{\textbf{2.64}} \\
\rowcolor{pennyColor!50} \cellcolor{white} \multirow{-2}{*}{\edit{Qwen2.5-0.5B}} & \edit{Synth} & \textbf{13.1} & \textbf{19.2} & \textbf{20.5} & \textbf{24.5} & \textbf{\phantom{0}2.3} & \textbf{\phantom{0}6.0} & \phantom{0}4.0 & \phantom{0}8.0 & \phantom{0}5.0 & \phantom{0}6.3 & \textbf{16.0} & \textbf{22.7} & \textbf{10.2} & \textbf{14.4} & \edit{2.60} \\
\bottomrule
\end{tabular}
\end{adjustbox}
\caption{\edit{{Gold versus constructed evidence.} EM and F1 of \ours trained on the 5{,}276 HotpotQA training examples with annotated evidence (\textit{gold}) and on the 4{,}000 instances constructed in \S\ref{sec:evidence_construction} (\textit{synth}, the default \ours). The higher value of each pair is \textbf{bolded}. \textit{Search} is the average number of search actions per question over the six benchmarks. \colorbox{cleanColor}{Teal} Antileak-m cells are contamination-free evaluations.}}
\label{tab:gold_d_ablation}
\end{table*}

\section{Analysis}
\label{sec:analysis}

% \begin{figure}[t]
%   \centering
% \includegraphics[width=\linewidth]{fig/ablation_benchmarks_em_davidgrpo_v8_italic_wo.pdf}
%   \caption{
%     Ablation study.
%   }
%   \label{fig:ablation}
% \end{figure}

% \input{table/warmup}

\edit{This section separates the two sources of the gains. In \S\ref{sec:retrieval_capabilities} and \S\ref{sec:ablation}, \ours is trained on the HotpotQA training set with its annotated evidence (gold), the same kind of supervision the retrieval-reward baselines use, so the comparisons isolate the RL mechanisms. \S\ref{sec:evidence_supervision_analysis} then keeps the mechanisms fixed and changes only the training data.}

\subsection{Multi-Hop Retrieval Capabilities}
\label{sec:retrieval_capabilities}
% To verify whether our small language agents have genuinely acquired multi-hop retrieval skills, we analyze their behavior on benchmarks that require distinct reasoning structures.

\paragraph{\edit{Retrieval Behavior.}}
\edit{Table~\ref{tab:retrieval_behavior} compares how the agents search on HotpotQA and MuSiQue. \ours issues about three searches per question and reaches the highest document recall, both for any support document (\textit{Any}) and for the complete support set (\textit{All}). Tree-GRPO stops searching entirely and Search-R1-v0.3 averages half a search per question, so both remain in retrieval collapse, and StepSearch searches once. The additional searches of \ours mostly target missing evidence. Only 9.8\% on HotpotQA and 2.6\% on MuSiQue are issued after the support set is already complete, and document precision stays within 12 points of the best baseline. Appendix~\ref{appendix:retrieval_behavior} breaks recall down into bridge and answer documents, and Appendix~\ref{appendix:musique_depth_intervention_analysis} reports HotpotQA question types and MuSiQue hop depths.}

\subsection{Ablation Study}
\label{sec:ablation}

\paragraph{Impact of Expert Trajectory Seeding.}
Figure~\ref{fig:warm-start} compares different ways to use the same four expert trajectories. \edit{Without them, the agent undergoes retrieval collapse and scores zero EM on every benchmark.}
Expert trajectory seeding, which combines off-policy trajectories from $X_{\text{exp}}$ (\S\ref{sec:warmstart}) with on-policy generations, achieves the highest performance.
When using $X_{\text{exp}}$ only, standard SFT performs comparably to or better than off-policy RL.
Adding on-policy rollouts gives a larger gain than either SFT or off-policy RL alone, which supports expert trajectory seeding.
\edit{Keeping the same four trajectories but replacing the rest of \ours with vanilla GRPO, StepSearch, Search-R1-v0.3, or Tree-GRPO gives at most 20.4 average EM against 22.0 (Appendix~\ref{appendix:expert_bootstrapped_rl_variants}), so the trajectories alone do not explain the gain.}

% \paragraph{Effectiveness of Grounding Modules.}
% We further dissect the impact of our grounding components in Table~\ref{tab:ablation_grounding_components}.
% The full \ours system demonstrates superior performance, ranking first in all six benchmarks.
% Removing the \textit{grounded expansion} module results in a performance drop, placing it second in most cases. %, except for one specific benchmark.
% Crucially, further removing the \textit{grounded retrieval reward}—leaving the agent without any grounding signals—leads to the lowest performance across the board.
% This step-wise degradation confirms that both the expansion mechanism and the retrieval-aware reward are essential for guiding the agent toward effective reasoning paths.

\paragraph{Effectiveness of Evidence-Guided Continuation.} Table~\ref{tab:ablation_grounding_components} isolates the evidence-guided continuation components. The full \ours system gives the best results across all six benchmarks. Removing \textit{prefix continuation} reduces performance, and removing both \textit{prefix continuation} and \textit{evidence-coverage reward} gives the lowest results.
\edit{The variant without both is trained with the outcome reward alone, so the lowest results on all six benchmarks indicate that the outcome reward is insufficient under the low budget.}
\edit{Appendix~\ref{appendix:retrieval_behavior} shows that the two components raise document recall cumulatively (Table~\ref{tab:ablation_retrieval_behavior}).}

Appendix~\ref{appendix:grounded_retrieval_reward} compares our evidence-coverage reward with StepSearch and Search-R1-v0.3 rewards, which do not explicitly reward coverage of the support set.

% We break down the MuSiQue performance according to the required number of reasoning hops, annotated in the dataset.
% Referring to Figure~\ref{fig:musique-em-hop}, \ours consistently ranks first or second regardless of reasoning depth.
% While Search-R1-v0.3 shows slightly superior EM scores on the 3- and 4-hop subsets, Figure~\ref{fig:musique-retrieval-hop} reveals a critical behavioral difference: similar to the HotpotQA results, Search-R1-v0.3 averages only a single retrieval action per question even for complex queries (see Appendix~\ref{appendix:case_study_search_r1}).
% This implies that its performance on higher-hop questions stems primarily from parametric memorization or shortcuts, whereas \ours faithfully executes the necessary sequence of retrieval steps to derive the answer.

\subsection{\edit{Gold versus Constructed Evidence}}
\label{sec:evidence_supervision_analysis}
\edit{Table~\ref{tab:gold_d_ablation} compares \ours trained on the HotpotQA training set with annotated evidence (gold) and on the constructed instances (synth). Average EM differs by at most 1.6 points on any backbone, and the direction of the difference depends on the backbone. The learned retrieval behavior is also similar. The two variants differ by at most 0.5 searches per question on every backbone except Qwen2.5-7B, where gold issues 0.8 more searches per question. Per benchmark, synth is at most 1.3 EM points below gold on HotpotQA, the training distribution of gold, and leads on MuSiQue and Bamboogle on the two largest backbones.}

% \subsection{Warm-Start with Extensive Annotations}
% \label{analysis:full_sft}
% Table~\ref{tab:full_sft_comparison_complete}

\section{Conclusion}
\edit{We studied RL for multi-hop retrieval agents on commodity GPUs, where few rollouts per question leave small agents in retrieval collapse. \ours restores the learning signal with expert trajectory seeding, which places a handful of expert trajectories in early GRPO groups, and evidence-guided continuation, which rewards coverage of evidence constructed from the corpus graph and resumes the most promising partial trajectory. With 144 rollouts per step on four RTX 3090 GPUs, \ours nearly doubles the average EM of the best low-budget baseline at 1.5B, is comparable to Tree-GRPO trained with 20 times more rollouts, and learns multi-turn retrieval where the baselines stop after at most one search.}

% \section{Conclusion}
% In this work, we presented \ours, a budget-aware reinforcement learning framework designed to empower small language models with robust multi-turn reasoning capabilities. By integrating a mixed off-/on-policy warm-start strategy, trajectory-level grounded retrieval rewards, and grounded expansion for efficient exploration, \ours effectively mitigates the cold-start and sparsity challenges inherent in training small agents under strict compute constraints. Our extensive experiments across six multi-hop QA benchmarks demonstrate that \ours consistently outperforms existing baselines designed for high-resource settings, achieving superior accuracy while utilizing significantly fewer rollouts.
% These findings challenge the assumption that multi-hop reasoning is exclusive to resource-rich \textit{Goliaths}, demonstrating that small language model agents can achieve this capability via efficient RL strategies.

% These findings challenge the prevailing assumption that effective reasoning is exclusive to large-scale computations (\textit{Goliath}), highlighting the potential of specialized, resource-efficient RL strategies to democratize intelligent agents on commodity hardware.

\section*{Limitations}
\edit{\ours targets small agents trained under a low rollout budget, the regime in which retrieval collapse arises. We train agents of up to 1.5B parameters and check the trend with Qwen2.5-7B; under high rollout budgets, where on-policy sampling already yields rewarded trajectories, the gain from seeding and continuation may be smaller. Part of the high-budget comparison relies on scores reported in the original papers, since reproducing them requires the original hardware. Expert trajectory seeding assumes a few expert trajectories (four in our experiments), and evidence construction assumes a corpus with link structure such as Wikipedia. Our evaluation covers English multi-hop QA, with a tool-use check on BFCL (Appendix~\ref{appendix:bfcl}); other languages and tool environments are left for future work.}

\section*{Ethics Statement}
This work studies how to train retrieval and tool-using agents with smaller models and commodity GPUs. Lowering the compute barrier can make agent research easier to reproduce and extend outside large industrial labs. The same techniques could also improve agents used for harmful information gathering or automated misuse if deployed without safeguards. The constructed training data is derived from Wikipedia articles and contains no personal or private information beyond the source articles. Deployment of such agents requires the usual access control, monitoring, and tool-use safeguards for agentic systems.

% acl.sty sets \bibliographystyle{acl_natbib}.
\bibliography{custom}

\appendix
% Let more full-width tables share a page so they do not pile up before \onecolumn.
\setcounter{dbltopnumber}{4}
\renewcommand{\dbltopfraction}{0.9}
\renewcommand{\dblfloatpagefraction}{0.7}

\section{\edit{Full MDP Formulation}}
\label{appendix:mdp_formulation}

\edit{This section provides the full specification of the MDP summarized in \S\ref{sec:formulation}.
We conceptualize the reasoning process of $\pi_\theta$ as a Markov decision process (MDP) $\langle S, A, P, R \rangle$,\footnote{In this setting, the discount factor is set to $1$, as the tasks are episodic and we are primarily concerned with the final correctness of the terminal state.} where $S$ and $A$ are sets of states and actions, respectively.
Let $a \oplus b$ signify the concatenation of strings $a$ and $b$.
The state transition is formulated as $s_{t+1} = s_t\oplus a_t \oplus \hat{D}_t$, where $a_t$ for $t<T$ is an intermediate query for retrieval, $\hat{D}_t$ is the set of retrieved documents queried by $a_t$,\footnote{$\hat{D}_t$ is converted into the textual sequence of documents when concatenated.} and the terminal action emits the predicted answer $\hat{y}$.\footnote{$\hat{D}_T{=}\varnothing$ as no retrieval is conducted in step $T$.}
The environment dynamics are deterministic, where the next state is formed by concatenating the chosen action to the current sequence, i.e., $P(s_{t+1}|s_{t},a_t)=1$.
For each rollout during on-policy RL and inference time, $\pi_\theta$ produces a trajectory $\tau = s_0\oplus \cdots \oplus s_T$ given by a user question $x$ and the task prompt $s_0$, i.e., $\tau\sim \pi_\theta(x)$.
During training, given a training instance $(x, y, D) \in X_{train}$, a reward $R(\tau;y,D)$ is calculated by the reward function $R$ where $y$ and $D$ are respectively the answer and the evidence to be retrieved.
The policy parameters $\theta$ are then updated with the GRPO objective. Defining the probability ratio $\rho_i(\theta) = \frac{\pi_\theta(\tau_i)}{\pi_{\theta_{old}}(\tau_i)}$, the clipped surrogate function $f_{clip}(\rho, A) = \min ( \rho A, \text{clip}( \rho, 1-\epsilon, 1+\epsilon ) A )$ with a clipping parameter $\epsilon$, and the KL divergence term against the reference model $\pi_{ref}$ as $\mathbb{D}_{KL}$, the objective is:
\begin{equation}
\label{eq:grpo}
\begin{split}
    \mathcal{L}_{GRPO}(\theta) = {} & \mathbb{E} \bigg[ \frac{1}{G} \sum_{i=1}^{G} f_{clip}(\rho_i(\theta), \hat{A}_i) \bigg] \\
    & - \beta \mathbb{D}_{KL}.
\end{split}
\end{equation}}

\section{Further Analysis}
\label{appendix:analysis}
\edit{Unless noted otherwise, the analyses in this appendix use the gold variant of \ours, as in \S\ref{sec:analysis}.}

\subsection{Additional Computation Analysis}
\label{appendix:computation_analysis}
\begin{table}[t]
\centering
\renewcommand{\arraystretch}{1.15}
\setlength{\tabcolsep}{5pt}
\scriptsize
\begin{adjustbox}{max width=\columnwidth}
\begin{tabular}{l c}
\toprule
\textbf{Method} & \textbf{GPU-days (RTX 3090)} \\
\midrule
\rowcolor{pennyColor!50}
\textbf{\ours} & \textbf{9.96} \\
\textit{\phantom{0}w/o continuation} & 8.07 \\
\textit{\phantom{00}w/o coverage reward} & 8.74 \\
\bottomrule
\end{tabular}
\end{adjustbox}
\caption{Training cost measured on RTX 3090 GPUs with a fixed peak memory budget of 20,890 MiB per GPU and no CPU offload.}
\label{tab:training_efficiency}
% \vspace{-5.0em}
\end{table}

Table~\ref{tab:training_efficiency} reports the training cost of \ours and its ablations in GPU-days measured on RTX 3090 GPUs.
Under a fixed peak memory budget of 20,890 MiB per GPU and no CPU offload, \ours takes 9.96 GPU-days, compared with 8.07 without prefix continuation and 8.74 without evidence-coverage reward.
On four RTX 3090 GPUs, this adds less than 0.5 wall-clock days over the ablations.

\begin{table*}[t]
\centering
\renewcommand{\arraystretch}{1.15}
\setlength{\tabcolsep}{4pt}
\begin{adjustbox}{max width=\textwidth}
\scriptsize
\begin{tabular}{l c c c c c >{\columncolor{cleanColor}}c c}
\toprule
\textbf{Method} & \textbf{HotpotQA} & \textbf{2wiki} & \textbf{Musique} & \textbf{Bamboogle} & \textbf{Bamtwoogle} & \cellcolor{cleanColor}\textbf{Antileak-m} & \textbf{Avg.} \\
\midrule
\rowcolor{pennyColor!50}
\textbf{\ours} & \textbf{1468.1} & \textbf{1595.6} & \textbf{1647.2} & \textbf{1466.4} & \textbf{1537.1} & \textbf{1160.8} & \textbf{1479.2} \\
\textit{\phantom{0}w/o continuation} & 1253.1 & 1443.4 & 1441.4 & 1229.6 & 1406.1 & 1074.8 & 1308.1 \\
\textit{\phantom{00}w/o coverage reward} & 1212.5 & 1456.0 & 1353.2 & 1108.9 & 1418.8 & 1139.6 & 1281.5 \\
Search-R1-v0.3 & 245.6 & 177.6 & 270.0 & 323.9 & 234.2 & 267.6 & 253.2 \\
StepSearch & 550.6 & 591.1 & 529.3 & 534.1 & 554.7 & 468.7 & 538.1 \\
Tree-GRPO & 0.0 & 0.0 & 0.0 & 0.0 & 0.0 & 0.0 & 0.0 \\
\bottomrule
\end{tabular}
\end{adjustbox}
\caption{Average number of retrieved tokens at inference time with Qwen2.5-1.5B.}
\label{tab:inference_retrieved_tokens}
\end{table*}

\begin{table*}[t]
\centering
\renewcommand{\arraystretch}{1.15}
\setlength{\tabcolsep}{4pt}
\begin{adjustbox}{max width=\textwidth}
\scriptsize
\begin{tabular}{l c c c c c >{\columncolor{cleanColor}}c c}
\toprule
\textbf{Method} & \textbf{HotpotQA} & \textbf{2wiki} & \textbf{Musique} & \textbf{Bamboogle} & \textbf{Bamtwoogle} & \cellcolor{cleanColor}\textbf{Antileak-m} & \textbf{Avg.} \\
\midrule
\rowcolor{pennyColor!50}
\textbf{\ours} & \textbf{479.7} & 449.0 & \textbf{537.9} & \textbf{455.2} & 509.9 & 383.4 & 469.2 \\
\textit{\phantom{0}w/o continuation} & 454.0 & 433.4 & 532.4 & 440.7 & 511.6 & 368.6 & 456.8 \\
\textit{\phantom{00}w/o coverage reward} & 462.3 & \textbf{465.9} & 527.6 & 433.1 & \textbf{530.9} & \textbf{441.8} & \textbf{476.9} \\
Search-R1-v0.3 & 325.2 & 412.2 & 307.2 & 223.6 & 371.4 & 256.6 & 316.0 \\
StepSearch & 126.7 & 126.0 & 127.1 & 109.7 & 174.6 & 117.9 & 130.3 \\
Tree-GRPO & 94.5 & 87.7 & 105.9 & 85.2 & 98.3 & 73.3 & 90.8 \\
\bottomrule
\end{tabular}
\end{adjustbox}
\caption{Average number of output tokens generated by Qwen2.5-1.5B at inference time.}
\label{tab:inference_output_tokens}
\end{table*}

Tables~\ref{tab:inference_retrieved_tokens} and~\ref{tab:inference_output_tokens} report inference-time retrieved and output token counts.
\ours uses more retrieved tokens than the baselines because it performs more retrieval actions.
The lower token counts of the baselines reflect their retrieval behavior. Tree-GRPO skips retrieval, and the other baselines stop after shallow retrieval (Figures~\ref{fig:hotpot-retrieval-hop} and~\ref{fig:musique-retrieval-hop}).

\subsection{\edit{HotpotQA Question Types and MuSiQue Reasoning Depth}}
\label{appendix:musique_depth_intervention_analysis}
\begin{figure*}[t]
  \centering

  % (a) Exact Match
  \begin{subfigure}{0.49\linewidth}
    \centering
    \includegraphics[width=\linewidth]{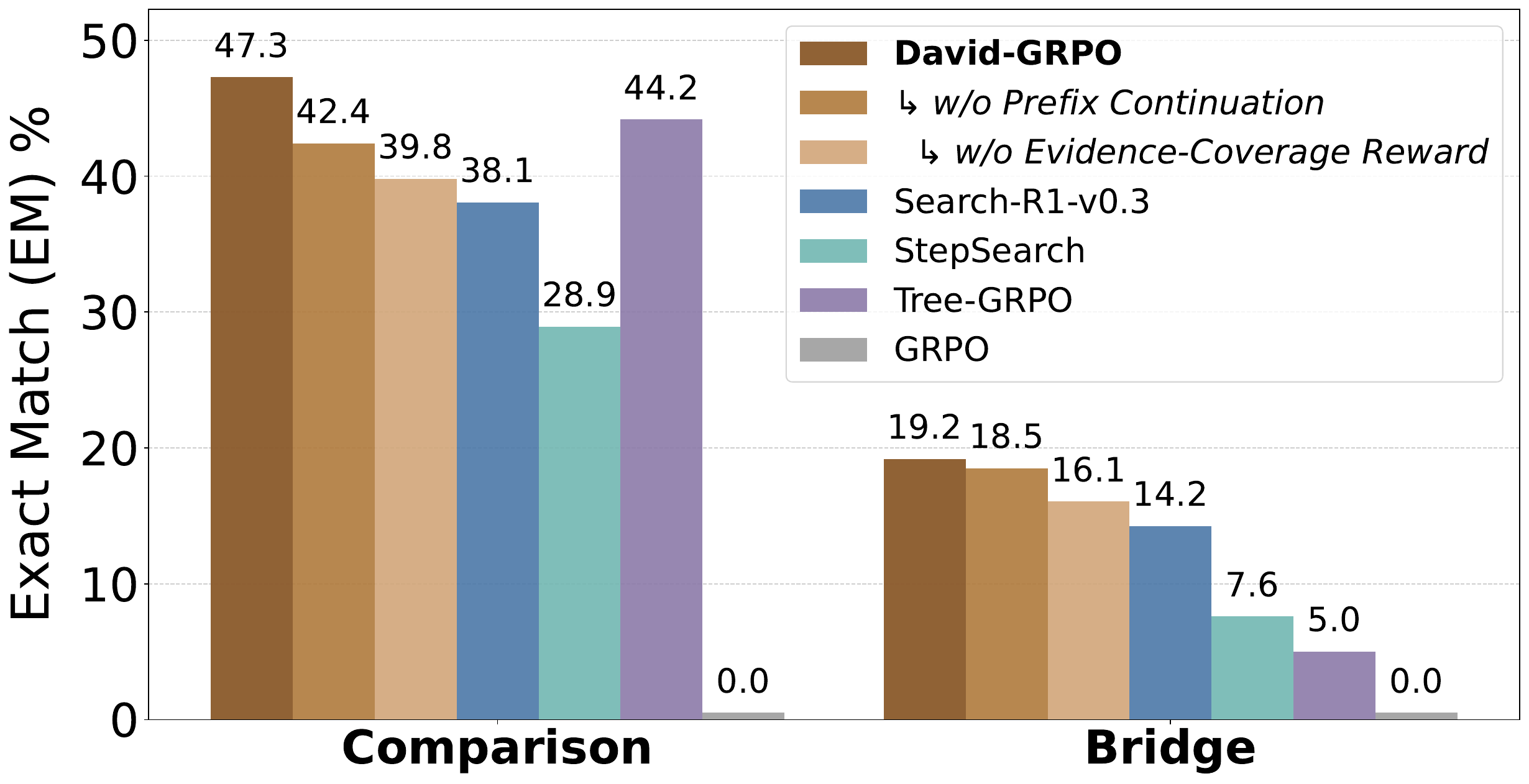}
    \caption{EM by user question types.}
    \label{fig:hotpot-em-hop}
  \end{subfigure}\hfill
  % (b) Retrieval Actions
  \begin{subfigure}{0.49\linewidth}
    \centering
    \includegraphics[width=\linewidth]{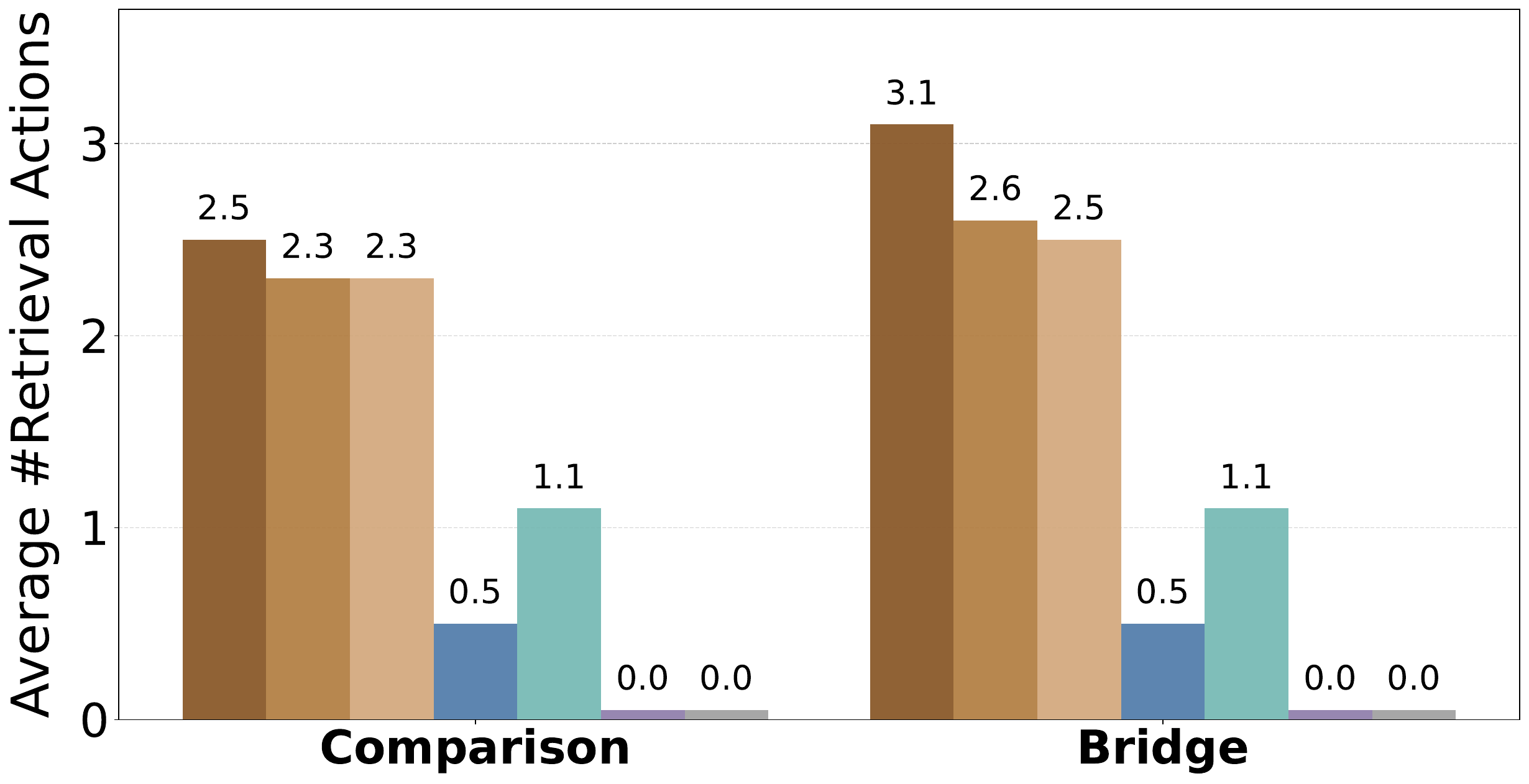}
    \caption{Avg. retrieval actions by user question types.}
    \label{fig:hotpot-retrieval-hop}
  \end{subfigure}

  \caption{
    EM and average number of unique retrieval actions on HotpotQA by user question types with Qwen2.5-1.5B. Search-R1-v0.3 is trained along with its retrieval reward.
  }
  \label{fig:hotpot-subset}
\end{figure*}
\paragraph{Reasoning Types in HotpotQA.}
We analyze HotpotQA performance across two reasoning types: \textit{comparison} (comparing mentioned entities) and \textit{bridge} (identifying missing intermediate entities). As shown in Figure~\ref{fig:hotpot-em-hop}, \ours leads in both categories. Tree-GRPO performs well on \textit{comparison} tasks but drops to the second-lowest on \textit{bridge} tasks, where retrieval is needed to locate hidden entities.
Figure~\ref{fig:hotpot-retrieval-hop} shows the retrieval behavior behind this gap. \ours and its ablations perform 2--3 retrieval steps. StepSearch averages 1.1 actions, and Tree-GRPO records zero actions, relying only on parametric knowledge (see Appendix~\ref{appendix:case_study_tree_grpo}).

\begin{figure*}[t]
  \centering

  \begin{subfigure}{0.49\linewidth}
    \centering
    \includegraphics[width=\linewidth] {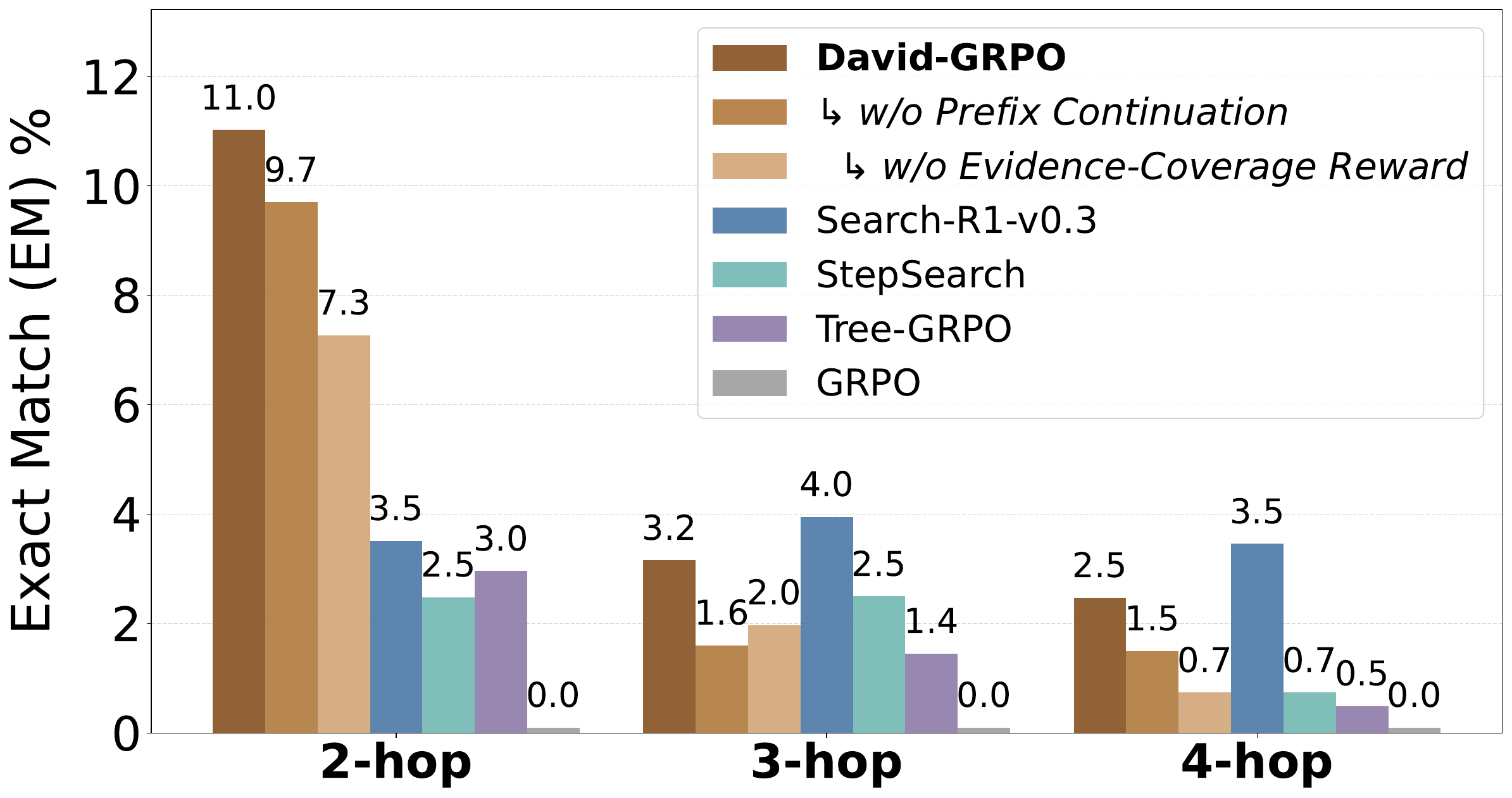}
    \caption{EM by reasoning hops.}
    \label{fig:musique-em-hop}
  \end{subfigure}\hfill
  \begin{subfigure}{0.49\linewidth}
    \centering
    \includegraphics[width=\linewidth]{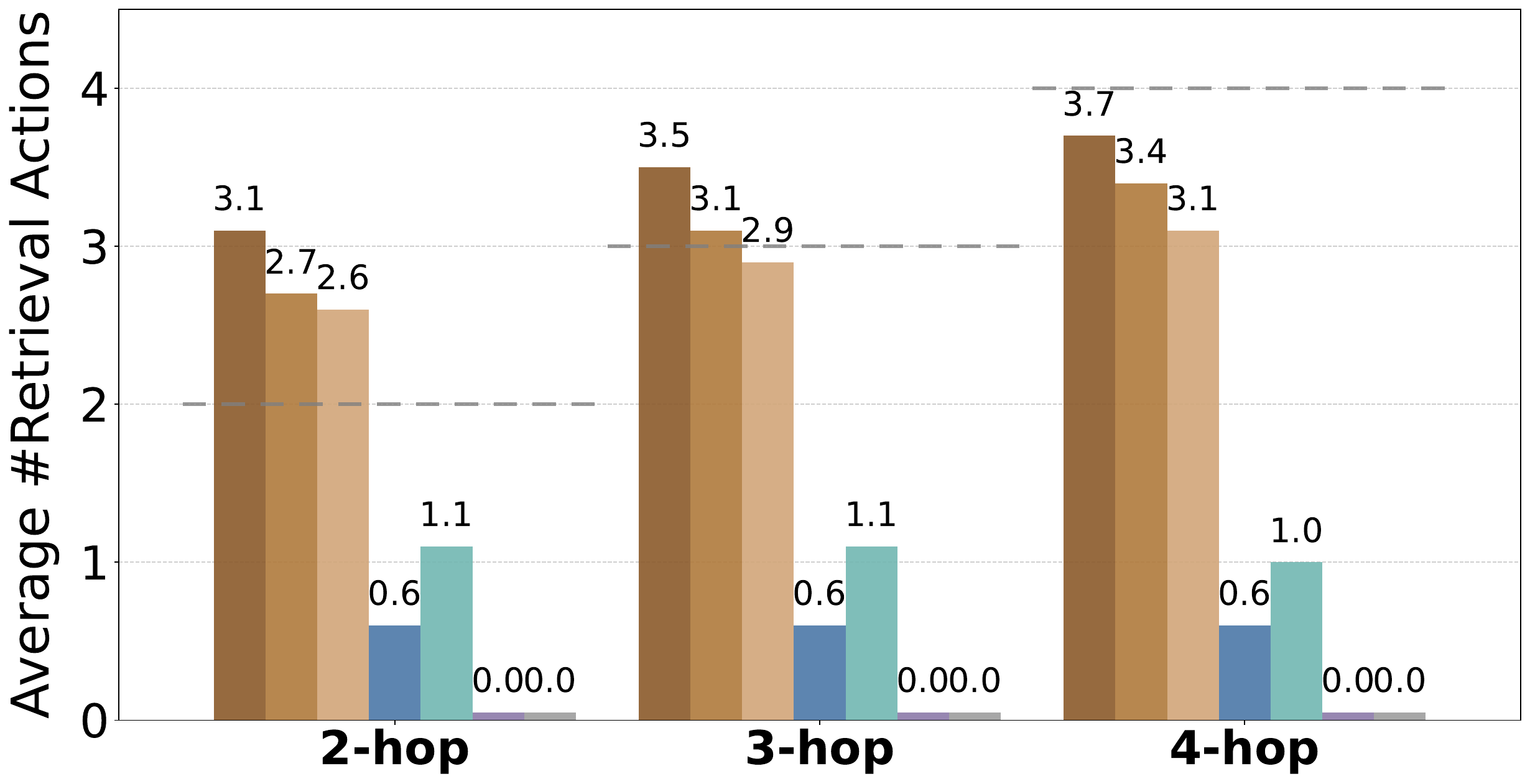}
    \caption{Avg. retrieval actions by reasoning hops.}
    \label{fig:musique-retrieval-hop}
  \end{subfigure}

  \caption{
    EM and average number of unique retrieval actions on MuSiQue by reasoning hops with Qwen2.5-1.5B.
    Dashed lines indicate the minimum number of retrieval actions required for each hop subset. Search-R1-v0.3 is trained along with its retrieval reward.
  }
  \label{fig:musique-hop-breakdown}
\end{figure*}

We stratify MuSiQue performance by reasoning hops.
As shown in Figure~\ref{fig:musique-em-hop}, \ours ranks first or second across all depths.
Search-R1-v0.3 scores slightly higher on the 3- and 4-hop subsets, but Figure~\ref{fig:musique-retrieval-hop} shows that it averages only one retrieval action even for complex queries (see Appendix~\ref{appendix:case_study_search_r1}).
This indicates that its higher-hop scores are not tied to multi-turn retrieval.
In contrast, \ours increases retrieval depth with the annotated hop count.

\begin{table}[h]
\centering
\setlength{\tabcolsep}{4pt}
\scriptsize
\begin{adjustbox}{max width=\columnwidth}
\begin{tabular}{l c c c c}
\toprule
\multirow{2}{*}{\textbf{Method}} & \multicolumn{2}{c}{\textbf{EM}} & \multicolumn{2}{c}{\textbf{Retrieval}} \\
\cmidrule(lr){2-3} \cmidrule(lr){4-5}
& \textbf{3-hop} & \textbf{4-hop} & \textbf{3-hop} & \textbf{4-hop} \\
\midrule
Search-R1-v0.3 & 4.0 & 3.5 & 0.6 & 0.6 \\\midrule
% \rowcolor{pennyColor!50}
\textbf{\ours} & 3.2 & 2.5 & 3.5 & 3.7 \\
\textit{\phantom{0}+ oracle answer doc append} & 3.2 & 1.7 & 3.9 & 4.2 \\
\textit{\phantom{0}+ oracle answer doc replace} & 8.2 & 5.7 & 3.9 & 4.3 \\
\phantom{0}+ continuation only & 3.3 & 1.0 & \textbf{4.2} & 4.3 \\
\phantom{0}+ early-hop truncation + continuation & \textbf{4.1} & \textbf{4.2} & 4.1 & \textbf{4.8} \\
\bottomrule
\end{tabular}
\end{adjustbox}
\caption{{Controlled interventions on MuSiQue 3-hop and 4-hop subsets.} EM and average number of retrieval actions with Qwen2.5-1.5B.}
\label{tab:musique_depth_intervention}
\end{table}

We further analyze the 3-hop and 4-hop subsets in Table~\ref{tab:musique_depth_intervention}.
The average retrieval depth of \ours increases only from 3.5 to 3.7 when moving from 3-hop to 4-hop questions, leaving little margin for missed evidence.
Appending an oracle answer document does not improve EM, whereas replacing one final-hop retrieved document with it gives larger gains.
Recall of later hops is not the only bottleneck. The gain from replacement points to the retrieved context as a second bottleneck, either through length or through distracting documents.
Early-hop truncation followed by continuation improves 3-hop EM from 3.2 to 4.1 and 4-hop EM from 2.5 to 4.2.

\subsection{Full SFT with Large Annotation Sets} \label{analysis:full_sft}
\begin{table*}[t]
\centering
\setlength{\tabcolsep}{3pt} % 열 간격을 조금 넓혀 가독성을 높였습니다.

% 테이블 너비를 페이지 너비에 맞게 자동 조절
\begin{adjustbox}{max width=\textwidth}
\scriptsize
\begin{tabular}{l cc cc cc cc cc >{\columncolor{cleanColor}}c >{\columncolor{cleanColor}}c cc}
\toprule
\multirow{2.5}{*}{\textbf{Method}} 
& \multicolumn{2}{c}{\textbf{HotpotQA}} 
& \multicolumn{2}{c}{\textbf{2wiki}} 
& \multicolumn{2}{c}{\textbf{Musique}} 
& \multicolumn{2}{c}{\textbf{Bamboogle}} 
& \multicolumn{2}{c}{\textbf{Bamtwoogle}} 
& \multicolumn{2}{c}{\cellcolor{cleanColor}\textbf{Antileak-m}}
& \multicolumn{2}{c}{\textbf{Avg.}} \\
\cmidrule(lr){2-3} \cmidrule(lr){4-5} \cmidrule(lr){6-7} \cmidrule(lr){8-9} \cmidrule(lr){10-11} \cmidrule(lr){12-13} \cmidrule(lr){14-15}
 & \textbf{EM} & \textbf{F1} 
 & \textbf{EM} & \textbf{F1} 
 & \textbf{EM} & \textbf{F1} 
 & \textbf{EM} & \textbf{F1} 
 & \textbf{EM} & \textbf{F1} 
 & \textbf{EM} & \textbf{F1}
 & \textbf{EM} & \textbf{F1} \\
\midrule

\multicolumn{15}{c}{\textbf{\textsc{High Training Budget}}} \\ \cmidrule{1-15}
Tree-GRPO$^\dagger$ & 29.5  & - & 26.8 & - & \phantom{0}6.6 & - & 13.6 & - & - & - & - & - & - & - \\
\cmidrule{1-15}
\multicolumn{15}{c}{\textbf{\textsc{Low Training Budget}}} \\ \cmidrule{1-15}

% --- Full SFT ---
Full SFT 
& 23.8 & 32.4
& 20.6 & 26.6
& \phantom{0}\textcolor{red}{7.1} & 12.9
& \textcolor{red}{22.4} & 32.4 
& 20.0 & 21.7 
& 37.1 & 46.1
& 21.8 & 28.7 \\

% --- AutoCoA ---
\phantom{0}+ GRPO (AutoCoA)
& \textcolor{red}{\underline{32.1}} &  {\underline{42.6}}
& \textcolor{red}{\underline{27.2}} & \underline{34.5}
& \textcolor{red}{\underline{11.2}} & \underline{17.6}
& \textcolor{red}{\underline{32.8}} & \textbf{40.5}
& \underline{23.0} & \underline{29.4} 
& \underline{45.9} & \underline{52.1}
& \underline{28.7} & \underline{36.1} \\

% --- Ours ---
\rowcolor{pennyColor!50}
\phantom{0}\textbf{+ \ours}
& \textcolor{red}{\textbf{32.6}} & \textbf{42.7}
& \textcolor{red}{\textbf{28.1}} & \textbf{36.2}
& \textcolor{red}{\textbf{12.2}} & \textbf{19.0}
& \textcolor{red}{\textbf{33.6}} & \underline{39.4} 
& \textbf{32.0} & \textbf{37.8} 
& \textbf{46.2} & \textbf{53.6}
& \textbf{30.8} & \textbf{38.1} \\

\bottomrule
\end{tabular}
\end{adjustbox}
\caption{{Comparison with Full SFT Initialization.} Results across multi-hop QA benchmarks including EM and F1 scores. Results marked with $^\dagger$ are reported from \citet{ji2025treesearch}. Scores marked in \textcolor{red}{\textbf{red}} indicate instances where methods under \textsc{Low Training Budget} outperform the best \textsc{High Training Budget} baseline. Base model is Qwen2.5-1.5B.}
\label{tab:full_sft_comparison_complete}
\end{table*}

We also test \ours with Full SFT initialization when large trajectory annotation sets are available. Table~\ref{tab:full_sft_comparison_complete} shows that GRPO after Full SFT improves over the Full SFT baseline for both AutoCoA and \ours.
Both AutoCoA and \ours surpass the high-budget Tree-GRPO baseline across the reported benchmarks (marked in \textcolor{red}{red}). \ours also outperforms AutoCoA on most benchmarks, with the largest gap on BamTwoogle: $9.0$ pp in EM and $8.4$ pp in F1.

\subsection{\edit{Evaluation on Qwen3-0.6B}}
\label{appendix:qwen3}
\begin{table*}[t]
\centering
\scriptsize
\setlength{\tabcolsep}{2.5pt}
\begin{adjustbox}{max width=\textwidth}
\begin{tabular}{l cc cc cc cc cc cc cc c}
\toprule
\multirow{2.5}{*}{\edit{\textbf{Method}}}
& \multicolumn{2}{c}{\edit{\textbf{HotpotQA}}}
& \multicolumn{2}{c}{\edit{\textbf{2wiki}}}
& \multicolumn{2}{c}{\edit{\textbf{Musique}}}
& \multicolumn{2}{c}{\edit{\textbf{Bamboogle}}}
& \multicolumn{2}{c}{\edit{\textbf{Bamtwoogle}}}
& \multicolumn{2}{c}{\edit{\textbf{Antileak-m}}}
& \multicolumn{3}{c}{\edit{\textbf{Avg.}}} \\
\cmidrule(lr){2-3} \cmidrule(lr){4-5} \cmidrule(lr){6-7} \cmidrule(lr){8-9} \cmidrule(lr){10-11} \cmidrule(lr){12-13} \cmidrule(lr){14-16}
& \edit{\textbf{EM}} & \edit{\textbf{F1}}
& \edit{\textbf{EM}} & \edit{\textbf{F1}}
& \edit{\textbf{EM}} & \edit{\textbf{F1}}
& \edit{\textbf{EM}} & \edit{\textbf{F1}}
& \edit{\textbf{EM}} & \edit{\textbf{F1}}
& \edit{\textbf{EM}} & \edit{\textbf{F1}}
& \edit{\textbf{EM}} & \edit{\textbf{F1}} & \edit{\textbf{Search}} \\
\midrule
\edit{Tree-GRPO} & 10.1 & 14.7 & \underline{22.8} & 25.0 & \phantom{0}0.7 & \phantom{0}4.8 & \phantom{0}7.2 & 10.4 & \phantom{0}4.0 & \phantom{0}6.9 & 11.7 & 15.5 & \phantom{0}9.4 & 12.9 & \edit{0.00} \\
\edit{StepSearch} & \underline{19.8} & \textbf{30.9} & 11.8 & 21.1 & \textbf{\phantom{0}8.7} & \textbf{16.8} & \textbf{23.2} & \textbf{34.7} & \textbf{17.0} & \textbf{20.7} & \textbf{35.4} & \textbf{45.0} & \textbf{19.3} & \textbf{28.2} & \edit{\underline{1.96}} \\
\edit{Search-R1-v0.3 \textit{w/ retrieval reward}} & \phantom{0}9.7 & 14.6 & \textbf{24.2} & \underline{27.3} & \phantom{0}1.0 & \phantom{0}5.4 & \phantom{0}5.6 & \phantom{0}8.8 & \phantom{0}4.0 & \phantom{0}7.9 & 12.5 & 15.8 & \phantom{0}9.5 & 13.3 & \edit{0.01} \\
\rowcolor{pennyColor!50}
\edit{\textbf{\ours}} & \textbf{20.2} & \underline{28.9} & 22.7 & \textbf{29.2} & \underline{\phantom{0}5.9} & \underline{11.9} & \underline{17.6} & \underline{26.0} & \underline{13.0} & \underline{17.6} & \underline{34.5} & \underline{41.9} & \underline{19.0} & \underline{25.9} & \edit{\textbf{2.60}} \\
\bottomrule
\end{tabular}
\end{adjustbox}
\caption{\edit{{Qwen3-0.6B evaluation.} EM and F1 of low-budget methods with the Qwen3-0.6B backbone. Best results are \textbf{bolded} and second-best \underline{underlined}. \textit{Search} is the average number of search actions per question over the six benchmarks. Qwen3-0.6B post-dates the 2024 collection of Antileak-m, so the Antileak-m scores are not contamination-free (Appendix~\ref{appendix:cutoff}).}}
\label{tab:qwen3_results}
\end{table*}
\edit{Table~\ref{tab:qwen3_results} repeats the low-budget comparison with Qwen3-0.6B~\citep{yang2025qwen3}, a backbone released in 2025. StepSearch ties \ours at 19 average EM, and both learn multi-turn retrieval, with 2.0 and 2.6 searches per question. Tree-GRPO and Search-R1-v0.3 stop searching on this backbone as well and stay below 10 EM. StepSearch sustains retrieval on only two of the five backbones we test, Qwen3-0.6B and Qwen2.5-7B, and on Qwen2.5-7B it trails \ours by 9.1 average EM (Table~\ref{tab:qwen7b_results}).}

\subsection{\edit{Ablation on Qwen2.5-7B}}
\label{appendix:ablation_7b}
\begin{table*}[t]
\scriptsize
\centering
\setlength{\tabcolsep}{3.0pt}
\begin{adjustbox}{max width=\textwidth}
\begin{tabular}{l cc cc cc cc cc >{\columncolor{cleanColor}}c >{\columncolor{cleanColor}}c cc}
\toprule
\multirow{2.5}{*}{\edit{\textbf{Method}}}
& \multicolumn{2}{c}{\edit{\textbf{HotpotQA}}}
& \multicolumn{2}{c}{\edit{\textbf{2wiki}}}
& \multicolumn{2}{c}{\edit{\textbf{Musique}}}
& \multicolumn{2}{c}{\edit{\textbf{Bamboogle}}}
& \multicolumn{2}{c}{\edit{\textbf{Bamtwoogle}}}
& \multicolumn{2}{c}{\cellcolor{cleanColor}\edit{\textbf{Antileak-m}}}
& \multicolumn{2}{c}{\edit{\textbf{Avg.}}} \\
\cmidrule(lr){2-3} \cmidrule(lr){4-5} \cmidrule(lr){6-7} \cmidrule(lr){8-9} \cmidrule(lr){10-11} \cmidrule(lr){12-13} \cmidrule(lr){14-15}
& \edit{\textbf{EM}} & \edit{\textbf{F1}}
& \edit{\textbf{EM}} & \edit{\textbf{F1}}
& \edit{\textbf{EM}} & \edit{\textbf{F1}}
& \edit{\textbf{EM}} & \edit{\textbf{F1}}
& \edit{\textbf{EM}} & \edit{\textbf{F1}}
& \edit{\textbf{EM}} & \edit{\textbf{F1}}
& \edit{\textbf{EM}} & \edit{\textbf{F1}} \\
\midrule
\rowcolor{pennyColor!50}
\edit{\textbf{\ours}} & \edit{\textbf{35.2}} & \edit{\textbf{45.6}} & \edit{\textbf{37.7}} & \edit{\textbf{44.5}} & \edit{\textbf{14.2}} & \edit{\textbf{21.6}} & \edit{\textbf{40.0}} & \edit{\textbf{52.3}} & \edit{\textbf{44.0}} & \edit{\textbf{47.1}} & \edit{\textbf{55.2}} & \edit{\textbf{59.8}} & \edit{\textbf{37.7}} & \edit{\textbf{45.1}} \\
\rowcolor{abl1Color!45}
\edit{\textit{\phantom{0}w/o continuation}} & \edit{31.9} & \edit{41.6} & \edit{31.2} & \edit{37.9} & \edit{12.4} & \edit{19.8} & \edit{32.0} & \edit{40.6} & \edit{41.0} & \edit{44.2} & \edit{52.5} & \edit{57.0} & \edit{33.5} & \edit{40.2} \\
\rowcolor{abl2Color!30}
\edit{\textit{\phantom{00}w/o coverage reward}} & \edit{\underline{33.4}} & \edit{\underline{43.6}} & \edit{\underline{35.1}} & \edit{\underline{42.1}} & \edit{\underline{13.4}} & \edit{\underline{21.1}} & \edit{\underline{37.6}} & \edit{\underline{47.3}} & \edit{\underline{42.0}} & \edit{\underline{44.5}} & \edit{\underline{54.5}} & \edit{\underline{58.6}} & \edit{\underline{36.0}} & \edit{\underline{42.9}} \\
\bottomrule
\end{tabular}
\end{adjustbox}
\caption{\edit{{Ablation on evidence-guided continuation with Qwen2.5-7B.} EM and F1 of the gold variant of \ours and its ablations under the low-budget setting. The variants follow Table~\ref{tab:ablation_grounding_components}.}}
\label{tab:ablation_7b}
\end{table*}
\edit{Table~\ref{tab:ablation_7b} repeats the ablation of Table~\ref{tab:ablation_grounding_components} with Qwen2.5-7B for the gold variant. The full \ours system gives the best EM and F1 on all six benchmarks, and the two ablations fall below it by 4.2 and 1.7 average EM.}

\subsection{Generalization to Longer Tool-use Chains}
\label{appendix:bfcl}
\begin{table}[h]
\centering
\renewcommand{\arraystretch}{0.95}
\setlength{\tabcolsep}{2pt}
\scriptsize
\begin{adjustbox}{max width=\columnwidth}
\begin{tabular}{l c c}
\toprule
\textbf{Method} & \textbf{Size} & \textbf{Acc. (\%)} \\
\midrule
GPT-4.1-2025-04-14 & - & 43.0 \\
Grok-4-0709 & - & 44.0 \\
\midrule
xLAM-2-1b-fc-r (reported) & 1B & 25.5 \\
xLAM-2-1b-fc-r (64K override) & 1B & 41.5 \\
\phantom{0}+ GRPO & 1B & 42.5 \\
\rowcolor{pennyColor!50}
\phantom{0}\textbf{+ \ours} & \textbf{1B} & \textbf{44.5} \\
\bottomrule
\end{tabular}
\end{adjustbox}
\caption{{BFCL tool-use accuracy.}}
\label{tab:bfcl_tooluse}
% \vspace{-1.2em}
\end{table}

We also evaluate on the BFCL multi-turn long-context subset~\citep{patil2025the}, where each problem requires 6.02 tool calls on average.

We use xLAM-2-1b-fc-r~\citep{prabhakar2026apigenmt} as the base model. 
Because it already supports multi-turn tool calling, we omit expert trajectory seeding and train with the same GRPO setup under a 64K context override.
As shown in Table~\ref{tab:bfcl_tooluse}, \ours improves accuracy from 41.5 to 44.5 and outperforms standard GRPO by 2.0 pp.
This suggests that \ours can help outside Wikipedia-style multi-hop QA.

\subsection{\edit{Tree-GRPO under a Controlled Setting}}
\label{appendix:tree_grpo_controlled}
\begin{table}[h]
\centering
\scriptsize
\setlength{\tabcolsep}{6pt}
\begin{adjustbox}{max width=\columnwidth}
\begin{tabular}{l c}
\toprule
\edit{\textbf{Tree-GRPO backbone}} & \edit{\textbf{HotpotQA validation EM}} \\
\midrule
\edit{Qwen2.5-1.5B} & \phantom{0}\edit{0.0} \\
\edit{Qwen2.5-1.5B-Instruct} & \edit{14.5} \\
\bottomrule
\end{tabular}
\end{adjustbox}
\caption{\edit{{Tree-GRPO backbone choice.} HotpotQA validation EM (\%) of Tree-GRPO trained under the low-budget setting from the base and the instruct checkpoint of Qwen2.5-1.5B.}}
\label{tab:tree_grpo_backbone}
\end{table}

\begin{table}[h]
\centering
\scriptsize
\setlength{\tabcolsep}{6pt}
\begin{adjustbox}{max width=\columnwidth}
\begin{tabular}{l l c}
\toprule
\edit{\textbf{7B agent}} & \edit{\textbf{Checkpoint or stage}} & \edit{\textbf{Searches}} \\
\midrule
\edit{Tree-GRPO, high budget} & \edit{Step 40 (minimum)} & \edit{1.29} \\
\edit{Tree-GRPO, high budget} & \edit{Step 80 (maximum)} & \edit{1.96} \\
\edit{Tree-GRPO, high budget} & \edit{Step 150 (best checkpoint)} & \edit{1.82} \\
\rowcolor{pennyColor!50}
\edit{\textbf{\ours}, low budget} & \edit{After expert trajectory seeding} & \edit{2.99} \\
\bottomrule
\end{tabular}
\end{adjustbox}
\caption{\edit{{Search counts of 7B agents on HotpotQA.} Average number of search actions per question. The Tree-GRPO rows are the released high-budget run at three checkpoints; the \ours row is measured right after expert trajectory seeding, before the main GRPO stage.}}
\label{tab:tree_grpo_7b_search}
\end{table}

\begin{table*}[t]
\centering
\scriptsize
\setlength{\tabcolsep}{3.0pt}
\begin{adjustbox}{max width=\textwidth}
\begin{tabular}{l cc cc cc cc cc >{\columncolor{cleanColor}}c >{\columncolor{cleanColor}}c cc}
\toprule
\multirow{2.5}{*}{\edit{\textbf{Method}}}
& \multicolumn{2}{c}{\edit{\textbf{HotpotQA}}}
& \multicolumn{2}{c}{\edit{\textbf{2wiki}}}
& \multicolumn{2}{c}{\edit{\textbf{Musique}}}
& \multicolumn{2}{c}{\edit{\textbf{Bamboogle}}}
& \multicolumn{2}{c}{\edit{\textbf{Bamtwoogle}}}
& \multicolumn{2}{c}{\cellcolor{cleanColor}\edit{\textbf{Antileak-m}}}
& \multicolumn{2}{c}{\edit{\textbf{Avg.}}} \\
\cmidrule(lr){2-3} \cmidrule(lr){4-5} \cmidrule(lr){6-7} \cmidrule(lr){8-9} \cmidrule(lr){10-11} \cmidrule(lr){12-13} \cmidrule(lr){14-15}
& \edit{\textbf{EM}} & \edit{\textbf{F1}}
& \edit{\textbf{EM}} & \edit{\textbf{F1}}
& \edit{\textbf{EM}} & \edit{\textbf{F1}}
& \edit{\textbf{EM}} & \edit{\textbf{F1}}
& \edit{\textbf{EM}} & \edit{\textbf{F1}}
& \edit{\textbf{EM}} & \edit{\textbf{F1}}
& \edit{\textbf{EM}} & \edit{\textbf{F1}} \\
\midrule
\edit{Tree-GRPO (released configuration)} & \edit{12.9} & \edit{18.6} & \edit{20.5} & \edit{23.3} & \phantom{0}\edit{2.1} & \phantom{0}\edit{7.2} & \edit{12.0} & \edit{15.5} & \phantom{0}\edit{5.0} & \phantom{0}\edit{8.2} & \edit{11.7} & \edit{15.7} & \edit{10.7} & \edit{14.8} \\
\edit{Tree-GRPO (controlled)} & \edit{14.7} & \edit{21.3} & \edit{21.8} & \edit{25.9} & \phantom{0}\edit{1.9} & \phantom{0}\edit{8.9} & \phantom{0}\edit{4.8} & \edit{10.2} & \edit{10.0} & \edit{14.2} & \edit{16.5} & \edit{21.9} & \edit{11.6} & \edit{17.0} \\
\rowcolor{pennyColor!50}
\edit{\textbf{\ours}} & \edit{\textbf{24.8}} & \edit{\textbf{33.8}} & \edit{\textbf{27.2}} & \edit{\textbf{32.3}} & \phantom{0}\edit{\textbf{7.1}} & \edit{\textbf{12.6}} & \edit{\textbf{14.4}} & \edit{\textbf{24.2}} & \edit{\textbf{22.0}} & \edit{\textbf{25.4}} & \edit{\textbf{36.3}} & \edit{\textbf{41.1}} & \edit{\textbf{22.0}} & \edit{\textbf{28.2}} \\
\bottomrule
\end{tabular}
\end{adjustbox}
\caption{\edit{{Tree-GRPO under a controlled setting.} EM and F1 of Tree-GRPO with its released configuration and with the prompt, tag format, and training retrieval pool of \ours. The backbone is Qwen2.5-1.5B-Instruct for Tree-GRPO and Qwen2.5-1.5B for \ours.}}
\label{tab:tree_grpo_controlled}
\end{table*}
\edit{We check whether the shallow retrieval of Tree-GRPO in the low-budget setting is an artifact of our adaptation. First, the choice of the instruct checkpoint favors Tree-GRPO. With the Qwen2.5-1.5B base checkpoint, training fails, reaching 0.0 HotpotQA validation EM, whereas the instruct checkpoint reaches 14.5 (Table~\ref{tab:tree_grpo_backbone}). Second, we retrain Tree-GRPO with the prompt, tag format, and training retrieval pool of \ours. As shown in Table~\ref{tab:tree_grpo_controlled}, this controlled variant improves average EM from 10.7 to 11.6 and F1 from 14.8 to 17.0, but still issues almost no search actions (0.001 per question) and remains well below \ours. Third, shallow retrieval also appears in the publicly released high-budget runs of Tree-GRPO. In the 1.5B run, the number of valid searches per question stabilizes near 1.0 from step 49 onward, and all 23 readable responses in the final validation log perform exactly one search before answering. The 7B run averages 1.82 searches on HotpotQA at its best checkpoint and at most 1.96 at any checkpoint, whereas the low-budget \ours already averages 2.99 right after expert trajectory seeding, before the main GRPO stage (Table~\ref{tab:tree_grpo_7b_search}). Shallow retrieval of Tree-GRPO appears in every setting we examined, including the authors' own high-budget runs.}

\subsection{Dynamics of Prefix Continuation}
\label{appendix:grounded_expansion_dynamics}
\begin{figure}[h]
    \centering
    \includegraphics[width=0.85\columnwidth]{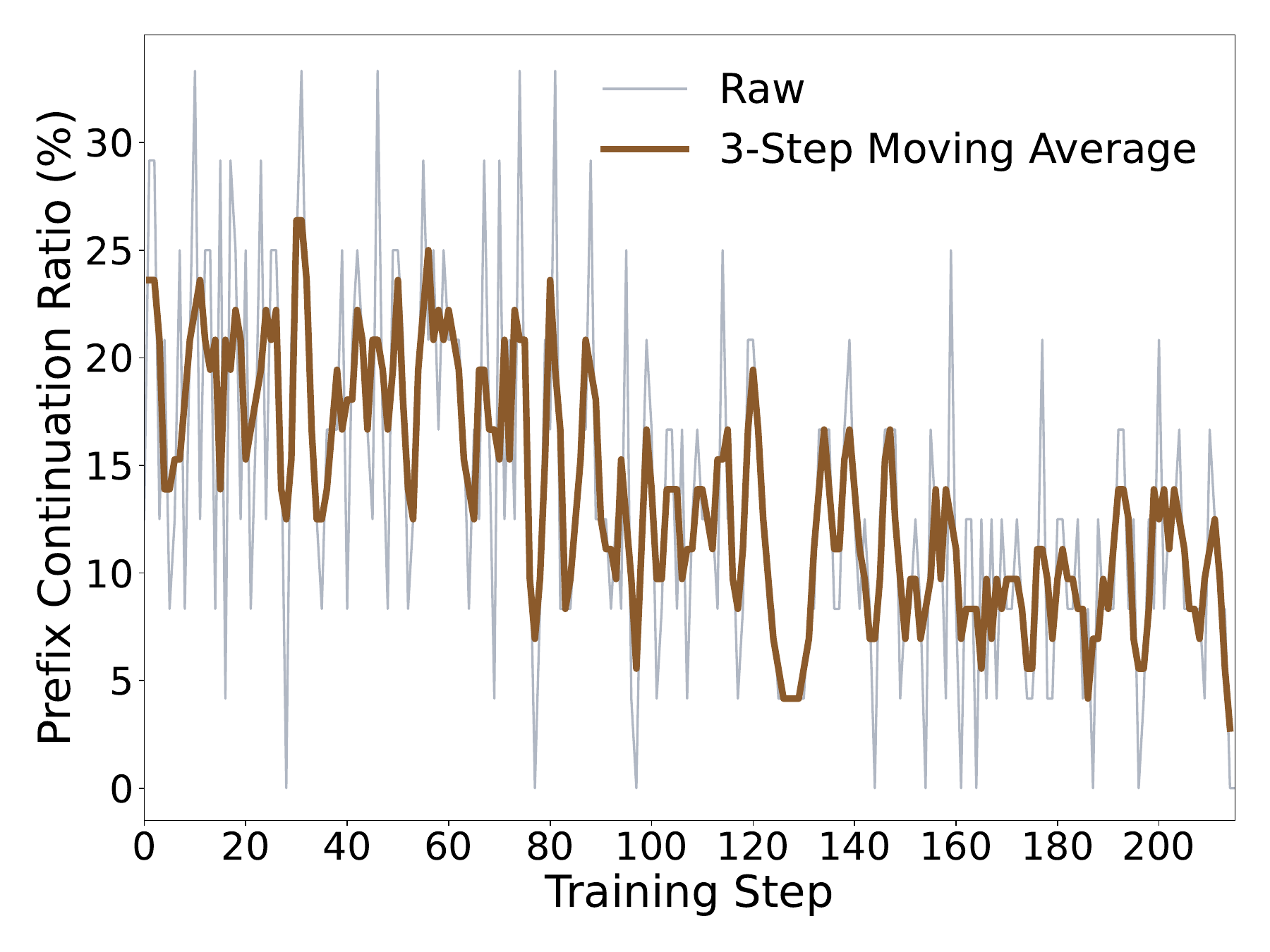}
    \caption{Prefix continuation ratio (\%) over training steps based on Qwen2.5-1.5B with 4 NVIDIA RTX 3090 GPUs. The light gray line indicates the raw values, while the thick brown line denotes the 3-step moving average.}
    \label{fig:grounded_expansion_ratio}
\end{figure}

We investigate the temporal behavior of prefix continuation during training, as illustrated in Figure~\ref{fig:grounded_expansion_ratio}.
The continuation ratio is higher in the early stages, between 20 and 33\% of a batch (about 25\% on the moving average), when the policy early in training rarely produces high-reward trajectories.
As training progresses and the agent generates more high-reward trajectories on its own, the continuation frequency declines.

\subsection{Training Stability}
\label{appendix:training_stability}
\begin{table*}[t]
\centering
\setlength{\tabcolsep}{4pt}
\begin{adjustbox}{max width=\textwidth}
\scriptsize
\begin{tabular}{l c c c c c >{\columncolor{cleanColor}}c c}
\toprule
\textbf{Method} & \textbf{HotpotQA} & \textbf{2wiki} & \textbf{Musique} & \textbf{Bamboogle} & \textbf{Bamtwoogle} & \cellcolor{cleanColor}\textbf{Antileak-m} & \textbf{Avg.} \\
\midrule
Search-R1-v0.3
& $14.5{\pm}5.0$ & $18.0{\pm}2.8$ & $2.4{\pm}1.4$ & $5.9{\pm}1.8$ & $4.0{\pm}1.0$ & $13.8{\pm}3.5$ & $9.8{\pm}2.1$ \\
\rowcolor{pennyColor!50}
\textbf{\ours}
& $\textbf{22.5}{\pm}2.1$ & $\textbf{24.4}{\pm}2.5$ & $\textbf{6.3}{\pm}1.0$ & $\textbf{13.6}{\pm}2.9$ & $\textbf{20.3}{\pm}2.1$ & $\textbf{33.1}{\pm}3.0$ & $\textbf{20.0}{\pm}1.7$ \\
\bottomrule
\end{tabular}
\end{adjustbox}
\caption{{Training stability.} Mean and std. deviation of EM over three Qwen2.5-1.5B training runs.}
\label{tab:stability}
\end{table*}

We repeat the Qwen2.5-1.5B training three times to check seed sensitivity.
We compare \ours against Search-R1-v0.3, the best low-budget baseline by average EM in Table~\ref{tab:main_results}.
As shown in Table~\ref{tab:stability}, the three-seed mean of average EM is $20.0{\pm}1.7$ for \ours against $9.8{\pm}2.1$ for Search-R1-v0.3, and \ours has higher mean EM on all six benchmarks.

\subsection{\edit{Robustness of Evaluation Metrics}}
\label{appendix:llm_judge}
\begin{table*}[t]
\centering
\setlength{\tabcolsep}{4pt}
\scriptsize
\begin{adjustbox}{max width=\textwidth}
\begin{tabular}{l c c c c c >{\columncolor{cleanColor}}c c}
\toprule
\textbf{Method} & \textbf{HotpotQA} & \textbf{2wiki} & \textbf{Musique} & \textbf{Bamboogle} & \textbf{Bamtwoogle} & \cellcolor{cleanColor}\textbf{Antileak-m} & \textbf{Avg.} \\
\midrule
Tree-GRPO & 21.1 & 23.8 & 5.8 & 15.2 & 7.0 & 12.3 & 14.2 \\
StepSearch & 21.9 & 18.4 & 5.8 & 8.0 & 7.0 & 14.5 & 12.6 \\
Search-R1-v0.3 & 31.4 & 27.2 & 9.1 & 15.2 & 8.0 & 19.8 & 18.5 \\
\rowcolor{pennyColor!50}
\textbf{\ours} & \textbf{40.6} & \textbf{34.7} & \textbf{13.4} & \textbf{28.8} & \textbf{28.0} & \textbf{40.9} & \textbf{31.1} \\
\bottomrule
\end{tabular}
\end{adjustbox}
\caption{\edit{{LLM-as-judge evaluation.} LLM-as-judge accuracy (\%) across six multi-hop QA benchmarks under the low training budget setting. The backbone model is Qwen2.5-1.5B except for Tree-GRPO with Qwen2.5-1.5B-Inst.}}
\label{tab:llm_judge}
\end{table*}

\edit{EM and F1 are the standard metrics on these benchmarks, whose gold answers contain 2.45 words on average (1.69 to 2.80 per benchmark), and are used by Tree-GRPO~\citep{ji2025treesearch}, StepSearch~\citep{zheng-etal-2025-stepsearch}, and Search-R1-v0.3~\citep{jin2025an}.}

\edit{To verify that the gains of \ours are not specific to lexical overlap, we additionally evaluate all methods with LLM-as-judge accuracy, following prior multi-hop QA work~\citep{sun2025rearter,song2025r1searcher}. We use Claude Sonnet 5 as the judge. Given the question, the gold answer, and the extracted predicted answer, the judge decides whether the prediction is semantically correct, with the prompt shown in Appendix~\ref{appendix:prompt}. As shown in Table~\ref{tab:llm_judge}, \ours remains the best-performing method on all six benchmarks, and the macro-average ranking of the low-budget methods is identical under EM, F1, and LLM-as-judge evaluation.}

\begin{table}[t]
\centering
\renewcommand{\arraystretch}{1.15}
\scriptsize
\begin{adjustbox}{max width=\columnwidth}
\begin{tabular}{l c}
\toprule
\textbf{Method} & \textbf{Extraction success (\%)} \\
\midrule
Tree-GRPO & \textbf{99.94} \\
StepSearch & 99.90 \\
Search-R1-v0.3 & 99.82 \\
\rowcolor{pennyColor!50}
\textbf{\ours} & 98.27 \\
\bottomrule
\end{tabular}
\end{adjustbox}
\caption{\edit{Final-answer extraction success rate (\%) aggregated across the six benchmarks under the low training budget setting.}}
\label{tab:answer_extraction}
\end{table}

\edit{We also confirm that answer-extraction differences do not explain the gains. Table~\ref{tab:answer_extraction} reports final-answer extraction success rates aggregated across the six benchmarks. Extraction succeeds for at least 98\% of the answers of every method, and the rate of \ours is slightly below the baselines, so the gains do not come from better compatibility with the extraction procedure.}

\subsection{\edit{No-Training Baselines}}
\label{appendix:no_training}
\begin{table*}[t]
\centering
\scriptsize
\begin{adjustbox}{max width=\textwidth}
\begin{tabular}{l c c c c c >{\columncolor{cleanColor}}c}
\toprule
\textbf{Setting} & \textbf{HotpotQA} & \textbf{2wiki} & \textbf{Musique} & \textbf{Bamboogle} & \textbf{Bamtwoogle} & \cellcolor{cleanColor}\textbf{Antileak-m} \\
\midrule
Base Agent w/o RL & 0.0 & 0.1 & 0.0 & 0.0 & 0.0 & 0.0 \\
Fixed Retrieval w/o Training & 0.0 & 0.1 & 0.0 & 0.0 & 0.0 & 0.0 \\
\bottomrule
\end{tabular}
\end{adjustbox}
\caption{\edit{{No-training baselines.} EM (\%) of Qwen2.5-1.5B under two no-training settings across six multi-hop QA benchmarks, using the same overall experimental setup as \ours.}}
\label{tab:no_training}
\end{table*}

\edit{We report no-training controls to verify that the gains of \ours do not come from the tool-use prompt scaffold, the additional retrieval context, or the existing capability of the base model. We evaluate two settings with Qwen2.5-1.5B on all six benchmarks under the same overall experimental setup as \ours.}
\begin{itemize}
    \item \edit{\textbf{Base Agent w/o RL}: the untrained model uses the same agent prompt and search interface as \ours.}
    \item \edit{\textbf{Fixed Retrieval w/o Training}: the question is used as the retrieval query, and the retrieved passages are provided as context without agentic search.}
\end{itemize}
\edit{As shown in Table~\ref{tab:no_training}, both settings achieve at most 0.1 EM on every benchmark. This result is consistent with the no-seeding baseline in Figure~\ref{fig:warm-start}, which uses the same agent prompt and obtains zero EM on all benchmarks.}

\subsection{\edit{Retrieval Behavior Analysis}}
\label{appendix:retrieval_behavior}
\begin{table}[h]
\centering

\tiny
\begin{adjustbox}{max width=\columnwidth}
\begin{tabular}{l cc cc cc cc}
\toprule
& \multicolumn{4}{c}{\textbf{HotpotQA}}
& \multicolumn{4}{c}{\textbf{Musique}} \\
\cmidrule(lr){2-5} \cmidrule(lr){6-9}
\textbf{Method}
& \multicolumn{2}{c}{\textbf{Bridge-doc}}
& \multicolumn{2}{c}{\textbf{Answer-doc}}
& \multicolumn{2}{c}{\textbf{Bridge-doc}}
& \multicolumn{2}{c}{\textbf{Answer-doc}} \\
\cmidrule(lr){2-3} \cmidrule(lr){4-5} \cmidrule(lr){6-7} \cmidrule(lr){8-9}
& \textbf{Any} & \textbf{All}
& \textbf{Any} & \textbf{All}
& \textbf{Any} & \textbf{All}
& \textbf{Any} & \textbf{All} \\
\midrule

Tree-GRPO
& \phantom{0}0.0 & \phantom{0}0.0 & \phantom{0}0.0 & \phantom{0}0.0
& \phantom{0}0.0 & \phantom{0}0.0 & \phantom{0}0.0 & \phantom{0}0.0 \\

StepSearch
& 50.6 & 48.1 & 38.0 & 26.6
& 28.1 & 18.4 & 11.9 & \phantom{0}9.5  \\

Search-R1-v0.3
& 26.3 & 25.1 & 19.8 & 14.0
& 17.2 & 10.4 & \phantom{0}6.4  & \phantom{0}5.0  \\

\rowcolor{pennyColor!50}
\textbf{\ours}
& \textbf{59.1} & \textbf{58.0} & \textbf{51.1} & \textbf{42.4}
& \textbf{48.5} & \textbf{27.5} & \textbf{18.5} & \textbf{15.8} \\

\bottomrule
\end{tabular}
\end{adjustbox}
\caption{Document recall (\%; $\uparrow$) of cumulative retrieved documents $\hat{\mathcal{D}}_{union}$ (Eq.~\eqref{eq:union}).
Any and All measure coverage of annotated answer/bridge documents in $D$. Search-R1-v0.3 is trained with its retrieval reward. Base model is Qwen2.5-1.5B.}
\label{tab:hit_analysis_ans_bridge_clean}
\end{table}

\edit{Table~\ref{tab:hit_analysis_ans_bridge_clean} splits the document recall of Table~\ref{tab:retrieval_behavior} into bridge and answer documents on the benchmarks that annotate them. \ours has the highest recall in every column, and Tree-GRPO has zero because it issues no search actions.}

% \edit{The notion of grounding in \S\ref{sec:related_work} does not mean that retrieving the complete evidence chain is necessary for producing a correct answer. Our position is that relying on parametric knowledge instead of retrieved evidence is undesirable in retrieval-based QA. Without supporting evidence, it is difficult to verify whether an answer is correct or hallucinated. Parametric knowledge is also costly to update when the underlying facts change.}

% \edit{A perfect per-example split between retrieval-dependent and parametrically answerable questions is not possible because the pretraining corpora of the backbone models are unknown. We instead use AntiLeakBench-Multihop, which consists of post-cutoff instances, as a reduced-contamination proxy. As reported in Table~\ref{tab:main_results}, \ours shows its largest gains over the low-budget baselines on this benchmark.}

\edit{\S\ref{sec:retrieval_capabilities} discusses the share of searches issued after the support set is complete (\textit{Unnec.} in Table~\ref{tab:retrieval_behavior}).}

\begin{table*}[t]
\centering
\scriptsize
\begin{adjustbox}{max width=\textwidth}
\begin{tabular}{l l c c c c c c}
\toprule
\edit{\textbf{Dataset}} & \edit{\textbf{Reward and continuation}} & \edit{\textbf{Search}} & \edit{\textbf{Prec.}} & \edit{\textbf{Any}} & \edit{\textbf{All}} & \edit{\textbf{Unnec.}} & \edit{\textbf{EM}} \\
\midrule
\multirow{3}{*}{\edit{HotpotQA}} & \edit{Outcome reward only (\textit{w/o coverage reward})} & \edit{2.52} & \edit{29.2} & \edit{42.9} & \edit{18.5} & \edit{6.4} & \edit{20.8} \\
 & \edit{+ Evidence-coverage reward (\textit{w/o continuation})} & \edit{2.60} & \edit{\textbf{30.8}} & \edit{49.7} & \edit{27.4} & \edit{9.3} & \edit{23.3} \\
 & \edit{+ Prefix continuation (\ours)} & \edit{3.05} & \edit{28.2} & \edit{\textbf{52.3}} & \edit{\textbf{31.6}} & \edit{9.8} & \edit{\textbf{24.8}} \\
\midrule
\multirow{3}{*}{\edit{MuSiQue}} & \edit{Outcome reward only (\textit{w/o coverage reward})} & \edit{2.86} & \edit{\textbf{14.6}} & \edit{22.5} & \edit{4.8} & \edit{1.9} & \edit{4.5} \\
 & \edit{+ Evidence-coverage reward (\textit{w/o continuation})} & \edit{3.05} & \edit{13.7} & \edit{24.5} & \edit{5.5} & \edit{2.1} & \edit{5.8} \\
 & \edit{+ Prefix continuation (\ours)} & \edit{3.48} & \edit{14.3} & \edit{\textbf{25.9}} & \edit{\textbf{8.2}} & \edit{2.6} & \edit{\textbf{7.1}} \\
\bottomrule
\end{tabular}
\end{adjustbox}
\caption{\edit{{Retrieval behavior of the ablations.} Retrieval statistics and EM (\%) on HotpotQA and MuSiQue with the Qwen2.5-1.5B backbone. Each row adds one component to the row above. The metrics follow Table~\ref{tab:retrieval_behavior}.}}
\label{tab:ablation_retrieval_behavior}
\end{table*}

\edit{Table~\ref{tab:ablation_retrieval_behavior} separates the two components of evidence-guided continuation at the level of retrieval behavior. Adding the evidence-coverage reward to the outcome-only variant raises complete-evidence recall (\textit{All}) from 18.5 to 27.4 on HotpotQA and from 4.8 to 5.5 on MuSiQue, and adding prefix continuation raises it further to 31.6 and 8.2. EM follows the same order. Neither component consistently changes document precision, and the increase in searches issued after the evidence chain is complete stays within 2.9 points for the coverage reward and 0.5 points for prefix continuation.}

\section{Benchmark Statistics}
\label{appendix:benchmark_statistics}
\begin{table*}[t]
\centering
\small
\begin{tabularx}{0.9\linewidth}{l r X}
\toprule
\textbf{Name} & \textbf{Size} & \textbf{Comment} \\ \midrule
% 첫 4개 데이터셋 (병합된 주석 사용)
HotpotQA~\citep{yang2018hotpotqa} & 7,405 & \multirow{4}{=}{Used in Search-R1, Tree-GRPO, etc.} \\
2WikiMultiHopQA~\citep{ho-etal-2020-constructing} & 12,576 & \\
MuSiQue~\cite{trivedi-etal-2022-musique} & 2,417 & \\
Bamboogle~\citep{press2023measuring} & 125 & \\ \midrule
% 개별 주석 데이터셋
BamTwoogle~\citep{aksitov2024rest} & 100 & Advanced version of Bamboogle \\ \midrule
AntiLeakBench (multi-hop;~\citealp{wu-etal-2025-antileakbench}) & 455 & 2024 subset \\
\bottomrule
\end{tabularx}
\caption{Statistics of the multi-hop QA evaluation benchmarks.}
\label{tab:eval_dataset}
\end{table*}

\section{Knowledge Cutoff Dates}
\label{appendix:cutoff}

\begin{table}[h]
    \centering

    \scriptsize
    \begin{tabular}{llc}
        \toprule
        \textbf{Model Family} & \textbf{Specific Model} & \textbf{Knowledge Cutoff} \\
        \midrule
        \multirow{5}{*}{Qwen2.5} & Qwen2.5-1.5B & \multirow{5}{*}{Dec 2023\textsuperscript{\textdagger}} \\
                                 & Qwen2.5-1.5B-Instruct &  \\
                                 & Qwen2.5-0.5B &  \\
                                 & Qwen2.5-0.5B-Instruct &  \\
                                 & \edit{Qwen2.5-7B} &  \\
        \midrule
        Llama 3.2 & Llama-3.2-1B-Instruct & Dec 2023\textsuperscript{$\ddagger$} \\
        \midrule
        \multirow{-2}{*}{Qwen3} & \multirow{-2}{*}{Qwen3-0.6B} & \edit{\shortstack{Not disclosed;\\ released Apr 2025\textsuperscript{\S}}} \\
        \bottomrule
    \end{tabular}
    \vspace{0.5ex}
    \begin{flushleft}
        \scriptsize
        \textsuperscript{\textdagger} \url{https://github.com/QwenLM/Qwen3/issues/525} \\
        \textsuperscript{$\ddagger$} \url{https://huggingface.co/meta-llama/Llama-3.2-1B} \\
        \textsuperscript{\S} \url{https://github.com/QwenLM/Qwen3}
    \end{flushleft}
\caption{Knowledge cutoff dates for the language agents evaluated in this work. \edit{The Qwen2.5 and Llama-3.2 models pre-date the data collection period of AntiLeakBench (2024); Qwen3-0.6B does not.}}
    \label{tab:cutoff}
\end{table}

To verify the validity of our evaluation on \textbf{AntiLeakBench} \citep{wu-etal-2025-antileakbench}, we report the official knowledge cutoff dates for the small language agents used in our experiments. As shown in Table~\ref{tab:cutoff}, \edit{the Qwen2.5 and Llama-3.2 models} have a knowledge cutoff of December 2023. Since the multi-hop tasks in AntiLeakBench are constructed from data generated in 2024 (Table~\ref{tab:eval_dataset}), the test samples were not available during pre-training. \edit{Qwen3-0.6B is the exception: its knowledge cutoff is not disclosed and it was released in April 2025, after the collection period, so we do not treat its Antileak-m scores (Table~\ref{tab:qwen3_results}) as contamination-free.}

\section{Implementation Details}
\label{appendix:implmentation}

\subsection{Training and Evaluation Setup}
\begin{table}[h]
\centering
\small
\begin{adjustbox}{max width=\columnwidth}
\begin{tabular}{l c c | c}
\toprule
\textbf{Parameter} & \textbf{Expert Trajectory Seeding} & \textbf{Main GRPO} & \textbf{Evaluation} \\
\midrule
Dataset Size & 4 (expert trajectories) & \edit{4,000} & - \\
Batch Size & 4 & 24 & 128 \\
Rollouts & 5 & 5--10 (avg. 6.1) & 1 \\
Group Size ($G$) & 5 & 5 & - \\
Max Search & 5 & 5 & - \\
Learning Rate & $1\times 10^{-5}$ & $1\times 10^{-6}$ & - \\
KL Loss Coefficient ($\beta$) &  $1\times 10^{-3}$ & $1\times 10^{-3}$ & - \\
Clipping Parameter ($\epsilon$) & 0.2 & 0.2 & - \\
Training Steps & 50 & 215 & - \\
Max Prompt Len. & 1,024 & 1,024 & 2,048 \\
Max Response Len. & 8,192 & 8,192 & 12,288 \\
Temperature & 0.6 & 0.6 & 0.6 \\
Top-$p$ & 1.0 & 1.0 & 1.0 \\
Top-$k$ & -1 & -1 & -1 \\
\bottomrule
\end{tabular}
\end{adjustbox}
\caption{Hyperparameters for Training and Evaluation.}
\label{tab:hyperparameters}
\end{table}

As described in \S\ref{sec:exp_setup}, both training and evaluation were performed on a system equipped with four NVIDIA RTX 3090 GPUs (24GB VRAM each)\edit{, except that the Qwen2.5-7B run was trained on eight NVIDIA RTX A6000 GPUs (48GB VRAM each) with the same batch size and rollout budget}.

\paragraph{Training.}
\edit{The default model trains on the 4{,}000 constructed instances of \S\ref{sec:evidence_construction}. The gold variant and the analyses in Appendix~\ref{appendix:analysis} use} a subset of 5,276 examples from the HotpotQA~\citep{yang2018hotpotqa} training set, provided by~\citet{zhang2025agent}.
The retrieval infrastructure uses \texttt{E5-base-v2}~\citep{wang2022text} as the dense encoder and FAISS~\citep{johnson2019billion} for efficient indexing and similarity search.

\paragraph{Evaluation.}
Our evaluation pipeline uses FlashRAG~\citep{10.1145/3701716.3715313}, following the architectures used in Search-R1(-v0.3)~\citep{jin2025searchr,jin2025an} and {Tree-GRPO~\citep{ji2025treesearch}}. The retrieval and corpus configuration has two settings to align the corpus date with the datasets:

\begin{description}[style=multiline, leftmargin=3cm, font=\bfseries]
    \item[Pipeline] A FlashRAG-based retrieval-augmented generation pipeline.
    \item[Retriever] The Search-R1 retriever, employing the same \texttt{E5-base-v2} and FAISS setup as described in the training phase.
    \item[Corpus (default)] For the majority of tasks, we use the \texttt{wiki-18.jsonl} corpus, following the standard setup in multi-hop reasoning evaluations.
    \item[Corpus (2024)] To accommodate the 2024 subset of AntiLeakBench, we constructed a dedicated index using the Wikipedia 2024 dump. This lets the retriever access information contemporary to the evaluation queries, while keeping the retriever model identical to the default setting.
\end{description}

\subsection{Existing Assets and Licenses}
\label{appendix:asset_licenses}

Table~\ref{tab:asset_licenses} summarizes the main external datasets, models, and software used in our experiments. When a public source does not specify a license, we mark it as not specified.

\begin{table*}[t]
\centering
\small
\begin{tabular}{p{0.37\linewidth} p{0.25\linewidth} p{0.28\linewidth}}
\toprule
\textbf{Asset} & \textbf{Use in this paper} & \textbf{License / terms} \\
\midrule
HotpotQA~\citep{yang2018hotpotqa} & Training and evaluation data & CC BY-SA 4.0 \\
2WikiMultiHopQA~\citep{ho-etal-2020-constructing} & Evaluation data & Apache-2.0 \\
MuSiQue~\citep{trivedi-etal-2022-musique} & Evaluation data & Not specified by public source \\
Bamboogle~\citep{press2023measuring} and BamTwoogle~\citep{aksitov2024rest} & Evaluation data & Not specified by public source \\
AntiLeakBench~\citep{wu-etal-2025-antileakbench} & Evaluation data & Not specified by public source \\
BFCL~\citep{patil2025the} & Tool-use evaluation data & Apache-2.0 \\
Qwen2.5 / Qwen3 models~\citep{qwen2025qwen25technicalreport,yang2025qwen3} & Base language agents & Apache-2.0 \\
Llama-3.2-1B~\citep{grattafiori2024llama} & Base language agent & Llama 3.2 Community License \\
xLAM-2-1b-fc-r~\citep{prabhakar2026apigenmt} & Base tool-use agent & Llama 3 Community License \\
\texttt{E5-base-v2}~\citep{wang2022text} & Dense retriever encoder & MIT \\
FAISS~\citep{johnson2019billion} & Vector index and retrieval & MIT \\
FlashRAG~\citep{10.1145/3701716.3715313} & Evaluation infrastructure & MIT \\
\bottomrule
\end{tabular}
\caption{Existing assets used in the experiments and their public licenses or terms.}
\label{tab:asset_licenses}
\end{table*}

% \paragraph{Evaluation}
% FlashRAG based (which Search-R1 and Tree-GRPO use) pipeline

% Retriever and corpus:
% - default: Search-R1 retriever (E5-base-v2 + FAISS), `data/wiki/wiki-18.jsonl` corpus
% - Antileakbench: to match the 2024 subset, build the 2024 version index using wiki-24 dump, then use the same Search-R1 retriever.

% \subsection{\ours}
\subsection{Expert Trajectory Seeding}
\label{appendix:fewshot_warm_start}
To build the expert trajectory set, we use the HotpotQA trajectory annotations from AutoCoA~\citep{zhang2025agent}, where the expert policy $\pi^*$ is DeepSeek-R1-Distill-Qwen-32B~\citep{guo2025deepseek}. We randomly selected $k{=}4$ trajectories returning correct answers as the set $X_{\text{exp}}$. The selected examples are shown in Appendix~\ref{appendix:fewshot_examples}.
For efficiency, we skip prefix continuation in the seeding phase.

\paragraph{\edit{Sensitivity to the Expert Trajectory Set.}}
\edit{To check whether the results depend on the specific four trajectories, we sample two additional sets of four expert trajectories independently and retrain \ours with the same setup. Table~\ref{tab:expert_trajectory_sets} shows average EM of 22.0, 24.3, and 22.3 for the three sets (22.9$\pm$1.3). Every set outperforms Search-R1-v0.3 on all six benchmarks. The largest variation is on Bamboogle and Antileak-m, which contain 125 and 455 questions.}
\begin{table*}[t]
\centering
\setlength{\tabcolsep}{4pt}
\scriptsize
\begin{adjustbox}{max width=\textwidth}
\begin{tabular}{l c c c c c >{\columncolor{cleanColor}}c c}
\toprule
\edit{\textbf{Method}} & \edit{\textbf{HotpotQA}} & \edit{\textbf{2wiki}} & \edit{\textbf{Musique}} & \edit{\textbf{Bamboogle}} & \edit{\textbf{Bamtwoogle}} & \cellcolor{cleanColor}\edit{\textbf{Antileak-m}} & \edit{\textbf{Avg.}} \\
\midrule
\edit{Search-R1-v0.3} & \edit{19.0} & \edit{21.3} & \edit{3.6} & \phantom{0}\edit{8.0} & \phantom{0}\edit{3.0} & \edit{16.7} & \edit{11.9} \\
\rowcolor{pennyColor!50}
\edit{\textbf{\ours} (set 1)} & \edit{24.8} & \edit{\textbf{27.2}} & \edit{7.1} & \edit{14.4} & \edit{22.0} & \edit{36.3} & \edit{22.0} \\
\rowcolor{pennyColor!50}
\edit{\textbf{\ours} (set 2)} & \edit{\textbf{26.0}} & \edit{26.9} & \edit{7.7} & \edit{\textbf{22.4}} & \edit{\textbf{23.0}} & \edit{\textbf{39.6}} & \edit{\textbf{24.3}} \\
\rowcolor{pennyColor!50}
\edit{\textbf{\ours} (set 3)} & \edit{24.9} & \edit{26.4} & \edit{\textbf{8.5}} & \edit{\textbf{22.4}} & \edit{20.0} & \edit{31.4} & \edit{22.3} \\
\midrule
\edit{Mean $\pm$ std.} & \edit{$25.2{\pm}0.7$} & \edit{$26.8{\pm}0.4$} & \edit{$7.8{\pm}0.7$} & \edit{$19.7{\pm}4.6$} & \edit{$21.7{\pm}1.5$} & \edit{$35.8{\pm}4.1$} & \edit{$22.9{\pm}1.3$} \\
\bottomrule
\end{tabular}
\end{adjustbox}
\caption{\edit{{Sensitivity to the expert trajectory set.} EM (\%) of \ours trained with three independently sampled sets of four expert trajectories on Qwen2.5-1.5B, with the best low-budget baseline for reference. Set 1 is the default set used in the main experiments.}}
\label{tab:expert_trajectory_sets}
\end{table*}

For the expert trajectory ratio, we keep the standard policy ratio $\pi_\theta(\tau^*)/\pi_{\theta_{old}}(\tau^*)$ and set the additional expert-policy correction $\omega^*(\theta){=}1$.
Table~\ref{tab:importance_ratio_ablation} compares different choices of $\omega^*(\theta)$ in expert trajectory seeding.
The exact trajectory-level correction yields 0 EM because the policy early in training assigns near-zero probability to expert trajectories, and the action-level correction gives lower EM than the constant correction.
\begin{table*}[t]
\centering
\setlength{\tabcolsep}{3pt}
\begin{adjustbox}{max width=\textwidth}
\scriptsize
\begin{tabular}{l c c c c c >{\columncolor{cleanColor}}c c}
\toprule
\textbf{$\omega^*(\theta)$} & \textbf{HotpotQA} & \textbf{2wiki} & \textbf{Musique} & \textbf{Bamboogle} & \textbf{Bamtwoogle} & \cellcolor{cleanColor}\textbf{Antileak-m} & \textbf{Avg.} \\
\midrule
$\pi_{\theta}(\tau^*) / \pi^*(\tau^*)$ & 0.0 & 0.0 & 0.0 & 0.0 & 0.0 & 0.0 & 0.0 \\
$\pi_{\theta}(a_t^* \mid s_t^*) / \pi^*(a_t^* \mid s_t^*)$ & 14.6 & 18.1 & 3.1 & 10.4 & 9.0 & 24.4 & 13.3 \\
\rowcolor{pennyColor!50}
$1$ & \textbf{24.8} & \textbf{27.2} & \textbf{7.1} & \textbf{14.4} & \textbf{22.0} & \textbf{36.3} & \textbf{22.0} \\
\bottomrule
\end{tabular}
\end{adjustbox}
\caption{{Off-policy correction ablation.} EM results for Qwen2.5-1.5B with different choices of the expert-policy correction $\omega^*(\theta)$ in expert trajectory seeding.}
\label{tab:importance_ratio_ablation}
\end{table*}

Table~\ref{tab:annotation_budget} compares \ours under different annotation budgets.
Increasing the number of expert trajectories from $k{=}4$ to $k{=}96$ improves average EM from 22.0 to 24.2, while using the full AutoCoA trajectory set with SFT initialization ($k{=}5{,}276$) further improves average EM to 30.8.
We keep $k{=}4$ as the default setting because the main goal is to test whether expert trajectory seeding can stabilize low-budget RL with only a handful of expert trajectories.
\begin{table*}[t]
\centering
\setlength{\tabcolsep}{4pt}
\scriptsize
\begin{adjustbox}{max width=\textwidth}
\begin{tabular}{l r c c c c c >{\columncolor{cleanColor}}c c}
\toprule
\textbf{Setting} & \textbf{$K$} & \textbf{HotpotQA} & \textbf{2wiki} & \textbf{Musique} & \textbf{Bamboogle} & \textbf{Bamtwoogle} & \cellcolor{cleanColor}\textbf{Antileak-m} & \textbf{Avg.} \\
\midrule
\ours & 4 & 24.8 & \textbf{27.2} & 7.1 & 14.4 & 22.0 & 36.3 & 22.0 \\
\ours & 96 & 27.6 & 26.9 & 8.6 & 20.8 & 22.0 & 39.1 & 24.2 \\
\ours with Full SFT & 5,276 & \textbf{32.6} & \textbf{28.1} & \textbf{12.2} & \textbf{33.6} & \textbf{32.0} & \textbf{46.2} & \textbf{30.8} \\
\bottomrule
\end{tabular}
\end{adjustbox}
\caption{{Effect of annotation budget.} EM results on Qwen2.5-1.5B with different numbers of expert trajectory annotations.}
\label{tab:annotation_budget}
\end{table*}

\paragraph{\edit{Controlled RL Variants.}}
\label{appendix:expert_bootstrapped_rl_variants}
\edit{To isolate the contribution beyond the expert trajectory seeding component of \ours, we retain this component and replace the subsequent RL method with vanilla GRPO~\citep{shao2024deepseekmath}, StepSearch~\citep{zheng-etal-2025-stepsearch}, Search-R1-v0.3~\citep{jin2025an}, or Tree-GRPO~\citep{ji2025treesearch}. All variants use the same four expert trajectories and low-budget setup. Table~\ref{tab:expert_bootstrapped_rl_variants} shows that \ours achieves the highest average EM of 22.0, exceeding the Tree-GRPO variant by 1.6 points. Access to the expert trajectories alone does not explain the gain.}
\begin{table*}[t]
\centering
\scriptsize
\setlength{\tabcolsep}{3.5pt}
\begin{adjustbox}{max width=\textwidth}
\begin{tabular}{l c c c c c >{\columncolor{cleanColor}}c c}
\toprule
\edit{\textbf{Method}} & \edit{\textbf{HotpotQA}} & \edit{\textbf{2wiki}} & \edit{\textbf{Musique}} & \edit{\textbf{Bamboogle}} & \edit{\textbf{Bamtwoogle}} & \cellcolor{cleanColor}\edit{\textbf{Antileak-m}} & \edit{\textbf{Avg.}} \\
\midrule
\edit{{Expert trajectory seeding + Vanilla GRPO}} & \edit{20.8} & \edit{22.0} & \edit{4.5} & \edit{10.4} & \edit{10.0} & \edit{31.4} & \edit{16.5} \\
\edit{{Expert trajectory seeding + StepSearch}} & \edit{20.4} & \edit{19.8} & \edit{3.6} & \phantom{0}\edit{9.6} & \phantom{0}\edit{9.0} & \edit{19.1} & \edit{13.6} \\
\edit{{Expert trajectory seeding + Search-R1-v0.3}} & \edit{22.7} & \edit{23.3} & \edit{6.6} & \edit{\underline{13.6}} & \edit{19.0} & \edit{28.8} & \edit{19.0} \\
\edit{{Expert trajectory seeding + Tree-GRPO}} & \edit{\underline{23.6}} & \edit{\underline{23.6}} & \edit{\textbf{7.5}} & \edit{12.8} & \edit{\textbf{22.0}} & \edit{\underline{33.2}} & \edit{\underline{20.4}} \\
\rowcolor{pennyColor!50}
\edit{\textbf{\ours}} & \edit{\textbf{24.8}} & \edit{\textbf{27.2}} & \edit{\underline{7.1}} & \edit{\textbf{14.4}} & \edit{\textbf{22.0}} & \edit{\textbf{36.3}} & \edit{\textbf{22.0}} \\
\bottomrule
\end{tabular}
\end{adjustbox}
\caption{\edit{EM (\%) for controlled variants of Qwen2.5-1.5B that retain the expert trajectory seeding component of \ours. All variants use the same four expert trajectories and low-budget setup. Best results are \textbf{bolded}; second-best results are \underline{underlined}.}}
\label{tab:expert_bootstrapped_rl_variants}
\end{table*}

\subsection{Evidence-Coverage Reward}
\label{appendix:grounded_retrieval_reward}

Table~\ref{tab:retrieval_reward_comparison_full} compares the proposed \textit{evidence-coverage reward} ($r_g$) with retrieval rewards from prior work.
Starting from the same expert trajectory seeding checkpoint, we train with three reward signals while keeping the other settings fixed.
\ours (Coverage) gives the best results on most benchmarks.
Search-R1-v0.3, which rewards answer-document retrieval, is second, while StepSearch, which uses step-wise lexical similarity, is lower in this setup.
On Antileak-m, \ours improves over Search-R1-v0.3 by 4.4 pp in EM and 5.4 pp in F1.
\begin{table*}[t]
\centering
\setlength{\tabcolsep}{3pt}

% 테이블 너비를 페이지 너비에 맞게 자동 조절
\begin{adjustbox}{max width=\textwidth}
\scriptsize
\begin{tabular}{l cc cc cc cc cc >{\columncolor{cleanColor}}c >{\columncolor{cleanColor}}c cc}
\toprule
\multirow{2.5}{*}{\textbf{Retrieval Reward}} 
& \multicolumn{2}{c}{\textbf{HotpotQA}} 
& \multicolumn{2}{c}{\textbf{2wiki}} 
& \multicolumn{2}{c}{\textbf{Musique}} 
& \multicolumn{2}{c}{\textbf{Bamboogle}} 
& \multicolumn{2}{c}{\textbf{Bamtwoogle}} 
& \multicolumn{2}{c}{\cellcolor{cleanColor}\textbf{Antileak-m}}
& \multicolumn{2}{c}{\textbf{Avg.}} \\
\cmidrule(lr){2-3} \cmidrule(lr){4-5} \cmidrule(lr){6-7} \cmidrule(lr){8-9} \cmidrule(lr){10-11} \cmidrule(lr){12-13} \cmidrule(lr){14-15}
 & \textbf{EM} & \textbf{F1} 
 & \textbf{EM} & \textbf{F1} 
 & \textbf{EM} & \textbf{F1} 
 & \textbf{EM} & \textbf{F1} 
 & \textbf{EM} & \textbf{F1} 
 & \textbf{EM} & \textbf{F1}
 & \textbf{EM} & \textbf{F1} \\
\midrule

% --- StepSearch ---
StepSearch
& 20.4 & 29.0
& 19.8 & 24.8
& \phantom{0}3.6 & \phantom{0}8.2
& \phantom{0}9.6 & 18.2
& \phantom{0}9.0 & 12.0
& 19.1 & 23.1
& 13.6 & 19.2 \\

% --- Search-R1 ---
Search-R1-v0.3
& \underline{22.7} & \underline{31.0}
& \underline{23.3} & \underline{28.0}
& \phantom{0}\textbf{6.6} & \textbf{11.4}
& \underline{13.6} & \textbf{22.8}
& \textbf{19.0} & \underline{23.4}
& \underline{28.8} & \underline{32.8}
& \underline{19.0} & \underline{24.9} \\

% --- Ours ---
% \rowcolor{pennyColor!50}
\rowcolor{abl1Color!45}
\textbf{\ours (Coverage)}
& \textbf{23.3} & \textbf{31.9}
& \textbf{24.4} & \textbf{30.0}
& \phantom{0}\underline{5.8} & \underline{11.3}
& \textbf{14.4} & \underline{20.9}
& \textbf{19.0} & \textbf{23.8}
& \textbf{33.2} & \textbf{38.2}
& \textbf{20.0} & \textbf{26.0} \\

\bottomrule
\end{tabular}
\end{adjustbox}
\caption{{Comparison of Retrieval Rewards.} Multi-hop QA results trained on Qwen2.5-1.5B.}
\label{tab:retrieval_reward_comparison_full}
\end{table*}

\edit{To examine whether explicitly accounting for retrieval precision improves the evidence-coverage reward, we replace our recall-oriented $r_g$ with an F1-oriented variant that equally weights the precision and recall of $\hat{\mathcal{D}}_{union}$ against $D$. As shown in Table~\ref{tab:f1_reward}, the F1-oriented reward reduces EM on all six benchmarks. Table~\ref{tab:f1_reward_behavior} explains this drop. Incorporating precision weakens the reward for additional searches because they may retrieve documents outside $D$, even when required evidence remains missing. This suppresses retrieval, reducing the average number of searches from 3.05 to 1.23 on HotpotQA and from 3.48 to 1.49 on MuSiQue, and consequently lowers both document recall metrics and QA performance (see also Table~\ref{tab:retrieval_behavior}).}
\begin{table*}[t]
\centering
\setlength{\tabcolsep}{4pt}
\scriptsize
\begin{adjustbox}{max width=\textwidth}
\begin{tabular}{l c c c c c >{\columncolor{cleanColor}}c c}
\toprule
\textbf{Retrieval reward} & \textbf{HotpotQA} & \textbf{2wiki} & \textbf{Musique} & \textbf{Bamboogle} & \textbf{Bamtwoogle} & \cellcolor{cleanColor}\textbf{Antileak-m} & \textbf{Avg.} \\
\midrule
\rowcolor{pennyColor!50}
\textbf{Recall-oriented (\ours)} & \textbf{24.8} & \textbf{27.2} & \textbf{7.1} & \textbf{14.4} & \textbf{22.0} & \textbf{36.3} & \textbf{22.0} \\
F1-oriented & 19.6 & 17.7 & 4.8 & \phantom{0}8.8 & 14.0 & 24.2 & 14.8 \\
\bottomrule
\end{tabular}
\end{adjustbox}
\caption{\edit{{Recall- and F1-oriented retrieval rewards.} EM (\%) when training with the recall-oriented evidence-coverage reward of \ours and with an F1-oriented variant that equally weights precision and recall. The backbone model is Qwen2.5-1.5B.}}
\label{tab:f1_reward}
\end{table*}

\begin{table*}[t]
\centering
\scriptsize
\begin{adjustbox}{max width=\textwidth}
\begin{tabular}{l l c c c c c c}
\toprule
\textbf{Dataset} & \textbf{Retrieval reward} & \textbf{Search} & \textbf{Prec.} & \textbf{Any} & \textbf{All} & \textbf{Unnec.} & \textbf{EM} \\
\midrule
\multirow{2}{*}{HotpotQA} & Recall-oriented (\ours) & 3.05 & 28.2 & \textbf{52.3} & \textbf{31.6} & 9.8 & \textbf{24.8} \\
 & F1-oriented & 1.23 & \textbf{39.1} & 41.5 & 15.1 & 1.4 & 19.6 \\
\midrule
\multirow{2}{*}{MuSiQue} & Recall-oriented (\ours) & 3.48 & 14.3 & \textbf{25.9} & \textbf{8.2} & 2.6 & \textbf{7.1} \\
 & F1-oriented & 1.49 & \textbf{18.0} & 18.4 & \phantom{0}2.8 & 0.4 & \phantom{0}4.8 \\
\bottomrule
\end{tabular}
\end{adjustbox}
\caption{\edit{{Retrieval behavior of recall- and F1-oriented rewards.} Retrieval statistics and EM (\%) on HotpotQA and MuSiQue with the Qwen2.5-1.5B backbone. The metrics follow Table~\ref{tab:retrieval_behavior}.}}
\label{tab:f1_reward_behavior}
\end{table*}

\edit{A related concern is the case where the retriever fails to return any document of the support set, since $r_g$ depends on exact membership in $D$. We measured this over a full training run of \ours. Only 0.88\% of the sampled groups contained no trajectory that retrieved a support document. In the remaining 99.12\%, at least one trajectory retrieved a support document, so the group-relative advantage carried an evidence signal. In the rare all-miss groups, $r_g$ is zero for every trajectory and the update falls back to the outcome reward.}

We set $\lambda{=}0.5$ in the main experiments, giving equal weight to the evidence-coverage reward $r_g$ and the outcome reward $r_o$. Table~\ref{tab:lambda_ablation} varies $\lambda$; although the best value differs by dataset, $\lambda{=}0.5$ achieves the highest average EM among the tested values.
\begin{table}[H]
\centering
\setlength{\tabcolsep}{3pt}
\begin{adjustbox}{max width=\columnwidth}
\scriptsize
\begin{tabular}{c c c c c c >{\columncolor{cleanColor}}c c}
\toprule
$\boldsymbol{\lambda}$ & \textbf{HotpotQA} & \textbf{2wiki} & \textbf{Musique} & \textbf{Bamboogle} & \textbf{Bamtwoogle} & \cellcolor{cleanColor}\textbf{Antileak-m} & \textbf{Avg.} \\
\midrule
\texttt{0.25} & 21.4 & 23.2 & 6.1 & 16.8 & 18.0 & 32.3 & 19.6 \\
\rowcolor{pennyColor!50}
\texttt{0.50} & \textbf{24.8} & \textbf{27.2} & \textbf{7.1} & 14.4 & \textbf{22.0} & \textbf{36.3} & \textbf{22.0} \\
\texttt{0.75} & 21.3 & 24.1 & 5.5 & \textbf{17.6} & 18.0 & 35.2 & 20.3 \\
\bottomrule
\end{tabular}
\end{adjustbox}
\caption{EM results for Qwen2.5-1.5B with different weights on the evidence-coverage reward.}
\label{tab:lambda_ablation}
\end{table}

\subsection{Prefix Continuation}
To efficiently generate the refined trajectory $\tau'_M$, we employ a rejection resampling strategy.
Given the truncated history $\tau_{M, \le t'}$ (where $t'$ is the last step with valid retrieval), we sample $l$ independent candidate completions $\{\tilde{\tau}^{(1)}, \dots, \tilde{\tau}^{(l)}\}$ from the policy $\pi_\theta$ using temperature sampling.
We then evaluate the total reward for each completion and select the one with the highest score as the final continued trajectory $\tau'_M$:
\begin{equation}
    \tau'_M = \operatorname*{arg\,max}_{\tilde{\tau} \in \{\tilde{\tau}^{(1)}, \dots, \tilde{\tau}^{(l)}\}} R(\tilde{\tau}).
\end{equation}
In our experiments, we set the number of continuation samples $l{=}5$.
% This selection process ensures that we inject the most plausible reasoning path into the batch.
% Finally, if the selected $\tau'_M$ yields a higher reward than the original best $\tau_M$, we replace the worst-performing trajectory $\tau_m$ in the current group with $\tau'_M$, thereby artificially increasing the density of high-reward trajectories without additional full rollouts.

\subsection{\edit{Details of Corpus-Graph Evidence Construction}}
\label{appendix:evidence_annotation_requirement}

\edit{\S\ref{sec:evidence_construction} summarizes the construction pipeline. Following the graph-based construction procedure of HotpotQA~\citep{yang2018hotpotqa}, we use the Wikipedia dump released with HotpotQA as follows:}
\begin{enumerate}[leftmargin=*,nosep]
    \item \edit{Build a hyperlink graph from the Wikipedia documents.}
    \item \edit{Sample subgraphs according to the target hop distribution. Among the 4{,}000 examples, 55\%, 44\%, 0.5\%, and 0.5\% are 2-, 3-, 4-, and 5-hop, respectively.}
    \item \edit{Extract entities and inter-document relations from each subgraph using the two-shot prompt in Appendix~\ref{appendix:synthetic_prompts}. We generate a question--answer pair from the leaf document and compose the multi-hop question backward using the one-shot prompt. We use Qwen2.5-3B~\citep{qwen2025qwen25technicalreport}, the smallest model that produced stable generations. Qwen2.5-1.5B produced degenerate outputs.}
    \item \edit{Retain an example only if (i) it follows the required format, (ii) its support documents do not occur in the top-20 BM25 results for the initial question, (iii) the answer cannot be produced without the support documents, (iv) no proper subset of the support documents is sufficient, and (v) the generated answer is correct.}
\end{enumerate}
\edit{The support set $D$ is the document chain of the sampled subgraph. The pipeline requires the source corpus but does not use manually annotated support sets.}

\clearpage
% Prompts, case studies, and few-shot examples are typeset in a single column
% so that the long verbatim boxes stay readable. These are the last appendix sections.
\onecolumn
\section{Prompts}
\label{appendix:prompt}
% \subsection{\ours}
\begin{fewshotbox}{\ours}
    % Metadata: ID 삭제, Level과 Source만 배치
    \small
    % Gold Response Section (연보라색 통일)
    \textbf{\large [Prompt]} \\
\begin{lstlisting}
You are an intelligent reasoning assistant with access to a web search tool to help you obtain accurate information. When answering the user's questions, please adhere to the following guidelines:

1. **Uncertainty Handling:**
   - Encourage the use of web search to solve the problems mentioned during reasoning.
   - If you are uncertain or do not have sufficient knowledge about a specific detail, you **must** perform a web search.
   - To perform a web search the format **must** be in the format mentioned in point 2 Web Search Format
   - You can perform multiple web searches to ensure that the answer is correct.
   - The web search should be initiated only within your internal reasoning (the "thinking" phase), and it should not appear in your final answer to the user.

2. **Web Search Format:**
   - When calling the web search tool, use the following exact format:
     <begin_search> your query here </end_search>
   - After you perform a web search, you will receive results enclosed in:
     <search_result> ... </search_result>
   - You can then use this information to further refine your reasoning and answer.

3. **Process:**
   - If you encounter any uncertainty or unknown knowledge, embed a web search query within your internal thoughts (surrounded by <think></think> tags) using the specified format.
   - Incorporate the search results into your reasoning before finalizing your answer.
   - You should provide your final answer in the format \boxed{YOUR_ANSWER}.

Now, please answer the user's question below:
{question}
\end{lstlisting}
\end{fewshotbox}
\begin{fewshotbox}{LLM-as-Judge}
    \small
    \textbf{\large [Prompt]} \\
\begin{lstlisting}
Given a Question and its Golden Answer, verify whether the Predicted Answer is correct. The prediction is correct if it fully aligns with the meaning and key information of the Golden Answer. Respond with True if the prediction is correct and False otherwise.

Question: {question}
Golden Answer: {gold_answer}
Predicted Answer: {predicted_answer}
\end{lstlisting}
\end{fewshotbox}

% \subsection{Tree-GRPO}
\subsection{\edit{Synthetic Data Generation}}
\label{appendix:synthetic_prompts}
\edit{Prompt A extracts a referring expression for an inter-document relation, and Prompt B generates a leaf-document question--answer pair. The referring expressions are substituted backward along the sampled chain to form a multi-hop question.}
\begin{fewshotbox}{\edit{Prompt A: Entity and Relation Extraction}}
\small
\edit{\textbf{[Prompt]}}\\
\begingroup
\lstset{basicstyle=\ttfamily\small\color{blue}}
\renewcommand{\lstlinecolor}{\color{blue}}
\begin{lstlisting}
System: "You rewrite Wikipedia facts into short referring expressions. You never add information that is not in the given paragraph."

Write a short noun phrase that uniquely identifies the target entity using ONLY facts stated in the paragraph.

Rules:
- Write the literal token <SOURCE> wherever the paragraph's title belongs. Never spell the title out.
- The phrase MUST NOT contain the target entity's name or any word from it.
- Use a definite noun phrase: "the director of <SOURCE>", "the town where <SOURCE> was founded".
- One line, no quotes, no explanation.

### Example 1
Paragraph (title: One Night Stand (1984 film)):
One Night Stand is a 1984 film directed by John Duigan.
Target entity: John Duigan
Phrase: the director of <SOURCE>

### Example 2
Paragraph (title: Then We Came to the End):
Then We Came to the End is a 2007 novel by Joshua Ferris, published by Little, Brown and Company.
Target entity: Little, Brown and Company
Phrase: the publisher of <SOURCE>

### Response
Paragraph (title: {src_title}): {src_text}
Target entity: {dst_title}
Phrase:
\end{lstlisting}
\endgroup
\end{fewshotbox}

\begin{fewshotbox}{\edit{Prompt B: Question and Answer Generation}}
\small
\edit{\textbf{[Prompt]}}\\
\begingroup
\lstset{basicstyle=\ttfamily\small\color{blue}}
\renewcommand{\lstlinecolor}{\color{blue}}
\begin{lstlisting}
System: "You write single-hop factual questions grounded in one Wikipedia paragraph. You copy answers verbatim from the paragraph."

Write ONE short factual question about the given title that the paragraph answers.

Rules:
- Write the literal token <TITLE> wherever the title belongs. Never spell it out.
- The answer MUST be COPIED CHARACTER FOR CHARACTER from the paragraph, at most 5 words.
- Prefer a name, a date, or a number as the answer -- not a descriptive phrase.
- The answer MUST NOT appear anywhere in the question.
- The question's wh-word must match the answer: "How many"/"How long" need a number, "When" needs a date, "Who" needs a person.
- Ask about an attribute, never about the title's own name.

### Example
Paragraph (title: Little, Brown and Company):
Little, Brown and Company is an American publisher founded in 1837 by Charles Little and James Brown.
{"question": "In what year was <TITLE> founded?", "answer": "1837", "answer_type": "date"}

### Response
Paragraph (title: {title}): {text}
Return JSON only:
{"question": "...", "answer": "...", "answer_type": "entity|date|number|span"}
\end{lstlisting}
\endgroup
\end{fewshotbox}

% \input{table/prompt_tree_grpo}
% \subsection{StepSearch}
% \input{table/prompt_stepsearch}
% \subsection{Search-R1-v0.3}
% \input{table/prompt_searchr1}

\clearpage
\section{Case Study}

\subsection{\ours}
\begin{fewshotbox}{\ours~{\color{my_green}{(Correct)}}}
    % Metadata: ID 삭제, Level과 Source만 배치
    \small
    \begin{tabularx}{\textwidth}{@{}l X  l X l X@{}}
        \textbf{Level:} & Hard & \textbf{Source:} & HotpotQA & \textbf{Reasoning Type:} Comparison\\
    \end{tabularx}
    \vspace{0.01mm}
    \hrule
    \vspace{4mm}

    % Prompt Section
    \textbf{\large [Question]} \\
\begin{lstlisting}
Are the Laleli Mosque and Esma Sultan Mansion located in the same neighborhood?
\end{lstlisting}  
    \vspace{6mm}

    \textbf{\large [Gold Answer]} \\
\begin{lstlisting}
No
\end{lstlisting}  
    \vspace{6mm}

    \textbf{\large [Titles of Support Evidence Documents ($D$)]} \\
\begin{lstlisting}
Laleli Mosque, Esma Sultan Mansion
\end{lstlisting}  
    \vspace{6mm}

    % Gold Response Section (연보라색 통일)
    \textbf{\large [Response]} \\
\begin{lstlisting}
<think>
Okay, I need to determine if the Laleli Mosque and Esma Sultan Mansion are located in the same neighborhood. Let me start by recalling what I know about these two landmarks.

First, I remember that I learned in history class that the Laleli Mosque is an important Islamic structure in Istanbul. I'll try to remember if it's part of a larger complex or if it's located in a specific neighborhood. I'll search for "Laleli Mosque location Istanbul" to get the basic information.

<begin_search>Laleli Mosque location Istanbul</end_search>

<search_result> result 1: "Laleli Mosque"
Laleli Mosque The Laleli Mosque () is an 18th-century Ottoman imperial mosque located in Laleli, Fatih, Istanbul, Turkey. The Laleli Mosque was built by Sultan Mustafa III from 1760–1763, designed in the baroque style by Ottoman imperial architect Mehmet Tahir Ağa. The complex was destroyed by a fire in 1783 shortly after its completion and was immediately rebuilt. A fire in 1911 destroyed the madrasah, and subsequent road construction work destroyed many other auxiliary structures to the mosque. The mosque was built on a high terrace over a complex of vaulted shops, whose rents were intended to financially support the
result 2: "Tekeli Mehmet Pasha Mosque"
Tekeli Mehmet Pasha Mosque The Tekeli Mehmet Paşa Mosque () is a mosque in the city of Antalya, Turkey. Mosque takes its name from Lala Mehmed Pasa. Built in the 18th century in the Kalekapisi district, the mosque is one of the most important Ottoman mosques in the city. The main dome, which rises on a high rim, is supported by three semi domes, one each in the east, west, and south directions, as well as three domes on the northern side. There are tiled panels decorated with Koranic verses in taliq script on the pointed arch-formed lunettes of the
result 3: "Tekeli Mehmet Pasha Mosque"
windows on the northern façade of the mosque and inside it. Tekeli Mehmet Pasha Mosque The Tekeli Mehmet Paşa Mosque () is a mosque in the city of Antalya, Turkey. Mosque takes its name from Lala Mehmed Pasa. Built in the 18th century in the Kalekapisi district, the mosque is one of the most important Ottoman mosques in the city. The main dome, which rises on a high rim, is supported by three semi domes, one each in the east, west, and south directions, as well as three domes on the northern side. There are tiled panels decorated with Koranic </search_result>



Alright, the first result clearly mentions that the Laleli Mosque is located in Laleli, Fatih, Istanbul, Turkey. Now, I need to find out about the Esma Sultan Mansion to see if they are in the same neighborhood.

I'll search for "Esma Sultan Mansion location Istanbul" to get that information.

<begin_search>Esma Sultan Mansion location Istanbul</end_search>

<search_result> result 1: "Esma Sultan Mansion"
Esma Sultan Mansion The Esma Sultan Mansion (), a historical yalı () located at Bosphorus in Ortaköy neighborhood of Istanbul, Turkey and named after its original owner Esma Sultan, is used today as a cultural center after being redeveloped. The three-storey brick building was designed by the renowned architect Sarkis Balyan and finished in 1875 next to Ortaköy Mosque. It was presented to Esma Sultan, the daughter of Ottoman Sultan Abdulaziz, as a wedding gift. The mansion remained in the possession of the Ottoman dynasty until 1915. The building was subsequently used first as a tobacco warehouse and then as
result 2: "Esma Sultan Mansion"
guests in the ground floor and 600 guests in the first floor. The venue also hosts concerts of the Istanbul International Jazz Festival and the Istanbul International Music Festival. Esma Sultan Mansion The Esma Sultan Mansion (), a historical yalı () located at Bosphorus in Ortaköy neighborhood of Istanbul, Turkey and named after its original owner Esma Sultan, is used today as a cultural center after being redeveloped. The three-storey brick building was designed by the renowned architect Sarkis Balyan and finished in 1875 next to Ortaköy Mosque. It was presented to Esma Sultan, the daughter of Ottoman Sultan Abdulaziz,
result 3: "Esma Sultan Mansion"
a coal depot from 1920 until 1975, when it was destroyed by a fire. The ruin, consisting only of the outer walls of the building, was purchased in the early 1990s by The Marmara hotel chain. Following a renovation with additions designed by architects Haluk Sezgin and Philippe Robert, the mansion was opened in 2001 as a multipurpose event venue. Within the brick exterior, which was left as the original, a steel and glass structure is incorporated. According to the GAD Architecture group, a redesign by the architect Gökhan Avcıoğlu was completed in 2005. The building includes a bar, a </search_result>



From the first result, the Esma Sultan Mansion is located at Ortaköy neighborhood of Istanbul. Now, I need to check if Laleli is part of the same neighborhood. I'll search for "Laleli neighborhood Istanbul" to confirm its location.

<begin_search>Laleli neighborhood Istanbul</end_search>

<search_result> result 1: "Laleli Mosque"
Laleli Mosque The Laleli Mosque () is an 18th-century Ottoman imperial mosque located in Laleli, Fatih, Istanbul, Turkey. The Laleli Mosque was built by Sultan Mustafa III from 1760–1763, designed in the baroque style by Ottoman imperial architect Mehmet Tahir Ağa. The complex was destroyed by a fire in 1783 shortly after its completion and was immediately rebuilt. A fire in 1911 destroyed the madrasah, and subsequent road construction work destroyed many other auxiliary structures to the mosque. The mosque was built on a high terrace over a complex of vaulted shops, whose rents were intended to financially support the
result 2: Etiler
Etiler Etiler is a neighbourhood on the European side of Istanbul, Turkey, in the district of Beşiktaş, close to the business quarters of Levent and Maslak. Etiler is famous for its upmarket cafés, pubs, night clubs, restaurants, gyms, coiffeurs, fashion shops and shopping malls, such as Akmerkez. It is a favourite area among Istanbul's elite. The quarter also has many villas and private residences. The name ""Etiler"" is an older name in Turkish for the ""Hittites"", as it was fashionable in the early years of the Turkish Republic to give the names of ancient Anatolian civilizations to the new districts
result 3: "Şişli"
Şişli Şişli () is one of 39 districts of Istanbul, Turkey. Located on the European side of the city, it is bordered by Beşiktaş to the east, Sarıyer to the north, Eyüp and Kağıthane to the west, and Beyoğlu to the south. In 2009, Şişli had a population of 316,058. Until the 1800s, Şişli was open countryside, used for hunting and agriculture and leisure. It was developed as a middle class residential district during the last years of the Ottoman Empire and the early years of the Turkish Republic (the late 19th-early 20th centuries). French culture was an important influence </search_result>



The first result clearly states that the Laleli Mosque is located in Laleli, Fatih, Istanbul, Turkey. It's located in the Fatih district, specifically in the Laleli neighborhood. Therefore, Laleli is part of the Fatih neighborhood, not Ortaköy.

Since the Esma Sultan Mansion is in Ortaköy, and the Laleli Mosque is in Laleli, they are indeed located in different neighborhoods.
</think>

The Laleli Mosque is in the Fatih neighborhood, and the Esma Sultan Mansion is in the Ortaköy neighborhood. Thus, they are not located in the same neighborhood.

\boxed{**No**}
\end{lstlisting}
\end{fewshotbox}

\subsection{Tree-GRPO}
\label{appendix:case_study_tree_grpo}
\begin{fewshotbox}{Tree-GRPO {\color{red}{(Incorrect)}}}
    % Metadata: ID 삭제, Level과 Source만 배치
    \small
    \begin{tabularx}{\textwidth}{@{}l X  l X l X@{}}
        \textbf{Level:} & Hard & \textbf{Source:} & HotpotQA & \textbf{Reasoning Type:} Comparison\\
    \end{tabularx}
    \vspace{0.01mm}
    \hrule
    \vspace{4mm}

    % Prompt Section
    \textbf{\large [Question]} \\
\begin{lstlisting}
Are the Laleli Mosque and Esma Sultan Mansion located in the same neighborhood?
\end{lstlisting}  
    \vspace{6mm}

    \textbf{\large [Gold Answer]} \\
\begin{lstlisting}
No
\end{lstlisting}  
    \vspace{6mm}

    \textbf{\large [Titles of Support Evidence Documents ($D$)]} \\
\begin{lstlisting}
Laleli Mosque, Esma Sultan Mansion
\end{lstlisting}  
    \vspace{6mm}

    % Gold Response Section (연보라색 통일)
    \textbf{\large [Response]} \\
\begin{lstlisting}
<think>
The Laleli Mosque is located in Istanbul, Turkey, which is the capital city of the country.
Esma Sultan Mansion is located in East Fourth district, Beyoglu neighborhood, Istanbul, Turkey.
Both are located in Istanbul, Turkey.
</think>
<answer>**Yes**</answer>
\end{lstlisting}
\end{fewshotbox}

\subsection{StepSearch}
\begin{fewshotbox}{StepSearch {\color{red}{(Incorrect)}}}
    % Metadata: ID 삭제, Level과 Source만 배치
    \small
    \begin{tabularx}{\textwidth}{@{}l X  l X l X@{}}
        \textbf{Level:} & Hard & \textbf{Source:} & HotpotQA & \textbf{Reasoning Type:} Comparison\\
    \end{tabularx}
    \vspace{0.01mm}
    \hrule
    \vspace{4mm}

    % Prompt Section
    \textbf{\large [Question]} \\
\begin{lstlisting}
Are the Laleli Mosque and Esma Sultan Mansion located in the same neighborhood?
\end{lstlisting}  
    \vspace{6mm}

    \textbf{\large [Gold Answer]} \\
\begin{lstlisting}
No
\end{lstlisting}  
    \vspace{6mm}

    \textbf{\large [Titles of Support Evidence Documents ($D$)]} \\
\begin{lstlisting}
Laleli Mosque, Esma Sultan Mansion
\end{lstlisting}  
    \vspace{6mm}

    % Gold Response Section (연보라색 통일)
    \textbf{\large [Response]} \\
\begin{lstlisting}
<answer> no. <
<warning> My previous action is invalid. If I want to search, I should put the query between <search> and </search>. If I want to give the final answer, I should put the answer between <answer> and </answer>. Let me try again. </warning>
<search> Are the Laleli Mosque and Esma Sultan Mansion located in the same neighborhood </search>

<information> Doc 1<## Title: "Esma Sultan Mansion" ##> Esma Sultan Mansion The Esma Sultan Mansion (), a historical yalı () located at Bosphorus in Ortaköy neighborhood of Istanbul, Turkey and named after its original owner Esma Sultan, is used today as a cultural center after being redeveloped. The three-storey brick building was designed by the renowned architect Sarkis Balyan and finished in 1875 next to Ortaköy Mosque. It was presented to Esma Sultan, the daughter of Ottoman Sultan Abdulaziz, as a wedding gift. The mansion remained in the possession of the Ottoman dynasty until 1915. The building was subsequently used first as a tobacco warehouse and then as
Doc 2<## Title: "Esma Sultan Mansion" ##> guests in the ground floor and 600 guests in the first floor. The venue also hosts concerts of the Istanbul International Jazz Festival and the Istanbul International Music Festival. Esma Sultan Mansion The Esma Sultan Mansion (), a historical yalı () located at Bosphorus in Ortaköy neighborhood of Istanbul, Turkey and named after its original owner Esma Sultan, is used today as a cultural center after being redeveloped. The three-storey brick building was designed by the renowned architect Sarkis Balyan and finished in 1875 next to Ortaköy Mosque. It was presented to Esma Sultan, the daughter of Ottoman Sultan Abdulaziz,
Doc 3<## Title: "Laleli Mosque" ##> Laleli Mosque The Laleli Mosque () is an 18th-century Ottoman imperial mosque located in Laleli, Fatih, Istanbul, Turkey. The Laleli Mosque was built by Sultan Mustafa III from 1760–1763, designed in the baroque style by Ottoman imperial architect Mehmet Tahir Ağa. The complex was destroyed by a fire in 1783 shortly after its completion and was immediately rebuilt. A fire in 1911 destroyed the madrasah, and subsequent road construction work destroyed many other auxiliary structures to the mosque. The mosque was built on a high terrace over a complex of vaulted shops, whose rents were intended to financially support the </information>

<answer> no. <
<warning> My previous action is invalid. If I want to search, I should put the query between <search> and </search>. If I want to give the final answer, I should put the answer between <answer> and </answer>. Let me try again. </warning>
<search> Are the Laleli Mosque and Esma Sultan Mansion located in the same neighborhood </search>

<information> Doc 1<## Title: "Esma Sultan Mansion" ##> Esma Sultan Mansion The Esma Sultan Mansion (), a historical yalı () located at Bosphorus in Ortaköy neighborhood of Istanbul, Turkey and named after its original owner Esma Sultan, is used today as a cultural center after being redeveloped. The three-storey brick building was designed by the renowned architect Sarkis Balyan and finished in 1875 next to Ortaköy Mosque. It was presented to Esma Sultan, the daughter of Ottoman Sultan Abdulaziz, as a wedding gift. The mansion remained in the possession of the Ottoman dynasty until 1915. The building was subsequently used first as a tobacco warehouse and then as
Doc 2<## Title: "Esma Sultan Mansion" ##> guests in the ground floor and 600 guests in the first floor. The venue also hosts concerts of the Istanbul International Jazz Festival and the Istanbul International Music Festival. Esma Sultan Mansion The Esma Sultan Mansion (), a historical yalı () located at Bosphorus in Ortaköy neighborhood of Istanbul, Turkey and named after its original owner Esma Sultan, is used today as a cultural center after being redeveloped. The three-storey brick building was designed by the renowned architect Sarkis Balyan and finished in 1875 next to Ortaköy Mosque. It was presented to Esma Sultan, the daughter of Ottoman Sultan Abdulaziz,
Doc 3<## Title: "Laleli Mosque" ##> Laleli Mosque The Laleli Mosque () is an 18th-century Ottoman imperial mosque located in Laleli, Fatih, Istanbul, Turkey. The Laleli Mosque was built by Sultan Mustafa III from 1760–1763, designed in the baroque style by Ottoman imperial architect Mehmet Tahir Ağa. The complex was destroyed by a fire in 1783 shortly after its completion and was immediately rebuilt. A fire in 1911 destroyed the madrasah, and subsequent road construction work destroyed many other auxiliary structures to the mosque. The mosque was built on a high terrace over a complex of vaulted shops, whose rents were intended to financially support the </information>

<answer> no. <
<warning> My previous action is invalid. If I want to search, I should put the query between <search> and </search>. If I want to give the final answer, I should put the answer between <answer> and </answer>. Let me try again. </warning>
<warning> Now, I can no longer call <search></search> for external information. I must analyze and answer based on the existing information. Now let me make a final thought and give an answer. </warning>

<answer> **Yes.** </answer>
\end{lstlisting}
\end{fewshotbox}

\subsection{Search-R1-v0.3}
\label{appendix:case_study_search_r1}
\begin{fewshotbox}{Search-R1-v0.3 {\color{red}{(Incorrect)}}}
    % Metadata: ID 삭제, Level과 Source만 배치
    \small
    \begin{tabularx}{\textwidth}{@{}l X  l X l X@{}}
        \textbf{Level:} & Hard & \textbf{Source:} & HotpotQA & \textbf{Reasoning Type:} Comparison\\
    \end{tabularx}
    \vspace{0.01mm}
    \hrule
    \vspace{4mm}

    % Prompt Section
    \textbf{\large [Question]} \\
\begin{lstlisting}
Are the Laleli Mosque and Esma Sultan Mansion located in the same neighborhood?
\end{lstlisting}  
    \vspace{6mm}

    \textbf{\large [Gold Answer]} \\
\begin{lstlisting}
No
\end{lstlisting}  
    \vspace{6mm}

    \textbf{\large [Titles of Support Evidence Documents ($D$)]} \\
\begin{lstlisting}
Laleli Mosque, Esma Sultan Mansion
\end{lstlisting}  
    \vspace{6mm}

    % Gold Response Section (연보라색 통일)
    \textbf{\large [Response]} \\
\begin{lstlisting}
A: <search>are the laleli mosque and esma sultan mansion located in the same neighborhood</search>

<information>Doc 1(Title: "Esma Sultan Mansion") Esma Sultan Mansion The Esma Sultan Mansion (), a historical yalı () located at Bosphorus in Ortaköy neighborhood of Istanbul, Turkey and named after its original owner Esma Sultan, is used today as a cultural center after being redeveloped. The three-storey brick building was designed by the renowned architect Sarkis Balyan and finished in 1875 next to Ortaköy Mosque. It was presented to Esma Sultan, the daughter of Ottoman Sultan Abdulaziz, as a wedding gift. The mansion remained in the possession of the Ottoman dynasty until 1915. The building was subsequently used first as a tobacco warehouse and then as
Doc 2(Title: "Esma Sultan Mansion") guests in the ground floor and 600 guests in the first floor. The venue also hosts concerts of the Istanbul International Jazz Festival and the Istanbul International Music Festival. Esma Sultan Mansion The Esma Sultan Mansion (), a historical yalı () located at Bosphorus in Ortaköy neighborhood of Istanbul, Turkey and named after its original owner Esma Sultan, is used today as a cultural center after being redeveloped. The three-storey brick building was designed by the renowned architect Sarkis Balyan and finished in 1875 next to Ortaköy Mosque. It was presented to Esma Sultan, the daughter of Ottoman Sultan Abdulaziz,
Doc 3(Title: "Laleli Mosque") Laleli Mosque The Laleli Mosque () is an 18th-century Ottoman imperial mosque located in Laleli, Fatih, Istanbul, Turkey. The Laleli Mosque was built by Sultan Mustafa III from 1760–1763, designed in the baroque style by Ottoman imperial architect Mehmet Tahir Ağa. The complex was destroyed by a fire in 1783 shortly after its completion and was immediately rebuilt. A fire in 1911 destroyed the madrasah, and subsequent road construction work destroyed many other auxiliary structures to the mosque. The mosque was built on a high terrace over a complex of vaulted shops, whose rents were intended to financially support the</information>

<answer> **Yes** </answer>
\end{lstlisting}
\end{fewshotbox}

\clearpage
\section{Few-Shot Examples}
\label{appendix:fewshot_examples}
 % $X_{warm}$ in Section~\ref{sec:warmstart}
\begin{fewshotbox}{Example 1}
    % Metadata: ID 삭제, Level과 Source만 배치
    \small
    \begin{tabularx}{\textwidth}{@{}l X l X l X@{}}
        \textbf{Level:} & Easy & \textbf{Source:} & HotpotQA & \textbf{Annotation by:} DeepSeek-R1-Distill-Qwen-32B\\
    \end{tabularx}
    \vspace{0.01mm}
    \hrule
    \vspace{4mm}

    % Prompt Section
    \textbf{\large [Input]} \\
\begin{lstlisting}
[user]
You are an intelligent reasoning assistant with access to a web search tool to help you obtain accurate information. When answering the user's questions, please adhere to the following guidelines:

1. **Uncertainty Handling:**  
   - Encourage the use of web search to solve the problems mentioned during reasoning. 
   - If you are uncertain or do not have sufficient knowledge about a specific detail, you **must** perform a web search. 
   - To perform a web search the format **must** be in the format mentioned in point 2 Web Search Format
   - You can perform multiple web searches to ensure that the answer is correct.
   - The web search should be initiated only within your internal reasoning (the "thinking" phase), and it should not appear in your final answer to the user.

2. **Web Search Format:**  
   - When calling the web search tool, use the following exact format:
     ```
     <begin_search> your query here </end_search>
     ```
   - After you perform a web search, you will receive results enclosed in:
     ```
     <search_result> ... </search_result>
     ```
   - You can then use this information to further refine your reasoning and answer.

3. **Process:**  
   - If you encounter any uncertainty or unknown knowledge, embed a web search query within your internal thoughts (surrounded by `<think></think>` tags) using the specified format.
   - Incorporate the search results into your reasoning before finalizing your answer.
   - You should provide your final answer in the format \boxed{YOUR_ANSWER}. 

Now, please answer the user's question below: 
The Voyage that Shook the World was commissioned by what non-profit organisation that has locations in Australia, Canada, New Zealand, South Africa, the United Kingdom, and the United States?
    \end{lstlisting}  
    \vspace{6mm}

    % Gold Response Section (연보라색 통일)
    \textbf{\large [Response]} \\
\begin{lstlisting}
<think>
Okay, so I need to figure out which non-profit organization commissioned the voyage known as "The Voyage that Shook the World." The user mentioned that this organization has locations in several countries, including Australia, Canada, New Zealand, South Africa, the UK, and the US. Hmm, that gives me some clues about its global presence.

First, I'm thinking about well-known non-profits that operate internationally. Organizations like the Red Cross come to mind, but I don't recall them being involved in something called "The Voyage that Shook the World." Maybe it's something related to exploration or maritime activities? That could make sense given the word "voyage."

I remember hearing about the HMS Beagle and Charles Darwin's voyage, which was significant in shaping evolutionary theory. Could this be related? There's a project or initiative that recreates historical voyages for educational purposes. I think it's called something like the Beagle Project. Let me check that.

Wait, the Beagle Project is indeed a non-profit organization. They focus on recreating historical scientific voyages, including the one that Darwin took. Their mission is to promote science education and exploration. Given that, it makes sense that they would commission a voyage that has historical significance, hence the name "The Voyage that Shook the World."

To confirm, I should probably look up if the Beagle Project operates in all those countries mentioned. I'll do a quick search to verify their international presence. 

[Initiating web search]
<begin_search>Beagle Project locations</end_search>

<search_result> result 1: "Beagle Brigade"
they are placed in adoptive homes. The USDA National Detector Dog Training Center in Newnan, Georgia, receives beagles donated by private owners, breeders, and animal shelters.
result 2: "Beagle station"
Beagle is a station on the Port Authority of Allegheny County's light rail network, located in Bethel Park, Pennsylvania. The street level stop is designed as a small commuter stop, serving area residents who walk to the train so they can be taken toward Downtown Pittsburgh. </search_result>

Hmm, the search results didn't directly mention the Beagle Project having locations in those countries. Maybe I need a different approach. Perhaps I should search specifically for "The Voyage that Shook the World" and see which organization commissioned it.

[Initiating another web search]
<begin_search>The Voyage that Shook the World commissioned by</end_search>

<search_result> result 1: "The Voyage that Shook the World"
The Voyage That Shook The World is a 2009 dramatised documentary film commissioned by Creation Ministries International, a Christian Young Earth creationist organisation, and produced by Fathom Media. It was released to mark the 200th anniversary of Charles Darwin's birth and the 150th anniversary of the publication of his seminal work "On the Origin of Species". A historian featured in the film has stated that the creationist backing of the film had been concealed when he agreed to take part, that the editing of his words could give a false impression of his views, and that the film presents a historically distorted portrait of Darwin. Creation Ministries agreed that they had set up a "front company" to approach experts. They denied any deception and stated that one of the interviewees had admitted that while the producers choose comments they "didn't distort what we said", and compared their approach to that used by the BBC in making documentaries. The three historians featured in the film subsequently issued a statement that they had been misrepresented by the film company's selective reconstruction of Darwin's voyage. CMI countered these claims with extended quotes from the interviews of the historians. Their response was described as appearing to be sound in specific aspects by American skeptic Jim Lippard, who had not seen the
result 2: "Henry Roberts (Royal Navy officer)"
Henry Roberts (1756-1796) was an officer in the Royal Navy who served with Captain Cook on his last two voyages. Roberts served as lieutenant on Cook's HMS "Discovery", where he was entrusted with many hydrographic and cartographic tasks, and also met then-midshipman George Vancouver. Roberts spent many years after that voyage preparing the detailed charts that went into Cook's posthumous "Atlas". Roberts also served on where he was to make a painting of the ship.  In 1790, Roberts was appointed to command a newly built HMS "Discovery" on another round-the-world voyage and selected George Vancouver as his first lieutenant. However, the Nootka Crisis called both men to duty elsewhere, and upon its resolution, Vancouver was given command of the historic voyage. Roberts died on 25 August 1796, Captain of HMS "Undaunted" in the West Indies, where he contracted yellow fever. Vancouver named Point Roberts after Henry Roberts. </search_result>

Ah, from the first result, it seems that "The Voyage that Shook the World" was commissioned by Creation Ministries International (CMI). They're a Christian Young Earth creationist organization. This aligns with the fact that the voyage in question is likely referring to Charles Darwin's journey on the HMS Beagle, which had a significant impact on the world through his theories on evolution. 

CMI operates in multiple countries, including those listed by the user: Australia, Canada, New Zealand, South Africa, the UK, and the US. This fits perfectly with the question's details. So, despite my initial thought about the Beagle Project, the correct answer is Creation Ministries International.
</think>

The non-profit organization that commissioned "The Voyage that Shook the World" is Creation Ministries International (CMI).

\boxed{**Creation Ministries International**}
\end{lstlisting}
\end{fewshotbox}
% --- Example 2 (동일 구조) ---
\begin{fewshotbox}{Example 2}
    \small
    \begin{tabularx}{\textwidth}{@{}l X l X l X@{}}
        \textbf{Level:} & Easy & \textbf{Source:} & HotpotQA & \textbf{Annotation by:} DeepSeek-R1-Distill-Qwen-32B\\
    \end{tabularx}
        \vspace{0.01mm}
 \hrule \vspace{4mm}

    \textbf{\large [Input]} \\
\begin{lstlisting}
[user]
You are an intelligent reasoning assistant with access to a web search tool to help you obtain accurate information. When answering the user's questions, please adhere to the following guidelines:

1. **Uncertainty Handling:**  
   - Encourage the use of web search to solve the problems mentioned during reasoning. 
   - If you are uncertain or do not have sufficient knowledge about a specific detail, you **must** perform a web search. 
   - To perform a web search the format **must** be in the format mentioned in point 2 Web Search Format
   - You can perform multiple web searches to ensure that the answer is correct.
   - The web search should be initiated only within your internal reasoning (the "thinking" phase), and it should not appear in your final answer to the user.

2. **Web Search Format:**  
   - When calling the web search tool, use the following exact format:
     ```
     <begin_search> your query here </end_search>
     ```
   - After you perform a web search, you will receive results enclosed in:
     ```
     <search_result> ... </search_result>
     ```
   - You can then use this information to further refine your reasoning and answer.

3. **Process:**  
   - If you encounter any uncertainty or unknown knowledge, embed a web search query within your internal thoughts (surrounded by `<think></think>` tags) using the specified format.
   - Incorporate the search results into your reasoning before finalizing your answer.
   - You should provide your final answer in the format \boxed{YOUR_ANSWER}. 

Now, please answer the user's question below: 
The World Classical Network (WCN) is an internet-delivery classical radio broadcasting service owned by which company, in Hyannis, the largest of the seven villages in the town of Barnstable, Massachusetts, in the United States?
\end{lstlisting}
    
    \vspace{6mm}

    \textbf{\large [Response]} \\
\begin{lstlisting}
<think>
Okay, so I need to figure out which company owns the World Classical Network (WCN) based in Hyannis, Massachusetts. Hmm, I don't know much about WCN, so I'll have to look it up. Let me think about how to approach this.

First, I know that Hyannis is part of Barnstable, Massachusetts. It's a town with several villages, and Hyannis is the largest one. So, WCN is located there. But I'm not sure which company operates it. Maybe it's a local media company or part of a larger network.

I remember that sometimes classical music stations are operated by public radio networks or independent broadcasters. Maybe WCN is part of a bigger organization. Alternatively, it could be a subsidiary of a media conglomerate.

To get accurate information, I should perform a web search. I'll search for "World Classical Network owner Hyannis Massachusetts" to see if any reliable sources mention the company. Let me structure the search query properly.

<begin_search>World Classical Network owner Hyannis Massachusetts</end_search>

<search_result> result 1: "World Classical Network"
The World Classical Network (WCN) is an internet-delivery classical radio broadcasting service owned by CCB Media in Hyannis, Massachusetts. The service has been in existence since February 1998, when it was owned and operated by Charles River Broadcasting. It filled a void left by the demise of the U. S. version of Classic FM, the classical music satellite service of SW Networks. Stations. WCN programming is carried on Cape Cod Broadcasting's WFCC-FM, licensed to Chatham, Massachusetts. WFCC is the WCN flagship station. The service is also syndicated, with the following affiliates: Hosts. Current hosts on the World Classical Network include Mark Calder, Don Spencer, and Larry King. Longtime classical announcer Steve Murphy joined the WCN as morning host on September 16, 2008.
result 2: "Charles River Broadcasting"
Charles River Broadcasting was the owner of three classical music stations, one classic rock station, one CNN Headline News relay in Rhode Island, and a syndicated classical satellite delivery program service. History. The company started in 1948 when WCRB first signed on at 1330 AM. In 1954, Charles River Broadcasting added WCRB-FM at 102.5 FM. In the mid 1970s, WCRB's programming was removed from the 1330 AM signal, which was relaunched as WHET, with a big-band/adult standards format. (WHET was sold in 1978 and is now WRCA.) By the early 1990s, Charles River Broadcasting and WCRB entered into a commitment for the station to continue running classical music until 2092. Later in the decade, the company purchased WFCC-FM Chatham, WKPE-FM Orleans (both on Cape Cod), WCRI Block Island/Westerly (formerly WVBI), and WCNX Hope Valley, Rhode Island (formerly WJJF) (both in Rhode Island) from their respective local owners. In addition, a satellite service named the World Classical Network was started by Charles River Broadcasting. The company explored a sale possibility for </search_result>

Alright, looking at the search results, the first result mentions that WCN is owned by CCB Media in Hyannis, Massachusetts. It also notes that Charles River Broadcasting was the original owner but later became CCB Media. The second result confirms that Charles River Broadcasting started the World Classical Network and later sold it or rebranded under CCB Media.

So, putting this together, the current owner of WCN is CCB Media. They took over from Charles River Broadcasting, which had initially established the service. Therefore, the company in Hyannis that owns WCN is CCB Media.
</think>

The World Classical Network (WCN) is owned by CCB Media in Hyannis, Massachusetts.

\boxed{**CCB Media**}
\end{lstlisting}

\end{fewshotbox}

\begin{fewshotbox}{Example 3}
    \small
    \begin{tabularx}{\textwidth}{@{}l X l X l X@{}}
        \textbf{Level:} & Medium & \textbf{Source:} & HotpotQA & \textbf{Annotation by:} DeepSeek-R1-Distill-Qwen-32B\\
    \end{tabularx}
        \vspace{0.01mm}
 \hrule \vspace{4mm}

    \textbf{\large [Input]} \\
\begin{lstlisting}
[user]
You are an intelligent reasoning assistant with access to a web search tool to help you obtain accurate information. When answering the user's questions, please adhere to the following guidelines:

1. **Uncertainty Handling:**  
   - Encourage the use of web search to solve the problems mentioned during reasoning. 
   - If you are uncertain or do not have sufficient knowledge about a specific detail, you **must** perform a web search. 
   - To perform a web search the format **must** be in the format mentioned in point 2 Web Search Format
   - You can perform multiple web searches to ensure that the answer is correct.
   - The web search should be initiated only within your internal reasoning (the "thinking" phase), and it should not appear in your final answer to the user.

2. **Web Search Format:**  
   - When calling the web search tool, use the following exact format:
     ```
     <begin_search> your query here </end_search>
     ```
   - After you perform a web search, you will receive results enclosed in:
     ```
     <search_result> ... </search_result>
     ```
   - You can then use this information to further refine your reasoning and answer.

3. **Process:**  
   - If you encounter any uncertainty or unknown knowledge, embed a web search query within your internal thoughts (surrounded by `<think></think>` tags) using the specified format.
   - Incorporate the search results into your reasoning before finalizing your answer.
   - You should provide your final answer in the format \boxed{YOUR_ANSWER}. 

Now, please answer the user's question below: 
The seventh FIFA Women's World Cup, in which Sukanya Chor Charoenying played, occurred in what year?
\end{lstlisting}
    
    \vspace{6mm}

    \textbf{\large [Response]} \\
\begin{lstlisting}
<think>
Alright, I need to figure out the year when the seventh FIFA Women's World Cup took place, specifically the one in which Sukanya Chor Charoenying played. Let me start by recalling what I know about the FIFA Women's World Cup.

I remember that the first Women's World Cup was held in 1991, so each subsequent tournament would follow every four years. Let me count them out:

1. 1991 - First edition
2. 1995 - Second
3. 1999 - Third
4. 2003 - Fourth
5. 2007 - Fifth
6. 2011 - Sixth
7. 2015 - Seventh

Wait, does that add up correctly? From 1991 to 1995 is four years, then each subsequent is another four. So yes, the seventh would be 2015. But I'm not entirely sure, so I should double-check this information.

Additionally, I need to confirm whether Sukanya Chor Charoenying participated in the 2015 tournament. I'm not certain about her participation in that specific year, so I should look into her career timeline as well.

Let me structure my thoughts:

1. Confirm the year of the seventh FIFA Women's World Cup.
2. Verify if Sukanya Chor Charoenying played in that edition.

Starting with the first point, I think the seventh edition was in 2015, but I want to make sure. I'll perform a web search to confirm the years of each Women's World Cup.

<begin_search>FIFA Women's World Cup editions years</end_search>

<search_result> result 1: "FIFA Women's World Cup"
The FIFA Women's World Cup is an international football competition contested by the senior women's national teams of the members of Fédération Internationale de Football Association (FIFA), the sport's international governing body. The competition has been held every four years since 1991, when the inaugural tournament, then called the FIFA Women's World Championship, was held in China. Under the tournament's current format, national teams vie for 23 slots in a three-year qualification phase. The host nation's team is automatically entered as the 24th slot. The tournament proper, alternatively called the "World Cup Finals", is contested at venues within the host nation(s) over a period of about one month. The eight FIFA Women's World Cup tournaments have been won by four national teams. The United States has won four times, and is the current champion after winning it at the 2019 tournament in France. The other winners are Germany, with two titles, and Japan and Norway with one title each. Six countries have hosted the Women's World Cup. China and the United States have each hosted the tournament twice, while Canada, France, Germany, and Sweden have each hosted it once. Format. Qualification. Qualifying tournaments are held within the six FIFA continental zones (Africa, Asia, North and Central America and Caribbean, South
result 2: "FIFA Women's World Cup"
automatically chosen to host the 2007 FIFA Women's World Cup. Germany hosted the 2011 FIFA Women's World Cup, as decided by vote in October 2007. In March 2011, FIFA awarded Canada the right to host the 2015 FIFA Women's World Cup. The 2015 edition saw the field expand from 16 to 24 teams. During the 2015 FIFA Women's World Cup, both Formiga of Brazil and Homare Sawa of Japan appeared in their record sixth World Cup, a feat that had never been achieved before by either female or male players. Christie Pearce is the oldest player to ever play in a Women's World Cup match, at the age of 40 years. In March 2015, FIFA awarded France the right to host the 2019 FIFA Women's World Cup over South Korea. Trophy. The current trophy was designed in 1998 for the 1999 tournament, and takes the form of a spiral band, enclosing a football at the top, that aims to capture the athleticism, dynamism, and elegance of international women's football. In the 2010s, it was fitted with a cone-shaped base. Underneath the base, the name of each of the tournament's previous winners is engraved. The trophy </search_result>

From the search results, I see that the 2015 FIFA Women's World Cup was the seventh edition. It was hosted by Canada, and it was the first time the tournament expanded to 24 teams. So that confirms the year as 2015.

Now, moving on to the second point: Sukanya Chor Charoenying's participation. I need to check if she played in the 2015 Women's World Cup.

<begin_search>Sukanya Chor Charoenying FIFA Women's World Cup 2015</end_search>

<search_result> result 1: "Sukanya Chor Charoenying"
Sukanya Chor Charoenying (born 24 November 1987) is a Thai women's international footballer who plays as a goalkeeper. She is a member of the Thailand women's national football team. She was part of the team at the 2015 FIFA Women's World Cup. On club level she plays for Air Force United in Thailand.
result 2: "Sudarat Chuchuen"
Sudarat Chuchuen (; born 19 June 1997) is a Thai footballer who plays as a defender for Sisaket FC and the Thailand women's national team. She was selected for the 2019 FIFA Women's World Cup. </search_result>

The first result clearly states that Sukanya Chor Charoenying was part of the Thailand women's national team at the 2015 FIFA Women's World Cup. The second result mentions another player, Sudarat Chuchuen, who played in the 2019 edition, but that's not relevant here.

Therefore, combining both pieces of information, the seventh FIFA Women's World Cup occurred in 2015, and Sukanya Chor Charoenying was indeed a participant in that tournament.
</think>

The seventh FIFA Women's World Cup took place in 2015, and Sukanya Chor Charoenying was part of the Thailand team during that event.

\boxed{**2015**}
\end{lstlisting}
\end{fewshotbox}

\begin{fewshotbox}{Example 4}
    \small
    \begin{tabularx}{\textwidth}{@{}l X l X l X@{}}
        \textbf{Level:} & Easy & \textbf{Source:} & HotpotQA & \textbf{Annotation by:} DeepSeek-R1-Distill-Qwen-32B\\
    \end{tabularx}
        \vspace{0.01mm}
 \hrule \vspace{4mm}

    \textbf{\large [Input]} \\
\begin{lstlisting}
[user]
You are an intelligent reasoning assistant with access to a web search tool to help you obtain accurate information. When answering the user's questions, please adhere to the following guidelines:

1. **Uncertainty Handling:**  
   - Encourage the use of web search to solve the problems mentioned during reasoning. 
   - If you are uncertain or do not have sufficient knowledge about a specific detail, you **must** perform a web search. 
   - To perform a web search the format **must** be in the format mentioned in point 2 Web Search Format
   - You can perform multiple web searches to ensure that the answer is correct.
   - The web search should be initiated only within your internal reasoning (the "thinking" phase), and it should not appear in your final answer to the user.

2. **Web Search Format:**  
   - When calling the web search tool, use the following exact format:
     ```
     <begin_search> your query here </end_search>
     ```
   - After you perform a web search, you will receive results enclosed in:
     ```
     <search_result> ... </search_result>
     ```
   - You can then use this information to further refine your reasoning and answer.

3. **Process:**  
   - If you encounter any uncertainty or unknown knowledge, embed a web search query within your internal thoughts (surrounded by `<think></think>` tags) using the specified format.
   - Incorporate the search results into your reasoning before finalizing your answer.
   - You should provide your final answer in the format \boxed{YOUR_ANSWER}. 

Now, please answer the user's question below: 
Who founded the Honest Conpnay along with Brian Lee
\end{lstlisting}
    
    \vspace{6mm}

    \textbf{\large [Response]} \\
\begin{lstlisting}
<think>
Okay, so I need to figure out who founded the Honest Company along with Brian Lee. I remember hearing about the Honest Company being a company that makes baby products and other household items, but I'm not exactly sure about their founders. I think it might involve someone well-known, maybe from the entertainment industry? 

Wait, wasn't there a famous actor involved? I think it was someone like Jessica Alba. I recall seeing her name associated with Honest Company somewhere. Let me try to remember. Yes, Jessica Alba is an actress, and she started a company focused on natural and safe products for babies. That sounds right. 

But I'm not entirely certain if Brian Lee is her co-founder or if there's another person. Maybe Brian Lee is another partner or investor? I'm a bit fuzzy on the details. To make sure, I should look up the founders of the Honest Company. 

I'll perform a quick search to confirm. 

<begin_search>founders of the honest company</end_search>

<search_result> result 1: "The Honest Company"
The Honest Company is an American consumer goods company, founded by actress Jessica Alba. It emphasizes household products to supply the marketplace for ethical consumerism. The company had $250 million in 2016 sales, and was valued shy of $1 billion as of October 2017. The Honest Company has raised multiple rounds of venture capital, and was anticipating an initial public offering as of 2016. Honest serves the United States and Canada, and planned to launch its beauty line products in Western Europe in 2019. History. Company founder Jessica Alba was inspired by the 2008 birth of her first child, Honor, and her own history of childhood illnesses to create a company that provided an alternative to the prevalent baby products with ingredients such as petrochemicals and synthetic fragrances. She was compelled to become serious about this venture when one of her mother's baby laundry recommendations caused her to have a welt outbreak. It took Alba three years to find her business partners Brian Lee, Sean Kane, and Christopher Gavigan. Despite advice that she should start small with a singular focus, Alba launched the company in 2012 with 17 products. Christopher Gavigan wrote the book, Healthy Child Healthy World, that addresses all the different toxins that children are exposed to in traditional household
result 2: "Honest Tea"
Honest Tea (U.S.) is a bottled organic tea company based in Bethesda, Maryland. It was founded in 1998 by Seth Goldman and Barry Nalebuff. The company is a wholly owned subsidiary of The Coca-Cola Company. The name is a pun on the word "honesty". History. Honest Tea was founded in 1998 by Seth Goldman, a graduate of Harvard and the Yale School of Management who got the idea to found a beverage company while he was at business school. There were two key ideas behind the company: to bring beverages to market that were organic and not as heavily sweetened, and to conduct the business using fair trade principles. Goldman and one of his Yale professors, Barry Nalebuff, conceived the idea for the company and raised $500,000 in startup funds from friends and family. The first order for Honest Tea was from Fresh Fields, an East Coast natural grocery chain that was later acquired by Whole Foods Market. By 2006, Honest Tea had revenue of $13.5 million and was selling about 1.5 million cases a year. The Coca-Cola Company purchased a 40% stake of the U.S. company at $43 million in 2008. It bought the rest of the company in </search_result>

Alright, from the search results, it seems that the Honest Company, which is different from Honest Tea, was indeed founded by Jessica Alba. The other co-founders listed are Brian Lee, Sean Kane, and Christopher Gavigan. So, Brian Lee is definitely one of the co-founders alongside Jessica Alba. 

Just to clarify, the search result mentions that Jessica Alba found the company with Brian Lee, Sean Kane, and Christopher Gavigan. They started it in 2012 with 17 products. So, the answer should be that Jessica Alba founded the Honest Company along with Brian Lee, Sean Kane, and Christopher Gavigan. However, the question specifically asks who founded it along with Brian Lee, so the primary co-founder is Jessica Alba.
</think>

The Honest Company was founded by Jessica Alba along with Brian Lee, Sean Kane, and Christopher Gavigan. Therefore, the co-founder alongside Brian Lee is Jessica Alba.

\boxed{**Jessica Alba**}
\end{lstlisting}
\end{fewshotbox}

\end{document}